\RequirePackage[svgnames]{xcolor}

\documentclass[10pt,letterpaper]{mystyle}

\usepackage[all]{hypcap}
\usepackage[comma,authoryear,compress]{natbib}
\usepackage{hyperref}
\hypersetup{
    colorlinks = true,
    citecolor = {YaleBlue},
}

\usepackage{algorithm}
\usepackage{algorithmicx}
\usepackage{algpseudocode}
\usepackage{microtype}
\usepackage{graphicx}
\expandafter\def\csname ver@subfig.sty\endcsname{}
\usepackage{booktabs}
\usepackage{float}
\usepackage{bigstrut}
\usepackage{amsmath}
\usepackage{amssymb}
\usepackage{mathtools}
\usepackage{amsthm}
\usepackage{mathrsfs}
\usepackage{nicefrac}
\usepackage{dsfont}
\usepackage{enumitem}
\usepackage{subcaption}
\usepackage{cleveref}
\usepackage{titletoc}
\usepackage{siunitx}
\usepackage{tabularx}
\usepackage{colortbl}
\usepackage{makecell}
\usepackage{array}
\usepackage{geometry}
\usepackage[utf8]{inputenc}
\usepackage[T1]{fontenc}
\usepackage{url}
\usepackage{amsfonts}
\usepackage{wrapfig}
\usepackage{stackengine}
\usepackage{adjustbox}
\usepackage{rotating}
\usepackage{pifont}
\usepackage{tikz}
\usepackage{multirow}
\usepackage{xcolor}
\usepackage{lmodern}

\usepackage{titlesec}
\titlespacing*{\paragraph}{0pt}{0ex plus 0ex minus 0ex}{0.5em}
\titlespacing*{\subsection}{0pt}{0ex plus 0ex minus 0ex}{0ex plus 0ex}

\definecolor{blanchedalmond}{rgb}{1.0, 0.92, 0.8}
\definecolor{carmine}{rgb}{0.59, 0.0, 0.09}
\definecolor{lightblue}{rgb}{0.22,0.45,0.70}%

\renewcommand{\mathbf}{\boldsymbol}

\makeatletter
\def\Ddots{\mathinner{\mkern1mu\raise\p@
\vbox{\kern7\p@\hbox{.}}\mkern2mu
\raise4\p@\hbox{.}\mkern2mu\raise7\p@\hbox{.}\mkern1mu}}
\makeatother

\definecolor{amaranth}{rgb}{0.9, 0.17, 0.31}
\definecolor{antiquebrass}{rgb}{0.8, 0.58, 0.46}
\definecolor{antiquefuchsia}{rgb}{0.57, 0.36, 0.51}
\definecolor{chromeyellow}{rgb}{0.31, 0.47, 0.26}

\newtcolorbox{AIbox}[2][]{aibox,title=#2,#1}
\definecolor{lightblue}{rgb}{0.22,0.45,0.70}
\definecolor{Gray}{gray}{0.95}
\definecolor{Cornsilk}{rgb}{1.0, 0.97, 0.86}
\definecolor{darkgreen}{RGB}{0, 150, 0}
\definecolor{bGreen}{HTML}{E5F5D3}  
\definecolor{bCyan}{HTML}{D0F5F5}   
\definecolor{bPink}{HTML}{F8E0F0}   
\definecolor{bYellow}{HTML}{FFF4CC} 
\definecolor{guidancecolor}{HTML}{4A8EB0}
\definecolor{LowBg}{HTML}{EEF5FF}
\definecolor{MidBg}{HTML}{EDF9EE}
\definecolor{HighBg}{HTML}{FFF4E8}
\definecolor{AllBg}{HTML}{F3F3F3}
\definecolor{LowHead}{HTML}{DCEBFF}
\definecolor{MidHead}{HTML}{DAF2DD}
\definecolor{HighHead}{HTML}{FFE6CC}
\definecolor{AllHead}{HTML}{E6E6E6}
\definecolor{SectionBg}{HTML}{F7F7F7}
\definecolor{plotblue}{RGB}{106,178,214}
\definecolor{plotgreen}{RGB}{118,190,150}
\definecolor{plotteal}{RGB}{108,190,178}
\definecolor{plotcoral}{RGB}{236,118,126}
\definecolor{plotcream}{RGB}{248,241,220}
\definecolor{plotgray}{RGB}{235,235,235}
\definecolor{AvgCell}{RGB}{59,130,246}
\definecolor{PassCell}{RGB}{249,115,22}
\definecolor{BlockHeader}{RGB}{242,244,247}
\definecolor{DeltaCell}{RGB}{239,68,68}
\definecolor{mildcolor}{RGB}{218, 156, 143}
\definecolor{EnsMainTwoAvgCell}{RGB}{88, 150, 95}
\definecolor{EnsMainTwoPassCell}{RGB}{170, 90, 135}
\definecolor{EnsMainTwoBlockHeader}{RGB}{242, 244, 247}
\definecolor{lightpink}{RGB}{255,245,245}
\definecolor{redtitle}{RGB}{210,50,50}
\definecolor{lightblue}{RGB}{240,247,255}
\definecolor{bluetitle}{RGB}{60,120,220}

\newtcolorbox{guidancebox}[1]{
    colback=white,               
    colframe=guidancecolor,      
    colbacktitle=guidancecolor,  
    coltitle=white,              
    fonttitle=\bfseries,         
    title={#1},                  
    boxrule=1pt,                 
    arc=2mm,                     
    auto outer arc,              
    left=6pt, right=6pt,         
    top=6pt, bottom=6pt,         
}

\newtcolorbox{discussion}[1]{
    colback=white,               
    colframe=mildcolor,          
    colbacktitle=mildcolor,      
    coltitle=white,              
    fonttitle=\bfseries,         
    title={#1},                  
    boxrule=1pt,                 
    arc=2mm,                     
    auto outer arc,              
    left=6pt, right=6pt,         
    top=6pt, bottom=6pt,         
}

\definecolor{colA}{HTML}{4682B4} 
\definecolor{colB}{HTML}{9370DB} 
\definecolor{colC}{HTML}{CD5C5C} 
\definecolor{colD}{HTML}{20B2AA} 
\definecolor{colE}{HTML}{D2691E} 
\definecolor{colF}{HTML}{3CB371} 
\definecolor{colG}{HTML}{708090} 
\definecolor{colTarget}{HTML}{DA70D6} 
\definecolor{EnsMainAvgCell}{RGB}{45, 160, 145}
\definecolor{EnsMainPassCell}{RGB}{145, 110, 185}
\definecolor{EnsMainBlockHeader}{RGB}{242, 244, 247}

\newcommand{\q}{\texttt{"}}
\definecolor{crystalC}{HTML}{1F77B4}
\definecolor{crystalB}{HTML}{DCEBFA}
\newcommand{\instCrystal}[1]{\colorbox{crystalB}{\strut #1}}
\newcommand{\catCrystal}[1]{\textcolor{crystalC}{#1}}
\definecolor{glowC}{HTML}{2CA02C}
\definecolor{glowB}{HTML}{DFF2DF}
\newcommand{\instGlow}[1]{\colorbox{glowB}{\strut #1}}
\newcommand{\catGlow}[1]{\textcolor{glowC}{#1}}
\definecolor{mithrilC}{HTML}{7F7F7F}
\definecolor{mithrilB}{HTML}{EAEAEA}
\newcommand{\instMithril}[1]{\colorbox{mithrilB}{\strut #1}}
\newcommand{\catMithril}[1]{\textcolor{mithrilC}{#1}}
\definecolor{sylphC}{HTML}{17BECF}
\definecolor{sylphB}{HTML}{DDF7FA}
\newcommand{\instSylph}[1]{\colorbox{sylphB}{\strut #1}}
\newcommand{\catSylph}[1]{\textcolor{sylphC}{#1}}
\definecolor{dragonC}{HTML}{D62728}
\definecolor{dragonB}{HTML}{F8DEDE}
\newcommand{\instDragon}[1]{\colorbox{dragonB}{\strut #1}}
\newcommand{\catDragon}[1]{\textcolor{dragonC}{#1}}
\definecolor{faerieC}{HTML}{E377C2}
\definecolor{faerieB}{HTML}{F8E1F1}
\newcommand{\instFaerie}[1]{\colorbox{faerieB}{\strut #1}}
\newcommand{\catFaerie}[1]{\textcolor{faerieC}{#1}}
\definecolor{moonC}{HTML}{9467BD}
\definecolor{moonB}{HTML}{ECE3F4}
\newcommand{\instMoon}[1]{\colorbox{moonB}{\strut #1}}
\newcommand{\catMoon}[1]{\textcolor{moonC}{#1}}
\definecolor{obsidianC}{HTML}{4D4D4D}
\definecolor{obsidianB}{HTML}{DDDDDD}
\newcommand{\instObsidian}[1]{\colorbox{obsidianB}{\strut #1}}
\newcommand{\catObsidian}[1]{\textcolor{obsidianC}{#1}}
\definecolor{sapphireC}{HTML}{003F9E}
\definecolor{sapphireB}{HTML}{DCE6FF}
\newcommand{\instSapphire}[1]{\colorbox{sapphireB}{\strut #1}}
\newcommand{\catSapphire}[1]{\textcolor{sapphireC}{#1}}
\definecolor{quartzC}{HTML}{00A6A6}
\definecolor{quartzB}{HTML}{D9F4F4}
\newcommand{\instQuartz}[1]{\colorbox{quartzB}{\strut #1}}
\newcommand{\catQuartz}[1]{\textcolor{quartzC}{#1}}
\definecolor{nymphC}{HTML}{6BAED6}
\definecolor{nymphB}{HTML}{E1F0F8}
\newcommand{\instNymph}[1]{\colorbox{nymphB}{\strut #1}}
\newcommand{\catNymph}[1]{\textcolor{nymphC}{#1}}
\definecolor{emberC}{HTML}{FF7F0E}
\definecolor{emberB}{HTML}{FFE8D6}
\newcommand{\instEmber}[1]{\colorbox{emberB}{\strut #1}}
\newcommand{\catEmber}[1]{\textcolor{emberC}{#1}}
\definecolor{rubyC}{HTML}{C51B7D}
\definecolor{rubyB}{HTML}{F6DCEB}
\newcommand{\instRuby}[1]{\colorbox{rubyB}{\strut #1}}
\newcommand{\catRuby}[1]{\textcolor{rubyC}{#1}}
\definecolor{unicornC}{HTML}{8C6BB1}
\definecolor{unicornB}{HTML}{EDE7F4}
\newcommand{\instUnicorn}[1]{\colorbox{unicornB}{\strut #1}}
\newcommand{\catUnicorn}[1]{\textcolor{unicornC}{#1}}
\definecolor{starC}{HTML}{B2182B}
\definecolor{starB}{HTML}{F8DDE1}
\newcommand{\instTarget}[1]{\colorbox{starB}{\strut #1}}
\newcommand{\catTarget}[1]{\textcolor{starC}{#1}}

\title{The Physics of Multi-Turn Long-Horizon Planning: From Pre-training to Post-training via Single- and Multi-Teacher On-Policy Agentic Distillation}

\runningtitle{The Physics of Multi-Turn Long-Horizon Planning}

\author[1,2]{Tianyi Men}
\author[1,2]{Zhuoran Jin}
\author[1,2]{Kang Liu}
\author[1,2,\dag]{Jun Zhao}

\affil[1]{The Key Laboratory of Cognition and Decision Intelligence for Complex Systems,\protect\newline Institute of Automation, Chinese Academy of Sciences, Beijing, China}

\affil[2]{School of Artificial Intelligence, University of Chinese Academy of Sciences, Beijing, China}

\affil[ ]{\{tianyi.men, zhuoran.jin, kliu, jzhao\}@nlpr.ia.ac.cn}

\begin{document}
\def\thefootnote{\dag}\footnotetext{Corresponding author.}\def\thefootnote{\arabic{footnote}}

\begin{abstract}
\textbf{Multi-turn long-horizon planning} is a critical capability for foundation model agents, yet how to fundamentally improve it across different training stages remains unclear. Existing models are trained on uncontrollable and opaque Internet data, making it difficult to identify how planning ability is acquired, shaped, and integrated. To address this challenge, we introduce a unified and controlled multi-turn environment that enables precise control over task length, data quality, planning knowledge, and planning patterns, and systematically study long-horizon planning across three stages.
\textbf{(1) Planning ability acquisition during pre-training.} We study data format, distribution, and quality. Explicit world model construction through chain-of-thought state transition modeling yields stronger long-horizon generalization than direct action prediction. Atomic skills alone are insufficient for compositional generalization, whereas a small amount of long-horizon data substantially improves planning ability. Moreover, suboptimal trajectories severely impair performance because decision errors accumulate and amplify over long horizons.
\textbf{(2) Planning ability shaping via GRPO and OPD post-training.} From the perspective of mutual information, we distinguish general planning patterns from task-specific planning knowledge. For planning patterns, we identify three applicability regions of post-training: unnecessary, effective, and unsupported. Compared with group relative policy optimization (GRPO), on-policy distillation (OPD) has a broader effective region under low-quality pre-training data and long planning horizons, as it provides more consistent update directions than sparse credit assignment when the teacher is ideal. For planning knowledge, however, distilling unseen multi-path procedures from a teacher with different underlying knowledge may impair the student's existing world modeling without fully establishing the new knowledge.
\textbf{(3) Planning ability integration through MOPD post-training.} We show that multi-teacher on-policy distillation (MOPD) integrates capabilities by converging to shared planning-pattern distributions across teachers. Compatible patterns enable cross-environment generalization, partially shared patterns support continual learning, while completely conflicting patterns cause severe catastrophic forgetting and cross-environment interference.
Overall, our study provides a unified understanding of how long-horizon planning ability is acquired, shaped, and integrated across training stages, offering practical insights for developing stronger agentic foundation models.


\textit{Homepage Link: \url{https://quester-one.github.io/PlanPhysWebsite/}}

\textit{Github Link: \url{https://github.com/Quester-one/PlanPhysCode}}

\textit{Huggingface Model: \url{https://huggingface.co/MultimodalAgent/TianyiMen_PlanPhys_Models}}

\textit{Huggingface Dataset: \url{https://huggingface.co/datasets/MultimodalAgent/TianyiMen_PlanPhys_Datasets}}

\end{abstract}

\maketitle
\vspace{3mm}
\clearpage
\section{Introduction}
\label{sec:introduction}

As foundation models advance toward agentic systems, \textbf{multi-turn long-horizon planning} ability becomes a critical capability for building real-world general intelligent agents~\citep{yuan2026osworld2,zhu2026edgebench,li2026long,zhang2026deepplanning}. Different from single-turn question answering or short-sequence decision making, long-horizon planning requires agents to decompose complex tasks, track environment states, model state transitions, and compose atomic skills through long-horizon interactions~\citep{li2026agentic,cao2025large}. Recently, benchmarks for multi-turn long-horizon planning, including OSWorld 2.0~\citep{yuan2026osworld2}, EdgeBench~\citep{zhu2026edgebench}, Long-Horizon-Terminal-Bench~\citep{li2026long}, and DeepPlanning~\citep{zhang2026deepplanning}, demonstrate that while foundation models achieve basic short-horizon planning capabilities, they still struggle significantly with long-horizon planning tasks.

Currently, improving the long-horizon planning ability of agents mainly relies on three stages: (1) \textbf{Large-scale pre-training} to learn general planning knowledge and planning patterns; (2) \textbf{RL-based post-training} to further optimize long-horizon planning strategies; and (3) \textbf{Multi-teacher model consolidation} in post-training to combine long-horizon planning abilities from multiple teachers. Despite these efforts, one core question remains unclear:

\begin{tcolorbox}[
    colback=lightpink,
    colframe=lightpink,
    boxrule=0pt,
    arc=8pt,
    left=12pt,
    right=12pt,
    top=3pt,
    bottom=3pt
]
\hspace*{-10pt}
\textcolor{redtitle}{\textbf{\textit{A Core Question for Multi-Turn Long-Horizon Planning.}}}

\hspace*{-12pt}
\textit{
How to \textbf{fundamentally} improve \textbf{long-horizon planning} ability of foundation models \textbf{at different training stages}?
}

\end{tcolorbox}

However, understanding and improving the long-horizon planning ability of foundation models still faces a fundamental challenge. The gap is that existing foundation models are trained on \textbf{uncontrollable and opaque Internet data}, making it difficult to systematically analyze where planning ability comes from and how it can be further strengthened. For example, it remains unclear whether the long-horizon planning ability exhibited by models already exists during pre-training, or it is a new ability activated during post-training. Therefore, when we seek to further strengthen long-horizon planning, as shown in Figure~\ref{fig:introduction}, we cannot answer the following three basic sub-questions from the perspective of the training stages:

\begin{tcolorbox}[
    colback=lightblue,
    colframe=lightblue,
    boxrule=0pt,
    arc=8pt,
    left=12pt,
    right=12pt,
    top=3pt,
    bottom=3pt
]
\hspace*{-10pt}
\textcolor{bluetitle}{\textbf{\textit{Three Basic Sub-questions for Multi-Turn Long-Horizon Planning.}}}

\hspace*{-6pt}
\textit{
\textbf{(1) Planning Ability Acquisition during Pre-training.} How does the model acquire long-horizon planning through pre-training, via world model internalization, atomic skill composition, and trajectory quality?
} 

\hspace*{-6pt}
\textit{
\textbf{(2) Planning Ability Shaping via GRPO and OPD Post-training.} How does RL shape long-horizon planning, and what are the applicable boundaries of RL algorithms in shaping planning knowledge and patterns?
}

\hspace*{-6pt}
\textit{
\textbf{(3) Planning Ability Integration through MOPD Post-training.} How does model consolidation integrate long-horizon planning abilities to enable cross-environment generalization, continual learning, and conflict resolution?
}

\end{tcolorbox}

To answer above questions, we propose \textbf{a unified and controlled multi-turn environment to study long-horizon planning ability (Section~\ref{sec:task})}. Unlike existing physics frameworks that are mainly based on single-turn mathematical reasoning~\citep{ye2025physics,zhang2025interplay}, our environment supports multi-turn long-horizon planning, as shown in Figure~\ref{fig:task}. It enables precise control over environment numbers, levels and categories, task lengths, data quality, planning knowledge, and planning patterns. Moreover, it is compatible with both SFT and RL paradigms. Based on this framework, we draw the following conclusions through three progressive stages:

\begin{figure}[h]
    \centering
    \includegraphics[width=\textwidth]{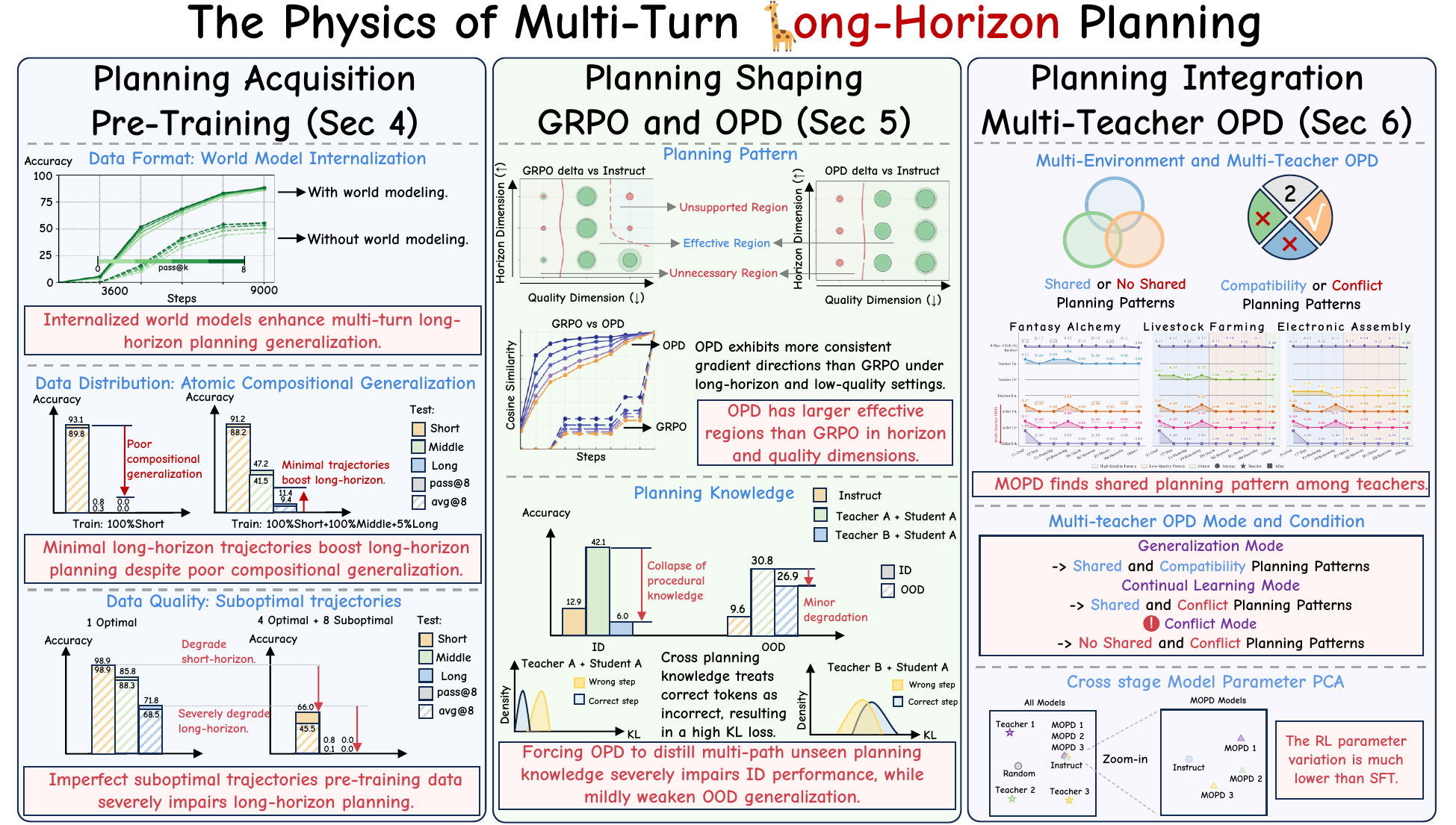}
    \captionsetup{skip=3pt}
    \caption{An overview of the studying into physics of multi-turn long-horizon planning. It studies the long-horizon planning ability across three training stages: \textbf{Large-scale pre-training}, \textbf{RL-based post-training (OPD and GRPO)}, and \textbf{Multi-teacher model consolidation post-training (MOPD)}. The giraffe icon is used to represent the “L” in “Long”. Its long neck also reflects that the agents need to look far ahead in long-horizon planning.}
    \label{fig:introduction} 
    \vspace{-15pt}
\end{figure}

\textbf{Planning ability acquisition during pre-training (Section~\ref{sec:pretrain}).}
At this stage, we analyze the acquisition of long-horizon planning ability from \textbf{three perspectives: data format, data distribution, and data quality}. 
(i) Data format: Models that explicitly construct an internal world model through chain-of-thought for state transition modeling exhibit stronger generalization ability in long-horizon planning than direct action prediction.
(ii) Data distribution: Although models struggle to achieve compositional generalization only through atomic skills, incorporating a small amount of long-horizon data during pre-training can enhance long-horizon planning ability. 
(iii) Data quality: Real-world pre-training data inevitably contains suboptimal trajectories alongside optimal ones. Since errors in long-sequence decision-making accumulate and amplify over time, such imperfect data severely impair long-horizon planning ability.

\textbf{Planning ability shaping via GRPO and OPD post-training (Section~\ref{sec:opd}).}
At this stage, we analyze planning ability through \textbf{the lens of mutual information and divide it into two aspects: planning patterns and planning knowledge}. 
The key difference is that planning patterns capture general planning behaviors shared across tasks, while planning knowledge contains task-specific procedures and solutions. (i) Planning patterns: Along the two dimensions of planning horizon and data quality mentioned in pre-training, the application space can be divided into three regions based on the applicability boundaries of RL: \textit{unnecessary region, effective region, unsupported region}. We compare mainstream post-training approaches including group relative policy optimization (GRPO) and on-policy distillation (OPD). GRPO has a narrower effective region. When pre-training data quality is low and the task requires long-horizon planning, OPD exhibits a broader effective boundary for shaping general planning patterns. This advantage stems from the fact that sparse credit assignment in RL introduces both correct and incorrect gradient update directions, whereas OPD provides more consistent and reliable update directions when the teacher is ideal. (ii) Planning knowledge: The situation is different. When the teacher and student models have different underlying procedural planning knowledge, directly applying OPD to distill unseen multi-path knowledge cannot guarantee effective transfer, even if the teacher achieves high accuracy. Instead, it may impair the student's existing in-domain world modeling and weaken its out-of-domain planning ability. This is because specific planning knowledge contains higher mutual information and requires precise, case-level updates. However, RL provides only small parameter updates, which are insufficient to fully replace old knowledge with new unseen knowledge. Therefore, the model may enter an intermediate state where prior knowledge is impaired while new planning knowledge remains not fully established.

\textbf{Planning ability integration through MOPD post-training (Section~\ref{sec:multi_opd}).}
At this stage, we analyze \textbf{cross-environment generalization, continual learning, and cross-environment conflicts from two aspects: whether different planning patterns are shared or compatible across different environments}. The experiments show that the key mechanism of multi-teacher on-policy distillation (MOPD) for cascaded consolidation is to find and converge to the shared distribution among multiple teachers. (i) Generalization mode: When different environments share compatible planning patterns, MOPD can break domain boundaries and successfully transfer abilities activated in one environment to unseen environments. (ii) Continual learning mode: When planning patterns conflict across environments, as long as some shared and useful pattern structures remain across environments, the model can adapt while mitigating severe forgetting, thereby supporting continual learning. (iii) Conflict mode: However, when different environments share no common planning patterns and exhibit completely conflict behavior, the student model may suffer from severe catastrophic forgetting as it overfits to the new teacher, leading to interference of cross-environment planning abilities.

\section{Preliminaries}
\label{sec:preliminaries}

\subsection{Agent Planning}

\paragraph{Planning Definition.} In this work, we formulate agent planning as a sequential decision-making problem. Given an initial environmental state $s_{0} \in \mathcal{S}$ and a specific target goal $g \in \mathcal{G}$, the objective of the agent is to generate a sequence of executable actions $A = \{a_1, a_2, \dots, a_T\}$. The execution of this sequence transitions the environment from the initial state $s_0$ to a final state $s_T$ that successfully satisfies the goal $g$. 

\paragraph{Agent Skill/Experience/Procedural Knowledge/Script Knowledge.} We define an agent skill (or experience or procedural knowledge or script knowledge) as a natural language mapping from a task description to a sequential list of sub-steps. Formally, a skill $k$ in the skill library $\mathcal{K}$ is represented as a tuple $k = (d, P)$. Here, $d$ represents the natural language description of the target task (e.g., "make a cup of coffee"), and $P = \langle p_1, p_2, \dots, p_m \rangle$ is an ordered sequence of sub-steps, where each step $p_i$ is also explicitly described in natural language (e.g., "find a mug", "pour water"). This format provides the agent with a clear recipe detailing exactly how a specific sub-goal should be systematically achieved. In practice, these skills can be utilized in two primary paradigms: either explicitly provided as external context, or implicitly internalized into the model's chain-of-thought (CoT) reasoning process. In this paper, we primarily focus on investigating the latter approach.

\paragraph{World Model.}
The world model aims to learn the rules of state transitions. Specifically, it models the state transition function $T(s_t, a_t) \rightarrow s_{t+1}$, which determines the next state given the current state and a specific action. In this process, procedural skills are not treated as separate, external modules; instead, they are directly internalized into the state transition function itself. By doing so, the model embeds these skills as the underlying rules driving state changes, enabling it to predict future states based on given actions.

\subsection{On-Policy Agentic Distillation}
Following~\citet{li2026rethinking}, we extend the setting from single-turn mathematical reasoning to multi-turn long-horizon planning.
On-Policy Agentic Distillation optimizes the student policy $\pi_\theta$ directly on its self-explored interactive trajectories. Formally, we define a $K$-turn trajectory $\tau = (g, s_0, \hat{Y}_1, s_1, \dots, \hat{Y}_K, s_K) \sim \pi_\theta$, where $g \in \mathcal{G}$ and $s_0 \in \mathcal{S}$ are the goal and initial state. At each turn $k$, conditioned on the accumulated context $H_k$, the student generates a unified sequence $\hat{Y}_k = (\hat{y}_{k,1}, \dots, \hat{y}_{k,T_k})$ that sequentially interleaves intermediate planning with an executable action, subsequently transitioning the environment to $s_k$.

At each token $t \in \{1, \dots, T_k\}$ within turn $k$, both the student and a teacher policy $\pi_{\text{teacher}}$ evaluate the sequence prefix to yield next-token distributions: 
\begin{equation}
    p_{k,t} \triangleq \pi_\theta(\cdot \mid H_k, \hat{Y}_{k,<t}) \quad \text{and} \quad q_{k,t} \triangleq \pi_{\text{teacher}}(\cdot \mid H_k, \hat{Y}_{k,<t}).
\end{equation}

By autoregressively factorizing the reverse Kullback-Leibler (KL) divergence over the entire multi-turn rollout, the Agent-OPD objective admits an exact token-level decomposition. This provides dense, step-by-step supervision on both the reasoning chain and the decision-making process:
\begin{equation}
    \mathcal{L}_{\text{Agent-OPD}}(\theta) = \mathbb{E}_{\tau \sim \pi_\theta} \left[ \sum_{k=1}^{K} \sum_{t=1}^{T_k} D_{\text{KL}}(p_{k,t} \parallel q_{k,t}) \right].
\end{equation}

\paragraph{PG-Style OPD vs GKD-Style OPD.} 
On-Policy Agentic Distillation primarily diverges into two paradigms based on how teacher feedback is utilized. Policy Gradient (PG) style OPD treats the per-token reverse KL divergence as a dense reward signal. Specifically, it computes the token-level advantage using a single-sample log-ratio approximation: $A_{k,t} \approx \log q_{k,t}(\hat{y}_{k,t}) - \log p_{k,t}(\hat{y}_{k,t})$ . The student is then updated via RL objectives to re-weight trajectory sampling probabilities: $\mathcal{L}_{\text{PG}} = \mathbb{E}_{\tau \sim \pi_\theta} [- \sum_{k,t} \text{sg}(A_{k,t}) \log p_{k,t}(\hat{y}_{k,t}) ]$, where $\text{sg}(\cdot)$ denotes the stop-gradient operator. In contrast, Generalized Knowledge Distillation (GKD) style OPD does not rely on sampled scalar advantages. Instead, it explicitly minimizes the statistical divergence between the full predictive distributions: $\mathcal{L}_{\text{GKD}} = \mathbb{E}_{\tau \sim \pi_\theta} [ \sum_{k,t} D_{\text{KL}}(p_{k,t} \parallel q_{k,t}) ]$. Consequently, while PG-style indirectly optimizes the policy by increasing the likelihood of trajectories with lower reverse KL, GKD-style provides explicit supervision to directly reshape the token-level distributions. Crucially, GKD-style ignores the global trajectory-sampling probability. Instead, it leverages direct gradient propagation to minimize the local divergence between the student and teacher distributions exclusively on the already visited prefixes. In practice, Sampled-Token OPD predominantly adopts the PG and GKD paradigm, whereas Top-$k$ and Full-Vocabulary OPD generally employ the GKD approach.

\paragraph{Sampled-Token OPD.}
Sampled-Token OPD evaluates the objective exclusively on the discrete tokens $\hat{y}_{k,t}$ sampled during the student's rollout, thereby avoiding the computation of the full vocabulary distribution. This approach is typically implemented in two ways. The \textbf{PG-Style} formulation utilizes a policy gradient estimator, treating the log-ratio as a token-level reward: $r_{k,t} = \log q_{k,t}(\hat{y}_{k,t}) - \log p_{k,t}(\hat{y}_{k,t})$. Alternatively, the \textbf{GKD-Style} formulation directly minimizes a local distance metric on the sampled tokens. Two common mathematical forms for the GKD-Style objective are based on Schulman's approximations: the \textbf{K1 approximation}, formulated as $\log p_{k,t}(\hat{y}_{k,t}) - \log q_{k,t}(\hat{y}_{k,t})$, and the \textbf{K2 approximation}, formulated as $\frac{1}{2}(\log p_{k,t}(\hat{y}_{k,t}) - \log q_{k,t}(\hat{y}_{k,t}))^2$.

\paragraph{Top-k OPD.}
Top-$k$ OPD limits the divergence penalty to a truncated vocabulary space. Specifically, at each decoding step $t$ within turn $k$, we isolate the subset of $k$ tokens that are most probable according to the student policy $p_{k,t}$. Let us denote this high-confidence token subset as $\mathcal{V}_{k,t} \subset \mathcal{V}$. We then construct truncated and re-normalized probability mass functions for both the student and the teacher, restricted strictly to $\mathcal{V}_{k,t}$:
\begin{equation}
\tilde{p}_{k,t}(v) = \frac{p_{k,t}(v)}{\sum_{w \in \mathcal{V}_{k,t}} p_{k,t}(w)}, \quad \tilde{q}_{k,t}(v) = \frac{q_{k,t}(v)}{\sum_{w \in \mathcal{V}_{k,t}} q_{k,t}(w)}, \quad \forall v \in \mathcal{V}_{k,t}.
\end{equation}

The distillation objective is subsequently formulated by evaluating the local KL divergence exclusively over these adjusted distributions. Integrating this into our agent's multi-turn trajectory framework yields the following formulation:
\begin{equation}
\mathcal{L}_{\text{Top-}k}(\theta) = \mathbb{E}_{\tau \sim \pi_\theta} \left[ \sum_{k=1}^{K} \sum_{t=1}^{T_k} D_{\text{KL}}(\tilde{p}_{k,t} \parallel \tilde{q}_{k,t}) \right].
\end{equation}

\paragraph{Full-Vocabulary OPD.}
In contrast to sampled or truncated approximations, Full-Vocabulary OPD evaluates the divergence penalty across the entire vocabulary $\mathcal{V}$ without any masking. At each decoding step $t$ within turn $k$, it directly aligns the complete predictive distribution of the student with that of the teacher. The distillation objective function computes the exact KL divergence over all possible tokens:
\begin{equation}
\mathcal{L}_{\text{Full-Vocab}}(\theta) = \mathbb{E}_{\tau \sim \pi_\theta} \left[ \sum_{k=1}^{K} \sum_{t=1}^{T_k} D_{\text{KL}}(p_{k,t} \parallel q_{k,t}) \right].
\end{equation}
\section{Task Formulation}
\label{sec:task}

\subsection{Controllable Planning Gym}

\begin{figure}[htbp]
    \centering
    \includegraphics[width=\textwidth]{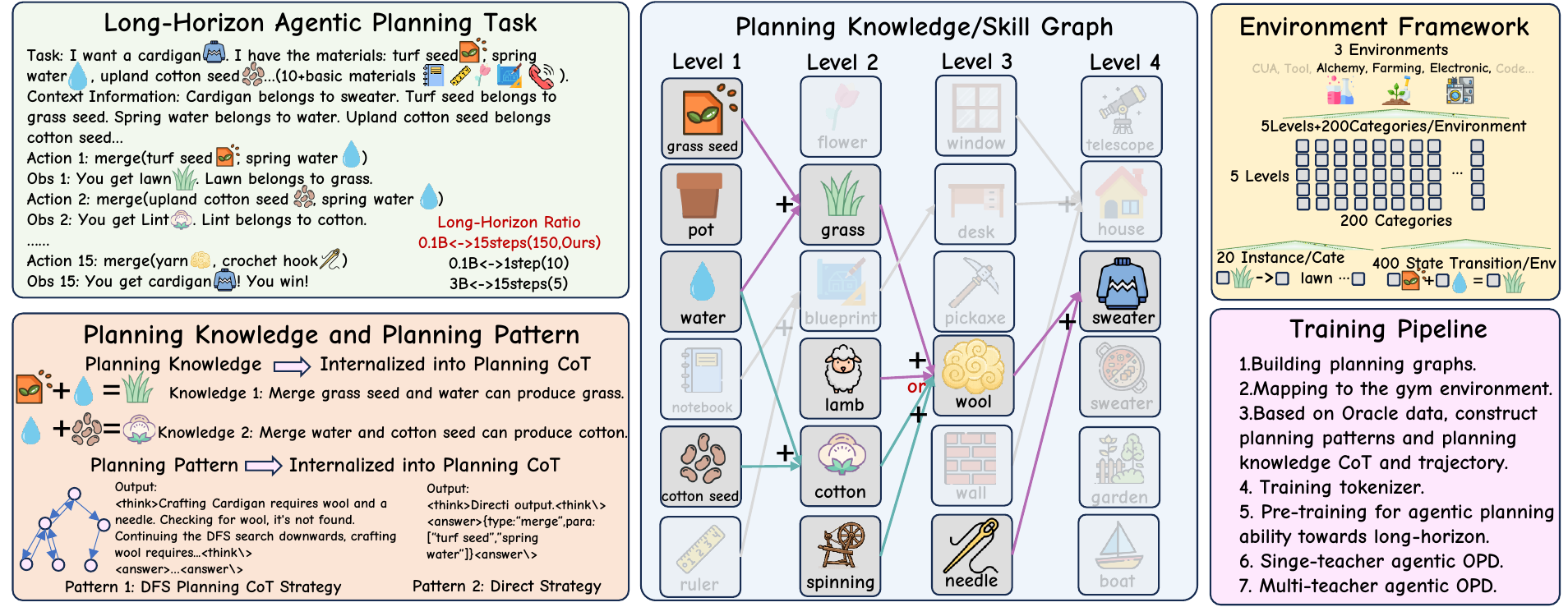}
    \caption{Overview of the proposed long-horizon agentic planning framework.}
    \label{fig:task} 
    \vspace{-10pt}
\end{figure}

\paragraph{Skill Graph Construction.}
As shown in Figure~\ref{fig:task}, we construct hierarchical skill graphs across three distinct domains (e.g., \textit{Fantasy Alchemy, Livestock Farming, Electronic Assembly}). To ensure rich semantic relationships, we leverage a large language model to generate category trees and concrete item entities. Each domain is structured as a tree with a height of $H$ layers, where each layer encompasses $W$ categories, and each category contains $N$ unique concrete items. 

The abstract graph transitions systematically connect nodes from lower layers to higher layers. \textbf{Crucially, the synthesis rules governing these transitions are formulated as a logical combination of AND/OR operations over prerequisite items.} For a target node $v$, its valid synthesis pathways are defined by an \textbf{OR} relation over multiple independent recipes, while each individual recipe requires the simultaneous availability (\textbf{AND}) of its constituent ingredients. Formally, the synthesis logic for node $v$ is expressed as:
\begin{equation}
v \iff (u_{1,1} \land \dots \land u_{1,k_1}) \lor \dots \lor (u_{m,1} \land \dots \land u_{m,k_m})
\end{equation}
where each conjunction block $(u_{i,1} \land \dots \land u_{i,k_i})$ represents a valid abstract recipe. To prevent high-step-count explosion, we employ a greedy minimization algorithm that prioritizes combining items with fewer preceding synthesis steps. For each domain, we generate disjoint abstract transition graphs (e.g., Graph A and Graph B). The transitions require varying numbers of prerequisite nodes governed by a fixed probability $p$, ensuring structurally sound and non-overlapping rule sets. Furthermore, to improve scalability, counterfactual synthesis is used for the state transition rules, since this is done from the pre-training stage, it does not affect the experimental conclusions.

\paragraph{Graph-to-Gym Mapping.}
The abstract skill graphs are instantiated into concrete simulated environments or gyms by mapping abstract nodes to specific items via an \textit{index map}. To foster environment diversity, we generate a combination of fixed and randomized configurations. Crucially, a subset of $K$ item indices combination per category is reserved for test set configurations. The remaining indices are randomly sampled to construct diverse training environments, preventing data leakage and ensuring evaluation on unseen entity mappings.

\paragraph{Task Difficulty Control.}
Task difficulty is determined by the minimum number of actions (i.e., synthesis steps) required to produce a target item from a given initial inventory. By sampling the target item from higher layers and starting materials from lower layers, we dynamically compute the exact step count required. Based on this, tasks are categorized into three difficulty levels: \textit{Short}, \textit{Middle}, and \textit{Long} step thresholds. To simulate realistic planning noise, we inject distractor items from the same starting layers into the initial inventory.

\paragraph{Dataset Splitting.}
To evaluate generalization, we split the instantiated configurations into datasets for pre-training, post-training, and testing. The test set exclusively employs novel combinations of familiar items. The training set is divided: the majority of configurations are utilized for pre-training to encourage exploratory learning across all trajectories, while a smaller subset is reserved for post-training, providing high-quality trajectories.

\subsection{Task Setup}
We formulate the planning problem as a sequential decision-making task. The agent is initialized with a starting inventory (prerequisites and distractors) and must synthesize a target item in limited steps. The state observation includes natural language descriptions of the category each item belongs to, facilitating category-level reasoning without exposing underlying IDs. At each step, the model must select the correct sequence of materials from the inventory to execute valid synthesis actions until the target item is successfully created.

\subsection{Evaluation}
For evaluation, we uniformly sample $M$ tasks across different step counts to ensure a balanced assessment spanning the difficulty spectrum. The evaluation is conducted on the reserved test configurations of Graph A for each difficulty level (\textit{Short}, \textit{Middle}, and \textit{Long}). The primary evaluation metric is the \textit{final success rate}, defined as the agent's ability to successfully synthesize the target item by strictly adhering to the unseen rules. For each test task, we independently run the agent $K$ times in a multi-turn setting. 
\textit{Avg@$K$} is computed by averaging \textit{Pass@1} over the $K$ independent runs. 
\textit{Pass@$K$} is the fraction of tasks solved in at least one of the $K$ runs.

\section{Multi-Turn Long-Horizon Planning Ability Acquisition during Pre-training}
\label{sec:pretrain}

Current large language model agents remain weak at multi-turn long-horizon planning. We investigate this problem by progressively relaxing three assumptions about pre-training data.
(1) \textbf{Data format: world-model internalization.} Given correct expert trajectories, does explicitly modeling intermediate state transitions via CoT to improve planning beyond only action-sequence imitation?
(2) \textbf{Data distribution: atomic compositional generalization.} When long-horizon trajectories are long-tail distribution, can models compose abundant atomic skills to generalize to compositional long-horizon planning tasks?
(3) \textbf{Data quality: suboptimal trajectories.} When expert trajectories are mixed with redundant, suboptimal, or erroneous solutions, how does this heterogeneity affect the long-horizon planning ability?
Throughout this section, pre-training data includes both pre-training and mid-training corpora, which are combined into a unified training mixture.
Accordingly, this section studies \textbf{Planning with Internalized World Model}, \textbf{Atomic Skill Compositional Generalization}, and \textbf{Impact of Suboptimal Trajectories}.

\subsection{Planning with Internalized World Model}
\paragraph{World Modeling.}
Fundamentally, a world model represents the underlying dynamics of an environment, serving as an internal simulator of how the environment transitions between states over time. It captures the relationship between current states, actions, and future states. The world model is defined by a state transition function:
\begin{equation}
\mathcal{T}: \mathcal{S} \times \mathcal{A} \rightarrow \mathcal{S}
\end{equation}
where $\mathcal{S}$ denotes the state space and $\mathcal{A}$ denotes the action space. Given a state $s_t \in \mathcal{S}$ and an action $a_t \in \mathcal{A}$, the transition function predicts the next state $s_{t+1} = \mathcal{T}(s_t, a_t)$. This predictive capability allows an agent to simulate the future consequences of its decisions through internal reasoning, without interacting with the external environment.
\paragraph{Skill Internalized World Modeling.}
We model planning as depth-first search over an internal simulated state space
$\mathcal{S}$. Let $\mathcal{K}$ denote the set of atomic skills, where each
skill $k \in \mathcal{K}$ is defined as a state transition function
$f_k : \mathcal{S} \rightarrow \mathcal{S}$. We define a planning step $t$ as one internal DFS node expansion. At each step,
the model selects an ordered skill sequence
$\mathbf{k}_t = (k_{t,1}, \ldots, k_{t,m_t}) \in \mathcal{K}^{*}$ and applies
the corresponding transition operators sequentially on the internal state:
\begin{equation}
s_{t+1} = f_{k_{t,m_t}} \circ \cdots \circ f_{k_{t,1}} (s_t),
\end{equation}
which represents the cumulative effect of executing the selected skills within
the internal world model.

The resulting transitions define a search tree over simulated states, and the
model performs depth-first search by recursively expanding one branch and exploring alternative skill sequences until a
valid plan is found. All transitions and search operations are conducted within
the internal CoT rather than through external
environment interaction.

\begin{figure}[h]
    \centering
    \includegraphics[width=\textwidth]{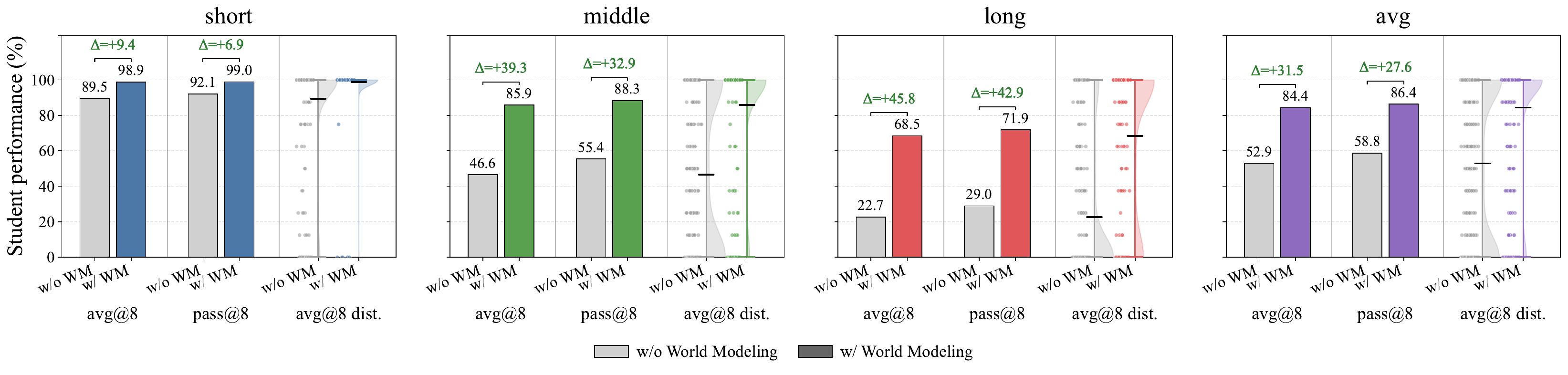}
    \captionsetup{skip=2pt}
    \caption{Final performance comparison between students with and without world modeling across three difficulty levels. For 3 domain and 3 difficulty level, each test subset contains 160 instances, resulting in 1,440 test instances for each checkpoint. During inference, we set the sampling temperature to 0.4 and independently sample 8 trajectories for each instance, with a limitation of 20 steps per trajectory. Therefore, evaluating one checkpoint requires a most total of 11,520 inference runs. We report $\mathrm{avg@8}$ as the average success rate across the 8 sampled trajectories, and $\mathrm{pass@8}$ as the proportion of instances for which at least one of the 8 trajectories successfully completes the task. Unless otherwise specified, we use the same evaluation protocol in all subsequent experiments.
}
    \label{fig:scaling_pretrain_bar} 
    \vspace{-10pt}
\end{figure}

\paragraph{Task Setup.}
(1) \textbf{Model configuration.} Since the pre-training corpora of existing models are proprietary and inaccessible, we instead construct a controlled experimental setting by randomly initializing a model under the Qwen2.5 architecture. To ensure consistent tokenization in the synthetic pre-training corpus, we train a byte-level BPE tokenizer from scratch. The full model configuration is detailed in Appendix~\ref{App:Pre_trained_Model_Configuration}.
(2) \textbf{Agent planning task configuration.} The corpus covers 3 domains. Each domain contains 5 levels, with 40 item categories per level and 20 instances per category.  These item categories have composition rules, meaning an item can be made from other items using AND and OR logic. The task requires the model to create a target item from basic materials, while extra noise items are added to the mix. You can find the details in Appendix~\ref{App:Pre_training_Schema_Configuration}.
(3) \textbf{Agent planning corpus configuration.} We compare the performance of direct answering and world modeling planning. For direct answering, the model predicts the action for each step directly. For world modeling planning, the model must first output a state transition function grounded in planning knowledge before acting. Specifically, it builds a chain of thought using post order traversal, state transition rules for item composition, and grounding from categories to exact instances. We train 1.2B corpus for the student model. You can find the exact corpus setup in Appendix~\ref{App:Pre_training_Corpus_Configuration}. And training details you can find in Appendix~\ref{App:Pre_training_Configuration}.

\begin{figure}[h]
    \centering
    \includegraphics[width=\textwidth]{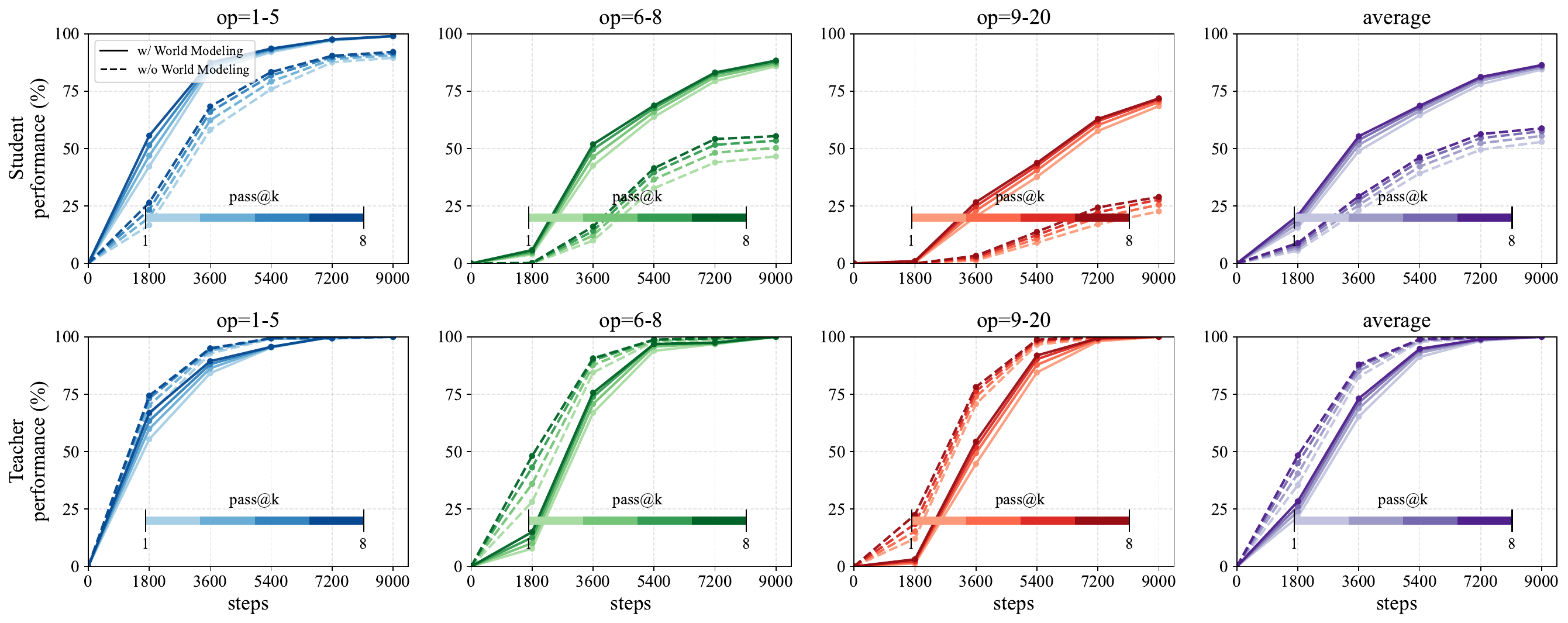}
    \captionsetup{skip=2pt}
    \caption{Performance scaling with pre-training corpus size for agent planning, comparing students with and without world modeling across three difficulty levels. The teacher is demonstrating the performance of the post-training set with the same distribution.}
    \label{fig:scaling_pretrain_passk} 
    \vspace{-10pt}
\end{figure}

\paragraph{Results.}
(1) \textbf{Internalizing a world model leads to stronger generalization ability.} As shown in Figure~\ref{fig:scaling_pretrain_bar}, the student model equipped with a world model (w/ WM) consistently outperforms the direct answering baseline (w/o WM) across all difficulty levels. Specifically, the model shows improvements in the short-horizon setting of +9.4\% (from 89.5\% to 98.9\%) and +6.9\% (from 92.1\% to 99.0\%) for avg@8 and pass@8, respectively. Similar enhancements are observed in the middle ($\Delta=+39.3\%$ for avg@8, $\Delta=+32.9\%$ for pass@8) and high ($\Delta=+45.8\%$ for avg@8, $\Delta=+42.9\%$ for pass@8) long-horizon planning, proving that planning with an internal world model improves the agent's ability to generalize to new and complex planning tasks.
(2) \textbf{Direct answering achieves higher efficiency with a lower token budget, but suffers from a lower performance upper bound.} As shown in Figure~\ref{fig:scaling_pretrain_passk}, analyzing the performance scaling curves over training steps reveals a distinct trade-off. In terms of learning speed, direct prediction (w/o WM, dashed lines) exhibits a faster initial trajectory in the teacher performance curves, where the dashed lines rise more rapidly and reach high performance at earlier stages (e.g., between 1800 and 5400 steps) compared to the solid lines (w/ WM). However, direct answering is limited by a strictly lower performance ceiling. As demonstrated by the student performance curves across all operation ranges (op=1-5, op=6-8, op=9-20) and their average, the solid lines (w/ WM) eventually surpass the dashed lines and stabilize at a significantly higher final accuracy. This demonstrates that using direct answering leads to faster convergence in training memory capabilities and requires a lower token budget, but it has much poorer generalization ability.

\begin{guidancebox}{Takeaway}
\textbf{Internalized world models enhance multi-turn long-horizon planning generalization.} While direct answering converges faster, requires a lower token budget, and is more efficient for tasks relying on memorized answers, it generalizes poorly to multi-turn long-horizon planning. In contrast, planning with a world model achieves stronger generalization and higher final accuracy on multi-turn long-horizon planning tasks.
\end{guidancebox}

\subsection{Atomic Skill Compositional Generalization}

\begin{figure}[h]
    \centering
    \includegraphics[width=\textwidth]{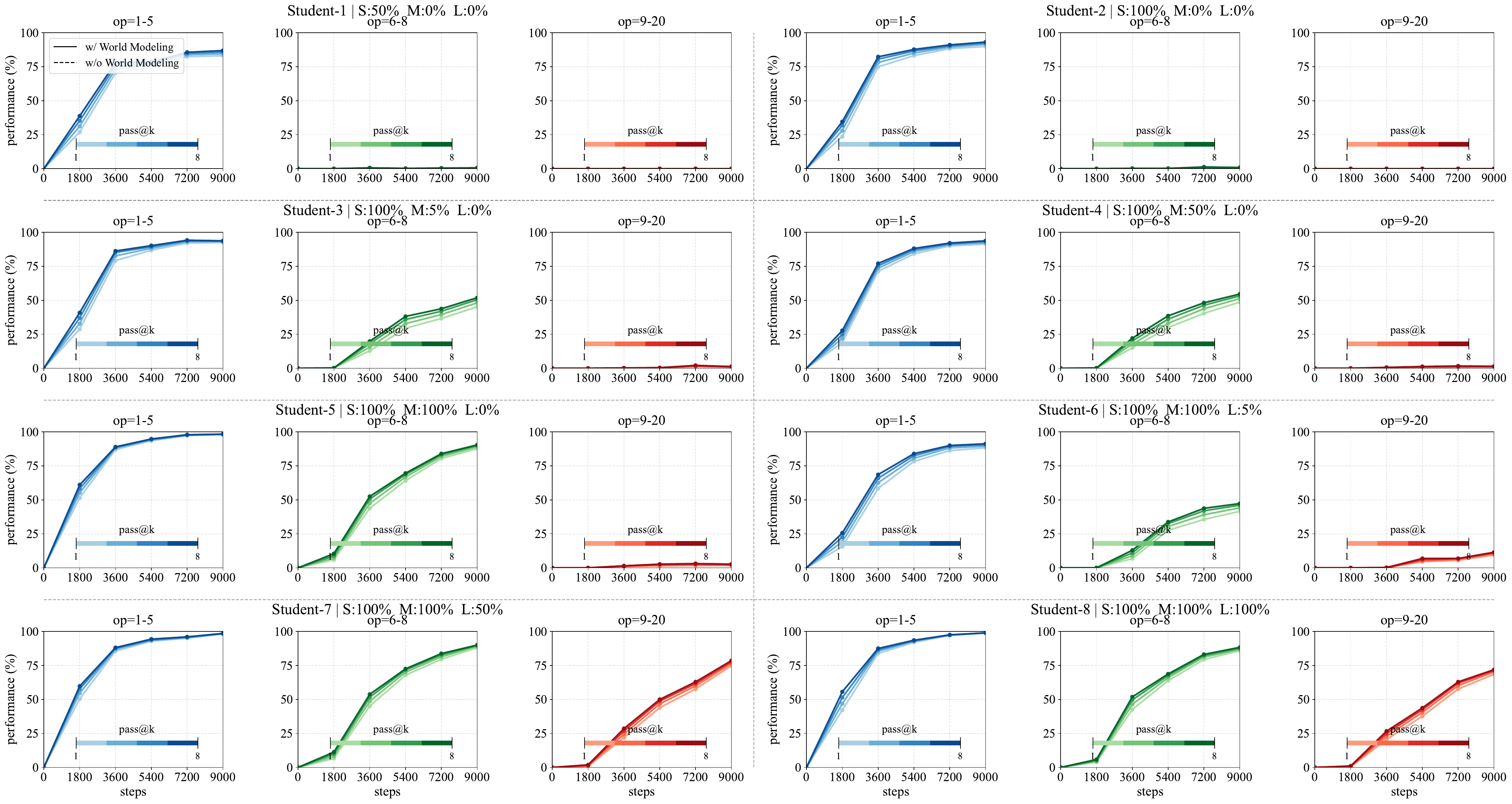}
    \caption{Performance scaling across different pretraining distributions.}
    \label{fig:Pre-Distribution} 
\end{figure}

\paragraph{Atomic Skill Composition.}
Atomic skill compositional generalization refers to the ability of an agent to combine known
\textbf{atomic skills} into new \textbf{compositional skills} for solving unseen
complex tasks~\citep{yuan2025f}. In this work, such compositional ability is acquired through
pre-training and encoded in the model parameters, rather than elicited through
test-time demonstrations, prompt engineering, or in-context learning. In our environment, an atomic skill is the ability to complete a single valid
synthesis step. We represent such a step as a basic function $f(x)$ that produces
a new item from a set of input materials. Suppose the model learns two recipes,
$f$ and $g$, from the training corpus. Compositional generalization occurs when,
without having observed their complete composition during training, the model can
correctly execute the novel nested pathway
\begin{equation}
h(x)=g(f(x))
\end{equation}
at test time. In our hierarchical graph tasks, this requires the model to
synthesize a target node at a higher level by chaining independently learned
lower-level recipes. This setting therefore evaluates whether pre-training
enables the model to internalize reusable synthesis rules and systematically
compose them, rather than merely memorizing complete trajectories or relying on
test-time prompting.

\definecolor{plotblue}{RGB}{106,178,214}
\definecolor{plotgreen}{RGB}{118,190,150}
\definecolor{plotteal}{RGB}{108,190,178}
\definecolor{plotcoral}{RGB}{236,118,126}
\definecolor{plotcream}{RGB}{248,241,220}
\definecolor{plotgray}{RGB}{235,235,235}
\begin{table}[h]
\centering
\footnotesize
\setlength{\tabcolsep}{2.2pt}
\caption{Pretrain distribution table with two global blocks (top: avg@8, bottom: pass@8). Ratio is split into S(Short)/M(Middle)/L(Long) columns. Each difficulty includes domain columns and a mean column. Two extra columns report difficulty gaps: Gap(L-S) and Gap(M-S). Domain abbreviations: FA = Fantasy Alchemy; LF = Livestock Farming; EA = Electronic Assembly.}
\label{tab:pretrain_distribution_region_ratio3_gap2}
\resizebox{\columnwidth}{!}{
\begin{tabular}{lccccccccccccccccc}
\toprule
\multirow{2}{*}{Region} & \multicolumn{3}{c}{Ratio} & \multicolumn{4}{c}{Short} & \multicolumn{4}{c}{Middle} & \multicolumn{4}{c}{Long} & \multicolumn{2}{c}{Gap} \\
\cmidrule(lr){2-4} \cmidrule(lr){5-8} \cmidrule(lr){9-12} \cmidrule(lr){13-16} \cmidrule(lr){17-18}
 & S & M & L & FA & LF & EA & Mean & FA & LF & EA & Mean & FA & LF & EA & Mean & L-S & M-S \\
\midrule
\rowcolor{plotgray}\multicolumn{18}{c}{\textbf{avg@8}} \\
\midrule
\multirow{2}{*}{S} & 50\% & 0\% & 0\% & \cellcolor{plotblue!44}90.62 & \cellcolor{plotblue!37}74.84 & \cellcolor{plotblue!41}83.75 & \cellcolor{plotblue!41}83.07 & \cellcolor{plotblue!6}0.23 & \cellcolor{plotblue!6}0.00 & \cellcolor{plotblue!6}0.55 & \cellcolor{plotblue!6}0.26 & \cellcolor{plotblue!6}0.00 & \cellcolor{plotblue!6}0.00 & \cellcolor{plotblue!6}0.00 & \cellcolor{plotblue!6}0.00 & \cellcolor{plotcoral!41}-83.07 & \cellcolor{plotcoral!41}-82.81 \\
 & 100\% & 0\% & 0\% & \cellcolor{plotblue!46}95.55 & \cellcolor{plotblue!40}80.70 & \cellcolor{plotblue!45}93.28 & \cellcolor{plotblue!44}89.84 & \cellcolor{plotblue!6}0.62 & \cellcolor{plotblue!6}0.16 & \cellcolor{plotblue!6}0.39 & \cellcolor{plotblue!6}0.39 & \cellcolor{plotblue!6}0.00 & \cellcolor{plotblue!6}0.00 & \cellcolor{plotblue!6}0.00 & \cellcolor{plotblue!6}0.00 & \cellcolor{plotcoral!44}-89.84 & \cellcolor{plotcoral!44}-89.45 \\
\cmidrule(lr){1-18}
\multirow{3}{*}{S+M} & 100\% & 5\% & 0\% & \cellcolor{plotblue!47}96.88 & \cellcolor{plotblue!42}84.53 & \cellcolor{plotblue!46}96.09 & \cellcolor{plotblue!45}92.50 & \cellcolor{plotblue!30}56.95 & \cellcolor{plotblue!22}37.19 & \cellcolor{plotblue!23}41.25 & \cellcolor{plotblue!25}45.13 & \cellcolor{plotblue!6}0.70 & \cellcolor{plotblue!6}0.62 & \cellcolor{plotblue!6}1.02 & \cellcolor{plotblue!6}0.78 & \cellcolor{plotcoral!45}-91.72 & \cellcolor{plotcoral!27}-47.37 \\
 & 100\% & 50\% & 0\% & \cellcolor{plotblue!46}94.14 & \cellcolor{plotblue!43}88.83 & \cellcolor{plotblue!44}90.86 & \cellcolor{plotblue!44}91.28 & \cellcolor{plotblue!31}59.69 & \cellcolor{plotblue!22}38.52 & \cellcolor{plotblue!26}47.73 & \cellcolor{plotblue!26}48.65 & \cellcolor{plotblue!6}0.55 & \cellcolor{plotblue!6}0.00 & \cellcolor{plotblue!7}1.95 & \cellcolor{plotblue!6}0.83 & \cellcolor{plotcoral!44}-90.44 & \cellcolor{plotcoral!25}-42.63 \\
 & 100\% & 100\% & 0\% & \cellcolor{plotblue!48}99.77 & \cellcolor{plotblue!47}97.42 & \cellcolor{plotblue!47}96.72 & \cellcolor{plotblue!47}97.97 & \cellcolor{plotblue!44}89.38 & \cellcolor{plotblue!43}87.50 & \cellcolor{plotblue!42}86.09 & \cellcolor{plotblue!43}87.66 & \cellcolor{plotblue!6}0.70 & \cellcolor{plotblue!7}1.72 & \cellcolor{plotblue!7}2.81 & \cellcolor{plotblue!7}1.74 & \cellcolor{plotcoral!46}-96.22 & \cellcolor{plotcoral!12}-10.31 \\
\cmidrule(lr){1-18}
\multirow{3}{*}{S+M+L} & 100\% & 100\% & 5\% & \cellcolor{plotblue!45}93.52 & \cellcolor{plotblue!40}80.62 & \cellcolor{plotblue!44}90.62 & \cellcolor{plotblue!43}88.26 & \cellcolor{plotblue!31}58.98 & \cellcolor{plotblue!19}31.02 & \cellcolor{plotblue!21}34.77 & \cellcolor{plotblue!23}41.59 & \cellcolor{plotblue!12}13.52 & \cellcolor{plotblue!7}3.20 & \cellcolor{plotblue!11}11.72 & \cellcolor{plotblue!10}9.48 & \cellcolor{plotcoral!40}-78.78 & \cellcolor{plotcoral!27}-46.67 \\
 & 100\% & 100\% & 50\% & \cellcolor{plotblue!48}99.92 & \cellcolor{plotblue!47}96.56 & \cellcolor{plotblue!47}98.12 & \cellcolor{plotblue!47}98.20 & \cellcolor{plotblue!43}88.12 & \cellcolor{plotblue!42}85.55 & \cellcolor{plotblue!44}90.55 & \cellcolor{plotblue!43}88.07 & \cellcolor{plotblue!39}78.05 & \cellcolor{plotblue!36}72.50 & \cellcolor{plotblue!37}73.98 & \cellcolor{plotblue!37}74.84 & \cellcolor{plotcoral!17}-23.36 & \cellcolor{plotcoral!12}-10.13 \\
 & 100\% & 100\% & 100\% & \cellcolor{plotblue!48}100.00 & \cellcolor{plotblue!48}99.22 & \cellcolor{plotblue!47}97.50 & \cellcolor{plotblue!48}98.91 & \cellcolor{plotblue!45}91.80 & \cellcolor{plotblue!41}82.19 & \cellcolor{plotblue!41}83.59 & \cellcolor{plotblue!42}85.86 & \cellcolor{plotblue!35}69.61 & \cellcolor{plotblue!34}65.55 & \cellcolor{plotblue!36}70.39 & \cellcolor{plotblue!35}68.52 & \cellcolor{plotcoral!20}-30.39 & \cellcolor{plotcoral!13}-13.05 \\
\midrule
\rowcolor{plotgray}\multicolumn{18}{c}{\textbf{pass@8}} \\
\midrule
\multirow{2}{*}{S} & 50\% & 0\% & 0\% & \cellcolor{plotgreen!45}92.50 & \cellcolor{plotgreen!40}81.25 & \cellcolor{plotgreen!42}86.88 & \cellcolor{plotgreen!42}86.88 & \cellcolor{plotgreen!7}1.25 & \cellcolor{plotgreen!6}0.00 & \cellcolor{plotgreen!6}0.62 & \cellcolor{plotgreen!6}0.62 & \cellcolor{plotgreen!6}0.00 & \cellcolor{plotgreen!6}0.00 & \cellcolor{plotgreen!6}0.00 & \cellcolor{plotgreen!6}0.00 & \cellcolor{plotcoral!43}-86.88 & \cellcolor{plotcoral!42}-86.25 \\
 & 100\% & 0\% & 0\% & \cellcolor{plotgreen!47}97.50 & \cellcolor{plotgreen!42}86.25 & \cellcolor{plotgreen!46}95.62 & \cellcolor{plotgreen!45}93.12 & \cellcolor{plotgreen!6}0.62 & \cellcolor{plotgreen!6}0.62 & \cellcolor{plotgreen!7}1.25 & \cellcolor{plotgreen!6}0.83 & \cellcolor{plotgreen!6}0.00 & \cellcolor{plotgreen!6}0.00 & \cellcolor{plotgreen!6}0.00 & \cellcolor{plotgreen!6}0.00 & \cellcolor{plotcoral!45}-93.12 & \cellcolor{plotcoral!45}-92.29 \\
\cmidrule(lr){1-18}
\multirow{3}{*}{S+M} & 100\% & 5\% & 0\% & \cellcolor{plotgreen!47}98.12 & \cellcolor{plotgreen!42}86.25 & \cellcolor{plotgreen!47}96.88 & \cellcolor{plotgreen!45}93.75 & \cellcolor{plotgreen!32}61.88 & \cellcolor{plotgreen!24}43.12 & \cellcolor{plotgreen!27}50.62 & \cellcolor{plotgreen!28}51.88 & \cellcolor{plotgreen!7}1.25 & \cellcolor{plotgreen!6}0.62 & \cellcolor{plotgreen!7}1.88 & \cellcolor{plotgreen!7}1.25 & \cellcolor{plotcoral!45}-92.50 & \cellcolor{plotcoral!25}-41.88 \\
 & 100\% & 50\% & 0\% & \cellcolor{plotgreen!46}95.00 & \cellcolor{plotgreen!45}93.12 & \cellcolor{plotgreen!45}93.12 & \cellcolor{plotgreen!45}93.75 & \cellcolor{plotgreen!33}63.12 & \cellcolor{plotgreen!25}45.00 & \cellcolor{plotgreen!29}55.62 & \cellcolor{plotgreen!29}54.58 & \cellcolor{plotgreen!6}0.62 & \cellcolor{plotgreen!6}0.00 & \cellcolor{plotgreen!8}3.75 & \cellcolor{plotgreen!7}1.46 & \cellcolor{plotcoral!45}-92.29 & \cellcolor{plotcoral!24}-39.17 \\
 & 100\% & 100\% & 0\% & \cellcolor{plotgreen!48}100.00 & \cellcolor{plotgreen!47}98.12 & \cellcolor{plotgreen!47}96.88 & \cellcolor{plotgreen!47}98.33 & \cellcolor{plotgreen!44}91.25 & \cellcolor{plotgreen!44}90.00 & \cellcolor{plotgreen!44}90.00 & \cellcolor{plotgreen!44}90.42 & \cellcolor{plotgreen!7}1.25 & \cellcolor{plotgreen!7}2.50 & \cellcolor{plotgreen!8}4.38 & \cellcolor{plotgreen!7}2.71 & \cellcolor{plotcoral!46}-95.62 & \cellcolor{plotcoral!11}-7.92 \\
\cmidrule(lr){1-18}
\multirow{3}{*}{S+M+L} & 100\% & 100\% & 5\% & \cellcolor{plotgreen!46}95.62 & \cellcolor{plotgreen!42}85.00 & \cellcolor{plotgreen!45}93.12 & \cellcolor{plotgreen!44}91.25 & \cellcolor{plotgreen!33}63.12 & \cellcolor{plotgreen!21}35.00 & \cellcolor{plotgreen!24}43.75 & \cellcolor{plotgreen!26}47.29 & \cellcolor{plotgreen!13}16.88 & \cellcolor{plotgreen!8}3.75 & \cellcolor{plotgreen!12}13.75 & \cellcolor{plotgreen!11}11.46 & \cellcolor{plotcoral!40}-79.79 & \cellcolor{plotcoral!26}-43.96 \\
 & 100\% & 100\% & 50\% & \cellcolor{plotgreen!48}100.00 & \cellcolor{plotgreen!47}97.50 & \cellcolor{plotgreen!47}98.12 & \cellcolor{plotgreen!47}98.54 & \cellcolor{plotgreen!44}89.38 & \cellcolor{plotgreen!43}88.75 & \cellcolor{plotgreen!45}91.88 & \cellcolor{plotgreen!44}90.00 & \cellcolor{plotgreen!40}81.88 & \cellcolor{plotgreen!38}76.25 & \cellcolor{plotgreen!39}77.50 & \cellcolor{plotgreen!39}78.54 & \cellcolor{plotcoral!16}-20.00 & \cellcolor{plotcoral!11}-8.54 \\
 & 100\% & 100\% & 100\% & \cellcolor{plotgreen!48}100.00 & \cellcolor{plotgreen!48}99.38 & \cellcolor{plotgreen!47}97.50 & \cellcolor{plotgreen!48}98.96 & \cellcolor{plotgreen!45}93.75 & \cellcolor{plotgreen!42}85.00 & \cellcolor{plotgreen!42}86.25 & \cellcolor{plotgreen!43}88.33 & \cellcolor{plotgreen!36}71.88 & \cellcolor{plotgreen!36}71.25 & \cellcolor{plotgreen!36}72.50 & \cellcolor{plotgreen!36}71.88 & \cellcolor{plotcoral!19}-27.08 & \cellcolor{plotcoral!12}-10.62 \\
\bottomrule
\end{tabular}
}
\end{table}

\paragraph{Task Setup.} (1) \textbf{Difficulty definition.} We divide the tasks into short, middle, and long levels based on the number of operation steps. Specifically, the short-horizon requires 1 to 5 steps, the middle-horizon requires 6 to 8 steps, and the long-horizon requires 9 or more steps. (2) \textbf{Data allocation.} We organize the training data into three difficulty intervals. The long-horizon interval includes all data from the short-horizon intervals. The short-horizon intervals do not contain complete tasks from the higher level, but only partial task components. Within each interval, we set different data ratios including 0\%, 5\%, 50\%, and 100\%. (3) \textbf{Experiment and evaluation.} All experiments use a unified number of training steps for 9000 steps. We then evaluate the models across the three horizon levels to test the compositional generalization ability of their internalized skills.

\paragraph{Results.} As shown in Figure~\ref{fig:Pre-Distribution} and Table~\ref{tab:pretrain_distribution_region_ratio3_gap2}.
(1) \textbf{Atomic skills struggle to compose automatically.} When the model is pretrained exclusively on short tasks (Region S at 100\%), it achieves strong performance on short-horizon tasks (pass@8 of 93.12\%), but accuracy on longer horizons drops to near zero (0.83\% for middle and 0.00\% for long-horizon subsets). For multi-turn agents, simply knowing individual actions is insufficient. They fail to connect atomic skills into long-horizon planning ability without explicitly observing extended state transitions during training. 
(2) \textbf{Minimal exposure to long-horizon trajectories activate planning generalization.} Adding just a small fraction of long-horizon trajectories causes a sudden jump in performance on longer levels. For instance, introducing merely 5\% of middle-horizon trajectories boosts middle pass@8 from 0.83\% to 51.88\%, while 5\% of long-horizon trajectories increases long-horizon subset in pass@8 from 0.00\% to 11.46\%.

\begin{guidancebox}{Takeaway}
\textbf{Models struggle with automatic compositional generalization, but minimal long-horizon trajectories activates long-horizon planning.} Learning atomic skills exclusively from short tasks is insufficient for solving multi-turn long-horizon planning tasks. However, introducing just a small fraction of long-horizon trajectories teaches the agent how to connect these skills sequentially, bridging the performance gap on longer tasks.
\end{guidancebox}

\subsection{Impact of Suboptimal Trajectories}
\label{subsec:planning_patterns}

\paragraph{Planning Pattern.} 
We define the planning pattern as the structural and algorithmic skeleton of the CoT. It dictates the general thinking approach required to solve the task, independent of the specific planning knowledge involved. As demonstrated in Table \ref{tab:starquench_skill_graph_synthesis}, the planning pattern involves a series of steps, including identifying the target category, recalling relevant planning knowledge or skills, linking abstract categories with specific inventory items, and generating the final action. This template uses standardized textual structures with consistent phrasing to explicitly define the target objective, analyze the required prerequisites, and finally generate the corresponding action. However, pre-training corpora may also contain other planning patterns, such as direct-answer patterns that bypass explicit planning processes, or low-quality planning patterns that involve incomplete, noisy, or irrelevant planning information.

\paragraph{Planning Knowledge.} 
In contrast, we define planning knowledge as task-specific procedural knowledge that specifies how actions interact, combine, and transform to produce valid outcomes. It provides the concrete rules that instantiate abstract planning patterns into executable plans, such as synthesis recipes or state-action transitions. As illustrated in Table \ref{tab:starquench_skill_graph_synthesis}, the planning knowledge encompasses the explicit synthesis rules defining exactly which materials combine to create a new item. This refers to the factual recipes within the skill graph, such as the specific knowledge that combining Crystalgrain and Glowfern synthesizes a Quartz Solution. To support scalable dataset construction, we introduce counterfactual synthesis rules to expand the skill graph with systematically generated yet logically consistent synthesis knowledge. As these rules are consistently adopted throughout both pre-training and all downstream experiments, the experimental settings remain aligned, and our conclusions are not affected by the use of counterfactual synthesis.

\begin{figure}[h]
    \centering
    \includegraphics[width=\textwidth]{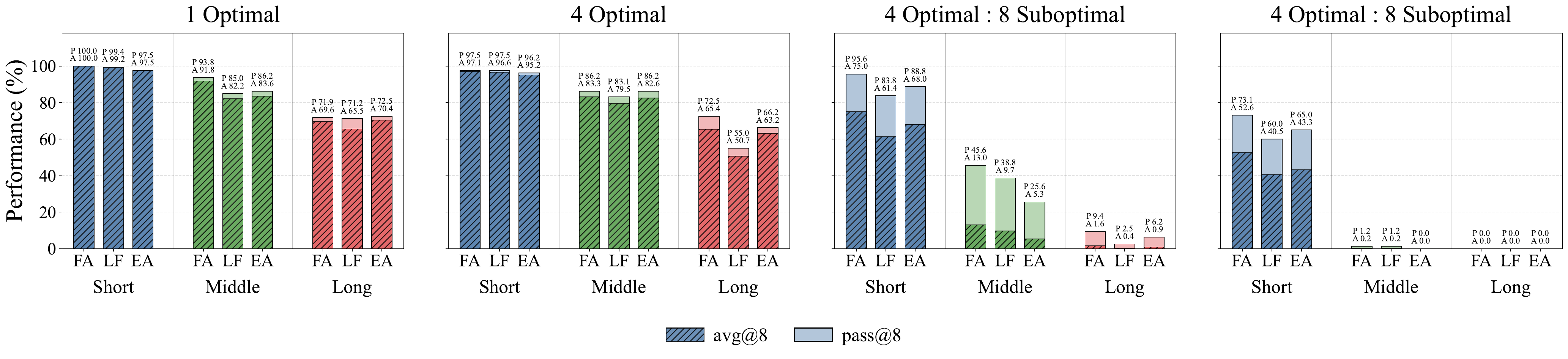}
    \caption{Effect of pretraining data composition on downstream performance. The figure compares avg@8 and pass@8 across three task domains and three difficulty levels under different pretraining configurations. FA, LF, and EA denote Fantasy Alchemy, Livestock Farming, and Electronic Assembly.}
    \label{fig:pretrain_pattern_bar} 
\end{figure}

\begin{table}[htbp]
\centering
\caption{Planning pattern and planning knowledge for the first fantasy alchemy test sample. The state transition and synthesis rules of the Skill Graph represent the underlying dynamics of the environment, and they must be learned through interaction with the environment, as they are not provided as part of the context during multi-turn interactions.}
\label{tab:starquench_skill_graph_synthesis}
\scriptsize
\renewcommand{\arraystretch}{0.85}
\setlength{\tabcolsep}{3pt}
\begin{tabularx}{\textwidth}{@{}l X@{}}
\toprule
\textbf{Part} & \textbf{JSON Content} \\
\midrule
\textbf{Input} &
<question>\newline
\{\q{}task\q{}: \q{}Synthesize \instTarget{starquench vial}\q{}, \newline
\q{}target\_category\q{}: \q{}\catTarget{Star Elixir}\q{}, \newline
\q{}inventory\q{}: [\q{}\instCrystal{Crystal Speck}\q{}, \q{}\instDragon{Dragon Scute}\q{}, \q{}\instFaerie{Faerie Dusting}\q{}, \q{}\instGlow{Glow Frond}\q{}, \q{}\instMoon{Lunar Sprig}\q{}, \q{}\instMithril{Mithril Lump}\q{}, \q{}\instObsidian{Obsidian Chip}\q{}, \q{}\instSapphire{Sapphire Grit}\q{}, \q{}\instSylph{Sylph Filament}\q{}], \newline
\q{}inventory\_category\_description\q{}: [\q{}\instCrystal{Crystal Speck} belongs to category \catCrystal{Crystalgrain}.\q{}, \q{}\instDragon{Dragon Scute} belongs to category \catDragon{Dragon Scale}.\q{}, \q{}\instFaerie{Faerie Dusting} belongs to category \catFaerie{Faerie Pollen}.\q{}, \q{}\instGlow{Glow Frond} belongs to category \catGlow{Glowfern}.\q{}, \q{}\instMoon{Lunar Sprig} belongs to category \catMoon{Moonleaf}.\q{}, \q{}\instMithril{Mithril Lump} belongs to category \catMithril{Mithril Nugget}.\q{}, \q{}\instObsidian{Obsidian Chip} belongs to category \catObsidian{Obsidian Shard}.\q{}, \q{}\instSapphire{Sapphire Grit} belongs to category \catSapphire{Sapphire Sand}.\q{}, \q{}\instSylph{Sylph Filament} belongs to category \catSylph{Sylph Thread}.\q{}], \newline
\q{}history\q{}: []\}\newline
</question>
\\
\midrule
\textbf{Skill Graph} &
\vspace{2mm}
\begin{tikzpicture}[
    xscale=1.0, yscale=0.72,
    activebox/.style={draw=black, text=black, thick, rectangle, rounded corners, align=center, inner sep=3pt, minimum width=2.65cm, fill=white},
    targetbox/.style={draw=starC, text=black, very thick, rectangle, rounded corners, align=center, inner sep=3pt, minimum width=2.65cm, fill=white},
    graybox/.style={draw=gray!75, text=gray!85, rectangle, rounded corners, align=center, inner sep=3pt, minimum width=2.65cm, fill=white}
]
    \node[activebox] (crystal)  at (0, 0)     {\instCrystal{Crystal Speck} \\ (\catCrystal{Crystalgrain})};
    \node[activebox] (glow)     at (0, -1.3)  {\instGlow{Glow Frond} \\ (\catGlow{Glowfern})};
    \node[activebox] (mithril)  at (0, -2.6)  {\instMithril{Mithril Lump} \\ (\catMithril{Mithril Nugget})};
    \node[activebox] (sylph)    at (0, -3.9)  {\instSylph{Sylph Filament} \\ (\catSylph{Sylph Thread})};
    \node[activebox] (dragon)   at (0, -5.2)  {\instDragon{Dragon Scute} \\ (\catDragon{Dragon Scale})};
    \node[activebox] (faerie)   at (0, -6.5)  {\instFaerie{Faerie Dusting} \\ (\catFaerie{Faerie Pollen})};
    \node[graybox]   (lunar)    at (0, -7.8)  {\instMoon{Lunar Sprig} \\ (\catMoon{Moonleaf})};
    \node[graybox]   (obsidian) at (0, -9.1)  {\instObsidian{Obsidian Chip} \\ (\catObsidian{Obsidian Shard})};
    \node[graybox]   (sapphire) at (0, -10.4) {\instSapphire{Sapphire Grit} \\ (\catSapphire{Sapphire Sand})};
    \node[activebox] (silica) at (4.3, -0.65) {\instQuartz{silica liquor} \\ (\catQuartz{Quartz Solution})};
    \node[activebox] (spring) at (4.3, -3.25) {\instNymph{spring jelly} \\ (\catNymph{Nymph Gel})};
    \node[activebox] (coal)   at (4.3, -5.85) {\instEmber{coalroot mash} \\ (\catEmber{Ember Paste})};
    \node[activebox] (ruby)    at (8.6, -1.95) {\instRuby{rubylace shard} \\ (\catRuby{Ruby Matrix})};
    \node[activebox] (unicorn) at (8.6, -5.85) {\instUnicorn{unicornlace node} \\ (\catUnicorn{Unicorn Matrix})};
    \node[targetbox] (target) at (12.9, -3.9) {\instTarget{starquench vial} \\ (\catTarget{Star Elixir})};
    \draw[->, thick, draw=quartzC] (crystal.east) to[out=0, in=180] (silica.west);
    \draw[->, thick, draw=quartzC] (glow.east)    to[out=0, in=180] (silica.west);
    \draw[->, thick, draw=nymphC] (mithril.east) to[out=0, in=180] (spring.west);
    \draw[->, thick, draw=nymphC] (sylph.east)   to[out=0, in=180] (spring.west);
    \draw[->, thick, draw=emberC] (dragon.east)  to[out=0, in=180] (coal.west);
    \draw[->, thick, draw=emberC] (faerie.east)  to[out=0, in=180] (coal.west);
    \draw[->, thick, draw=rubyC] (spring.east)  to[out=0, in=180] (ruby.west);
    \draw[->, thick, draw=rubyC] (silica.east)  to[out=0, in=180] (ruby.west);
    \draw[->, thick, draw=unicornC] (coal.east) to[out=0, in=180] (unicorn.west);
    \draw[->, very thick, draw=starC] (ruby.east)    to[out=0, in=180] (target.west);
    \draw[->, very thick, draw=starC] (unicorn.east) to[out=0, in=180] (target.west);
\end{tikzpicture}
\vspace{2mm}
\\
\midrule
\makecell[t]{\textbf{Pattern 1 Output}\\\textbf{(Turn 1)}} &
<solution>Goal: synthesize \catTarget{Star Elixir}. Category synthesis rule: combining \catRuby{Ruby Matrix}, \catUnicorn{Unicorn Matrix} can produce \catTarget{Star Elixir}. \catRuby{Ruby Matrix} is currently not represented in inventory. \catUnicorn{Unicorn Matrix} is currently not represented in inventory. Status: need to acquire \catRuby{Ruby Matrix}, \catUnicorn{Unicorn Matrix} first. Goal: synthesize \catRuby{Ruby Matrix}. Category synthesis rule: combining \catNymph{Nymph Gel}, \catQuartz{Quartz Solution} can produce \catRuby{Ruby Matrix}. \catNymph{Nymph Gel} is currently not represented in inventory. \catQuartz{Quartz Solution} is currently not represented in inventory. Status: need to acquire \catNymph{Nymph Gel}, \catQuartz{Quartz Solution} first. Goal: synthesize \catNymph{Nymph Gel}. Category synthesis rule: combining \catMithril{Mithril Nugget}, \catSylph{Sylph Thread} can produce \catNymph{Nymph Gel}. \catMithril{Mithril Nugget} is represented by items like \instMithril{Mithril Lump}. \catSylph{Sylph Thread} is represented by items like \instSylph{Sylph Filament}. Action: use \instMithril{Mithril Lump} (\catMithril{Mithril Nugget}), \instSylph{Sylph Filament} (\catSylph{Sylph Thread}) to fulfill this category requirement.</solution>\newline
<answer>\{\q{}type\q{}: \q{}process\q{}, \q{}materials\q{}: [\q{}\instMithril{Mithril Lump}\q{}, \q{}\instSylph{Sylph Filament}\q{}]\}</answer>
\\
\midrule
\makecell[t]{\textbf{Pattern 1 Obs}\\\textbf{(Turn 1)}} &
You get \instNymph{spring jelly}. It belongs to \catNymph{Nymph Gel}.\newline
......
\\
\midrule
\makecell[t]{\textbf{Pattern 1 Output}\\\textbf{(Turn 15)}} &
<solution>Goal: synthesize ... to fulfill this category requirement.</solution>\newline
<answer>\{\q{}type\q{}: \q{}process\q{}, \q{}materials\q{}: [\q{}\instRuby{rubylace shard}\q{}, \q{}\instUnicorn{unicornlace node}\q{}]\}</answer>
\\
\midrule
\makecell[t]{\textbf{Final Obs}} &
You get \instTarget{starquench vial}!
\\
\midrule
\makecell[t]{\textbf{Pattern 2 Output}\\\textbf{(Turn 1)}} &
<solution>Direct selection from current inventory.</solution>\newline
<answer>\{\q{}type\q{}: \q{}process\q{}, \q{}materials\q{}: [\q{}\instMoon{Lunar Sprig}\q{}, \q{}\instObsidian{Obsidian Chip}\q{}]\}</answer>
\\
\midrule
\makecell[t]{\textbf{Pattern 2 Obs}\\\textbf{(Turn 1)}} &
You get nothing.\newline
......
\\
\midrule
\makecell[t]{\textbf{Pattern 2 Output}\\\textbf{(Turn 20)}} &
<solution>Direct selection from current inventory.</solution>\newline
<answer>\{\q{}type\q{}: \q{}process\q{}, \q{}materials\q{}: [\q{}\instSapphire{Sapphire Grit}\q{}, \q{}\instDragon{Dragon Scute}\q{}]\}</answer>
\\
\midrule
\makecell[t]{\textbf{Final Obs}} &
The number of steps exceeds the maximum step budget.
\\
\bottomrule
\end{tabularx}
\end{table}

\paragraph{Task Setup.} (1) \textbf{Planning pattern.} We primarily establish 8 planning patterns, which can be classified into two categories. The first thinking pattern features complex templates and optimal decisions. The second thinking pattern features direct and suboptimal decisions. Each template corresponds directly to whether it represents an optimal decision. (2) \textbf{Planning quality distribution.} We configure 8 quality distributions. For example, these distributions include 4 optimal modes, 4 to 4 ratio between optimal and suboptimal modes, and 4 to 8 ratio between optimal and suboptimal modes. In addition, all data utilize the same metadata except for different rendering, while the data scale and training steps remain consistent.

\paragraph{Results.} As shown in Figure~\ref{fig:pretrain_pattern_bar}. 
(1) \textbf{Multiple optimal patterns maintain similar generalization.} Increasing the diversity of optimal reasoning templates from 1 Optimal to 4 Optimal does not significantly degrade or improve performance. The model maintains similar success rates across all three domains (FA, LF, EA). This indicates that agents can effectively internalize and utilize multiple high-quality structural skeletons without detrimental interference. 
(2) \textbf{Suboptimal trajectories cause catastrophic collapse due to compounding errors for long-horizon planning.} The introduction of suboptimal decisions (4 optimal : 8 suboptimal) severely undermines the model's long-horizon planning capabilities. While the model retains partial capability on short-horizon tasks, its performance plummets to near-zero on middle and long-horizon tasks. In sequential decision-making, a suboptimal step early in the trajectory does not merely cause a localized failure; rather, the error accumulates and amplifies step-by-step. While agents might tolerate minor deviations in short-horizon tasks, these cascading errors in extended plans steer the agent irrecoverably off course, preventing it from completing the objective within the required step limits.

\begin{guidancebox}{Takeaway}
\textbf{Imperfect pre-training data severely degrades multi-turn long-horizon planning.} In real-world environments, pre-training data naturally contains non-optimal trajectories instead of perfect expert demonstrations. These sub-optimal paths introduce reasoning errors that accumulate heavily over long horizons. When trained on this mixed data, the agent fails to complete long tasks. Trying to overcome these learned flaws by simply increasing sampling is ineffective, as it fails to improve the success rate in long-horizon tasks. Furthermore, in multi-turn scenarios, relying on massive sampling causes unacceptable computational costs, making it impractical for real-world deployments.
\end{guidancebox}

\section{Planning Ability Shaping via GRPO and On-Policy Agentic Distillation}
\label{sec:opd}

\subsection{Mutual Information for Planning Pattern and Planning Knowledge}

\paragraph{Mutual Information.}
Inspired by~\citet{wang2026ragen}, we use mutual information to measure the relevance between the input task ($T$) and the two distinct outputs: planning pattern ($P$) and planning knowledge ($K$). To reflect their different properties, we define their mutual information separately as:
\begin{equation}
    I(T; P) = H(P) - H(P | T), \quad I(T; K) = H(K) - H(K | T)
\end{equation}
where $I(T; P)$ and $I(T; K)$ quantify the reduction in uncertainty about $P$ and $K$, respectively, when the task $T$ is given. Due to the abstract and reusable characteristics of the planning pattern $P$, it is less dependent on the input task $T$ and can be shared across different samples. Therefore, knowing $T$ provides limited information about $P$, resulting in a lower mutual information $I(T;P)$. In contrast, the planning knowledge $K$ contains task-specific information and is closely related to the input task $T$. Thus, knowing $T$ can greatly reduce the uncertainty of $K$, leading to a higher mutual information $I(T;K)$.

To illustrate, general planning patterns $P$, such as reflection or backtracking, often span across diverse samples. Conversely, the planning knowledge $K$ (e.g., specific skills and procedural knowledge) is highly specific and tightly bound to the current state of a specific sample. While different tasks may certainly necessitate different planning patterns, the input task $T$ imposes far stricter constraints on the specific planning knowledge required than on the broad strategy used. This relative difference implies that knowing $T$ reduces significantly more uncertainty about $K$ than about $P$, establishing that $I(T; P) < I(T; K)$.

\subsection{Impact of Planning Pattern}

\subsubsection{Accuracy Analysis}

\begin{figure}[h]
    \centering
    \includegraphics[width=\textwidth]{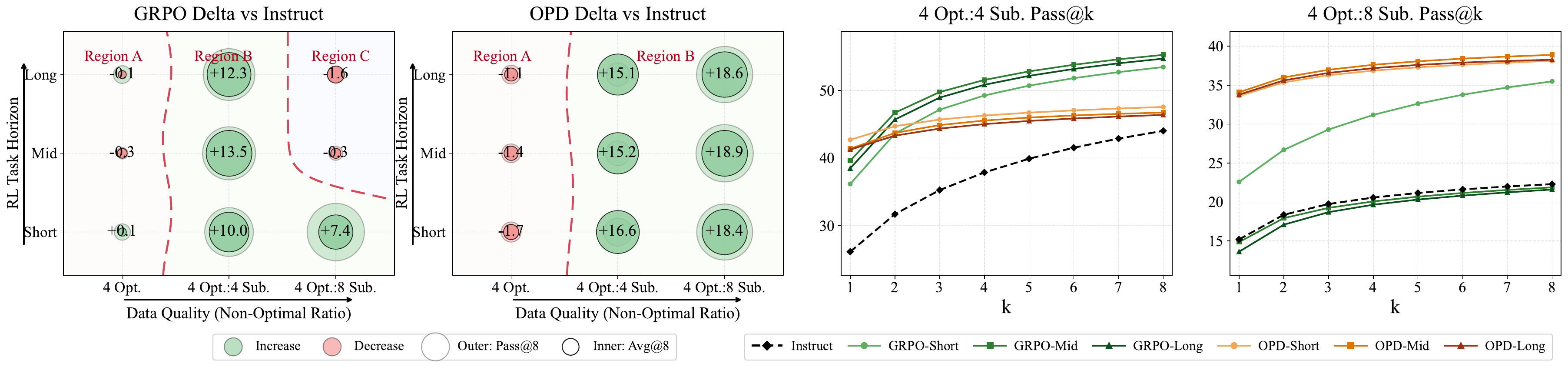}
    \caption{Applicability boundaries and post-training capability upper bounds of RL training. The first two panels show the applicability boundaries of multi-turn GRPO and OPD across RL task horizon and non-optimal action ratios (pre-training data quality). The last two panels compare the post-training capability upper bounds of different models.}
    \label{fig:reward_text_bar} 
\end{figure}

\paragraph{RL Applicability Regions.}
We define three regions to characterize when RL can improve planning.
The regions are determined by the performance gap among planning patterns and the ability of RL to discover better patterns.

\vspace{-10pt}
\begin{itemize}

\item \textbf{\colorbox{gray!20}{Region A: Unnecessary Region.}}
\quad
Different planning patterns achieve similar performance in this region.
RL-based selection is unnecessary because choosing different patterns leads to comparable results.

\item \textbf{\colorbox{green!20}{Region B: Effective Region.}}
\quad
Different planning patterns show clear performance differences in this region.
RL can discover better patterns through optimization and achieve consistent performance improvements.

\item \textbf{\colorbox{red!20}{Region C: Unsupported Region.}}
\quad
Better planning patterns exist in this region, but discovering them depends on the RL algorithm design.
An inappropriate RL algorithm may fail to activate the best planning pattern.

\end{itemize}

\vspace{-10pt}
\paragraph{Training Setup.} (1) \textbf{Base models.} As shown in Section \ref{subsec:planning_patterns}, we use three pre-trained base models: 4 Opt, 4 Opt:4 Sub, and 4 Opt:8 Sub. (2) \textbf{RL data.} We use the post-training data from Table \ref{tab:paper2datasetstats}. To test cross-task generalization, we entirely use the Fantasy Alchemy subset as the post-training data. To test cross-horizon generalization, we use data from short, middle, and long horizons. We downsample 3000 data items during post-training, but training process may not use all of them limited in training steps and batch size. (3) \textbf{RL algorithms.} We use multi-turn GRPO and multi-turn OPD algorithms. Multi-turn GRPO uses the binary final outcome reward. Multi-turn OPD uses Top-k OPD, where K is 100.


\begin{table}[!htbp]
\centering
\footnotesize
\setlength{\tabcolsep}{2.2pt}
\caption{Pattern-base comparison table. Rows are grouped by the pattern mixture: 4 Opt., 4 Opt.:4 Sub., and 4 Opt.:8 Sub. Each group contains the instruct checkpoint and multi-turn OPD/GRPO models trained on short, middle, and long levels. Two global blocks are reported (top: avg@8, bottom: pass@8). The last column reports the mean over short/mid/long evaluation difficulty means and its triangle delta vs. the instruct baseline in the same group. Domain abbreviations: FA = Fantasy Alchemy; LF = Livestock Farming; EA = Electronic Assembly.}
\label{tab:pattern_base_instruct_opd_grpo}
\resizebox{0.97\columnwidth}{!}{
\begin{tabular}{llccccccccccccc}
\toprule
\multirow{2}{*}{Pattern Mix} & \multirow{2}{*}{Model} & \multicolumn{4}{c}{Short} & \multicolumn{4}{c}{Mid} & \multicolumn{4}{c}{Long} & \multicolumn{1}{c}{Avg} \\
\cmidrule(lr){3-6} \cmidrule(lr){7-10} \cmidrule(lr){11-14} \cmidrule(lr){15-15}
 &  & FA & LF & EA & Mean & FA & LF & EA & Mean & FA & LF & EA & Mean & S/M/L Mean ($\triangle$) \\
\midrule
\rowcolor{BlockHeader}
\multicolumn{15}{c}{\textbf{avg@8}} \\
\midrule
\multirow{7}{*}{4 Opt.} & Instruct & \cellcolor{AvgCell!41}97.11 & \cellcolor{AvgCell!41}96.64 & \cellcolor{AvgCell!40}95.16 & \cellcolor{AvgCell!40}96.30 & \cellcolor{AvgCell!35}83.28 & \cellcolor{AvgCell!33}79.53 & \cellcolor{AvgCell!35}82.58 & \cellcolor{AvgCell!34}81.80 & \cellcolor{AvgCell!27}65.39 & \cellcolor{AvgCell!21}50.70 & \cellcolor{AvgCell!27}63.20 & \cellcolor{AvgCell!25}59.77 & \cellcolor{AvgCell!33}79.29\hspace{0.35em}\textcolor{green!55!black}{\scriptsize $\triangle$+0.00} \\
 & Short OPD & \cellcolor{AvgCell!41}97.27 & \cellcolor{AvgCell!40}94.77 & \cellcolor{AvgCell!40}95.55 & \cellcolor{AvgCell!40}95.86 & \cellcolor{AvgCell!34}81.95 & \cellcolor{AvgCell!32}76.64 & \cellcolor{AvgCell!33}78.52 & \cellcolor{AvgCell!33}79.04 & \cellcolor{AvgCell!27}63.12 & \cellcolor{AvgCell!21}48.98 & \cellcolor{AvgCell!26}61.88 & \cellcolor{AvgCell!24}57.99 & \cellcolor{AvgCell!33}77.63\hspace{0.35em}\textcolor{red!65!black}{\scriptsize $\triangle$-1.66} \\
 & Short GRPO & \cellcolor{AvgCell!41}97.03 & \cellcolor{AvgCell!41}96.88 & \cellcolor{AvgCell!40}95.16 & \cellcolor{AvgCell!40}96.35 & \cellcolor{AvgCell!35}83.28 & \cellcolor{AvgCell!33}79.14 & \cellcolor{AvgCell!35}82.89 & \cellcolor{AvgCell!34}81.77 & \cellcolor{AvgCell!27}65.47 & \cellcolor{AvgCell!21}50.86 & \cellcolor{AvgCell!27}63.75 & \cellcolor{AvgCell!25}60.03 & \cellcolor{AvgCell!33}79.38\hspace{0.35em}\textcolor{green!55!black}{\scriptsize $\triangle$+0.10} \\
 & Mid OPD & \cellcolor{AvgCell!41}97.81 & \cellcolor{AvgCell!40}95.47 & \cellcolor{AvgCell!40}94.92 & \cellcolor{AvgCell!40}96.07 & \cellcolor{AvgCell!34}81.33 & \cellcolor{AvgCell!33}78.83 & \cellcolor{AvgCell!33}78.67 & \cellcolor{AvgCell!33}79.61 & \cellcolor{AvgCell!27}64.61 & \cellcolor{AvgCell!20}48.44 & \cellcolor{AvgCell!26}61.09 & \cellcolor{AvgCell!24}58.05 & \cellcolor{AvgCell!33}77.91\hspace{0.35em}\textcolor{red!65!black}{\scriptsize $\triangle$-1.38} \\
 & Mid GRPO & \cellcolor{AvgCell!41}96.95 & \cellcolor{AvgCell!41}96.64 & \cellcolor{AvgCell!40}95.08 & \cellcolor{AvgCell!40}96.22 & \cellcolor{AvgCell!35}82.34 & \cellcolor{AvgCell!33}78.20 & \cellcolor{AvgCell!35}82.73 & \cellcolor{AvgCell!34}81.09 & \cellcolor{AvgCell!27}64.45 & \cellcolor{AvgCell!21}51.09 & \cellcolor{AvgCell!27}63.52 & \cellcolor{AvgCell!25}59.69 & \cellcolor{AvgCell!33}79.00\hspace{0.35em}\textcolor{red!65!black}{\scriptsize $\triangle$-0.29} \\
 & Long OPD & \cellcolor{AvgCell!41}96.88 & \cellcolor{AvgCell!40}95.55 & \cellcolor{AvgCell!40}95.31 & \cellcolor{AvgCell!40}95.91 & \cellcolor{AvgCell!34}81.17 & \cellcolor{AvgCell!33}78.83 & \cellcolor{AvgCell!35}82.66 & \cellcolor{AvgCell!34}80.89 & \cellcolor{AvgCell!26}61.88 & \cellcolor{AvgCell!21}50.08 & \cellcolor{AvgCell!26}61.48 & \cellcolor{AvgCell!24}57.81 & \cellcolor{AvgCell!33}78.20\hspace{0.35em}\textcolor{red!65!black}{\scriptsize $\triangle$-1.09} \\
 & Long GRPO & \cellcolor{AvgCell!41}96.80 & \cellcolor{AvgCell!41}96.48 & \cellcolor{AvgCell!40}95.62 & \cellcolor{AvgCell!40}96.30 & \cellcolor{AvgCell!34}82.11 & \cellcolor{AvgCell!33}77.66 & \cellcolor{AvgCell!35}82.66 & \cellcolor{AvgCell!34}80.81 & \cellcolor{AvgCell!27}65.47 & \cellcolor{AvgCell!22}52.50 & \cellcolor{AvgCell!27}63.52 & \cellcolor{AvgCell!25}60.49 & \cellcolor{AvgCell!33}79.20\hspace{0.35em}\textcolor{red!65!black}{\scriptsize $\triangle$-0.09} \\
\cmidrule(lr){1-15}
\multirow{7}{*}{4 Opt.:4 Sub.} & Instruct & \cellcolor{AvgCell!32}75.00 & \cellcolor{AvgCell!26}61.41 & \cellcolor{AvgCell!29}67.97 & \cellcolor{AvgCell!29}68.12 & \cellcolor{AvgCell!5}12.97 & \cellcolor{AvgCell!4}9.69 & \cellcolor{AvgCell!2}5.31 & \cellcolor{AvgCell!4}9.32 & \cellcolor{AvgCell!1}1.56 & \cellcolor{AvgCell!0}0.39 & \cellcolor{AvgCell!0}0.86 & \cellcolor{AvgCell!0}0.94 & \cellcolor{AvgCell!11}26.13\hspace{0.35em}\textcolor{green!55!black}{\scriptsize $\triangle$+0.00} \\
 & Short OPD & \cellcolor{AvgCell!39}93.52 & \cellcolor{AvgCell!30}71.02 & \cellcolor{AvgCell!34}80.08 & \cellcolor{AvgCell!34}81.54 & \cellcolor{AvgCell!19}45.78 & \cellcolor{AvgCell!14}32.97 & \cellcolor{AvgCell!11}25.39 & \cellcolor{AvgCell!15}34.71 & \cellcolor{AvgCell!6}15.47 & \cellcolor{AvgCell!3}6.02 & \cellcolor{AvgCell!6}14.06 & \cellcolor{AvgCell!5}11.85 & \cellcolor{AvgCell!18}42.70\hspace{0.35em}\textcolor{green!55!black}{\scriptsize $\triangle$+16.57} \\
 & Short GRPO & \cellcolor{AvgCell!36}85.78 & \cellcolor{AvgCell!29}69.84 & \cellcolor{AvgCell!32}75.39 & \cellcolor{AvgCell!32}77.01 & \cellcolor{AvgCell!14}34.14 & \cellcolor{AvgCell!11}26.80 & \cellcolor{AvgCell!6}15.47 & \cellcolor{AvgCell!11}25.47 & \cellcolor{AvgCell!4}9.30 & \cellcolor{AvgCell!1}2.42 & \cellcolor{AvgCell!3}6.33 & \cellcolor{AvgCell!3}6.02 & \cellcolor{AvgCell!15}36.16\hspace{0.35em}\textcolor{green!55!black}{\scriptsize $\triangle$+10.03} \\
 & Mid OPD & \cellcolor{AvgCell!38}91.25 & \cellcolor{AvgCell!30}71.56 & \cellcolor{AvgCell!34}79.77 & \cellcolor{AvgCell!34}80.86 & \cellcolor{AvgCell!18}42.19 & \cellcolor{AvgCell!13}30.86 & \cellcolor{AvgCell!11}26.48 & \cellcolor{AvgCell!14}33.18 & \cellcolor{AvgCell!5}12.89 & \cellcolor{AvgCell!3}6.41 & \cellcolor{AvgCell!5}10.94 & \cellcolor{AvgCell!4}10.08 & \cellcolor{AvgCell!17}41.37\hspace{0.35em}\textcolor{green!55!black}{\scriptsize $\triangle$+15.24} \\
 & Mid GRPO & \cellcolor{AvgCell!37}87.73 & \cellcolor{AvgCell!29}70.00 & \cellcolor{AvgCell!33}77.58 & \cellcolor{AvgCell!33}78.44 & \cellcolor{AvgCell!18}42.42 & \cellcolor{AvgCell!13}31.25 & \cellcolor{AvgCell!9}20.62 & \cellcolor{AvgCell!13}31.43 & \cellcolor{AvgCell!6}13.91 & \cellcolor{AvgCell!2}4.38 & \cellcolor{AvgCell!4}8.52 & \cellcolor{AvgCell!4}8.93 & \cellcolor{AvgCell!17}39.60\hspace{0.35em}\textcolor{green!55!black}{\scriptsize $\triangle$+13.47} \\
 & Long OPD & \cellcolor{AvgCell!39}92.03 & \cellcolor{AvgCell!30}70.78 & \cellcolor{AvgCell!34}79.92 & \cellcolor{AvgCell!34}80.91 & \cellcolor{AvgCell!17}40.39 & \cellcolor{AvgCell!14}33.12 & \cellcolor{AvgCell!10}24.61 & \cellcolor{AvgCell!14}32.71 & \cellcolor{AvgCell!6}13.28 & \cellcolor{AvgCell!2}4.53 & \cellcolor{AvgCell!5}12.58 & \cellcolor{AvgCell!4}10.13 & \cellcolor{AvgCell!17}41.25\hspace{0.35em}\textcolor{green!55!black}{\scriptsize $\triangle$+15.12} \\
 & Long GRPO & \cellcolor{AvgCell!37}86.95 & \cellcolor{AvgCell!29}69.30 & \cellcolor{AvgCell!33}77.66 & \cellcolor{AvgCell!33}77.97 & \cellcolor{AvgCell!16}37.66 & \cellcolor{AvgCell!12}29.38 & \cellcolor{AvgCell!8}19.77 & \cellcolor{AvgCell!12}28.93 & \cellcolor{AvgCell!6}14.22 & \cellcolor{AvgCell!1}2.97 & \cellcolor{AvgCell!4}8.36 & \cellcolor{AvgCell!4}8.52 & \cellcolor{AvgCell!16}38.47\hspace{0.35em}\textcolor{green!55!black}{\scriptsize $\triangle$+12.34} \\
\cmidrule(lr){1-15}
\multirow{7}{*}{4 Opt.:8 Sub.} & Instruct & \cellcolor{AvgCell!22}52.58 & \cellcolor{AvgCell!17}40.55 & \cellcolor{AvgCell!18}43.28 & \cellcolor{AvgCell!19}45.47 & \cellcolor{AvgCell!0}0.16 & \cellcolor{AvgCell!0}0.16 & \cellcolor{AvgCell!0}0.00 & \cellcolor{AvgCell!0}0.10 & \cellcolor{AvgCell!0}0.00 & \cellcolor{AvgCell!0}0.00 & \cellcolor{AvgCell!0}0.00 & \cellcolor{AvgCell!0}0.00 & \cellcolor{AvgCell!6}15.19\hspace{0.35em}\textcolor{green!55!black}{\scriptsize $\triangle$+0.00} \\
 & Short OPD & \cellcolor{AvgCell!36}85.86 & \cellcolor{AvgCell!26}62.58 & \cellcolor{AvgCell!31}72.66 & \cellcolor{AvgCell!31}73.70 & \cellcolor{AvgCell!12}29.22 & \cellcolor{AvgCell!9}22.34 & \cellcolor{AvgCell!7}16.02 & \cellcolor{AvgCell!9}22.53 & \cellcolor{AvgCell!3}6.88 & \cellcolor{AvgCell!1}3.52 & \cellcolor{AvgCell!1}3.44 & \cellcolor{AvgCell!2}4.61 & \cellcolor{AvgCell!14}33.61\hspace{0.35em}\textcolor{green!55!black}{\scriptsize $\triangle$+18.42} \\
 & Short GRPO & \cellcolor{AvgCell!30}71.48 & \cellcolor{AvgCell!23}55.08 & \cellcolor{AvgCell!25}60.55 & \cellcolor{AvgCell!26}62.37 & \cellcolor{AvgCell!3}6.64 & \cellcolor{AvgCell!3}6.80 & \cellcolor{AvgCell!1}1.88 & \cellcolor{AvgCell!2}5.10 & \cellcolor{AvgCell!0}0.23 & \cellcolor{AvgCell!0}0.39 & \cellcolor{AvgCell!0}0.16 & \cellcolor{AvgCell!0}0.26 & \cellcolor{AvgCell!9}22.58\hspace{0.35em}\textcolor{green!55!black}{\scriptsize $\triangle$+7.39} \\
 & Mid OPD & \cellcolor{AvgCell!36}86.72 & \cellcolor{AvgCell!27}64.06 & \cellcolor{AvgCell!30}71.95 & \cellcolor{AvgCell!31}74.24 & \cellcolor{AvgCell!14}33.44 & \cellcolor{AvgCell!9}20.47 & \cellcolor{AvgCell!6}14.84 & \cellcolor{AvgCell!10}22.92 & \cellcolor{AvgCell!4}8.75 & \cellcolor{AvgCell!2}3.91 & \cellcolor{AvgCell!1}2.66 & \cellcolor{AvgCell!2}5.10 & \cellcolor{AvgCell!14}34.09\hspace{0.35em}\textcolor{green!55!black}{\scriptsize $\triangle$+18.90} \\
 & Mid GRPO & \cellcolor{AvgCell!22}52.11 & \cellcolor{AvgCell!16}38.98 & \cellcolor{AvgCell!18}42.42 & \cellcolor{AvgCell!19}44.51 & \cellcolor{AvgCell!0}0.08 & \cellcolor{AvgCell!0}0.08 & \cellcolor{AvgCell!0}0.00 & \cellcolor{AvgCell!0}0.05 & \cellcolor{AvgCell!0}0.00 & \cellcolor{AvgCell!0}0.00 & \cellcolor{AvgCell!0}0.00 & \cellcolor{AvgCell!0}0.00 & \cellcolor{AvgCell!6}14.85\hspace{0.35em}\textcolor{red!65!black}{\scriptsize $\triangle$-0.34} \\
 & Long OPD & \cellcolor{AvgCell!36}85.00 & \cellcolor{AvgCell!26}62.89 & \cellcolor{AvgCell!30}71.80 & \cellcolor{AvgCell!31}73.23 & \cellcolor{AvgCell!13}31.64 & \cellcolor{AvgCell!9}20.78 & \cellcolor{AvgCell!7}16.09 & \cellcolor{AvgCell!10}22.84 & \cellcolor{AvgCell!3}7.81 & \cellcolor{AvgCell!2}3.67 & \cellcolor{AvgCell!2}4.38 & \cellcolor{AvgCell!2}5.29 & \cellcolor{AvgCell!14}33.78\hspace{0.35em}\textcolor{green!55!black}{\scriptsize $\triangle$+18.59} \\
 & Long GRPO & \cellcolor{AvgCell!19}45.55 & \cellcolor{AvgCell!16}37.81 & \cellcolor{AvgCell!16}38.91 & \cellcolor{AvgCell!17}40.76 & \cellcolor{AvgCell!0}0.08 & \cellcolor{AvgCell!0}0.16 & \cellcolor{AvgCell!0}0.00 & \cellcolor{AvgCell!0}0.08 & \cellcolor{AvgCell!0}0.00 & \cellcolor{AvgCell!0}0.00 & \cellcolor{AvgCell!0}0.00 & \cellcolor{AvgCell!0}0.00 & \cellcolor{AvgCell!6}13.61\hspace{0.35em}\textcolor{red!65!black}{\scriptsize $\triangle$-1.58} \\
\midrule
\rowcolor{BlockHeader}
\multicolumn{15}{c}{\textbf{pass@8}} \\
\midrule
\multirow{7}{*}{4 Opt.} & Instruct & \cellcolor{PassCell!41}97.50 & \cellcolor{PassCell!41}97.50 & \cellcolor{PassCell!40}96.25 & \cellcolor{PassCell!41}97.08 & \cellcolor{PassCell!36}86.25 & \cellcolor{PassCell!35}83.12 & \cellcolor{PassCell!36}86.25 & \cellcolor{PassCell!36}85.21 & \cellcolor{PassCell!30}72.50 & \cellcolor{PassCell!23}55.00 & \cellcolor{PassCell!28}66.25 & \cellcolor{PassCell!27}64.58 & \cellcolor{PassCell!35}82.29\hspace{0.35em}\textcolor{green!55!black}{\scriptsize $\triangle$+0.00} \\
 & Short OPD & \cellcolor{PassCell!41}98.12 & \cellcolor{PassCell!41}96.88 & \cellcolor{PassCell!40}95.62 & \cellcolor{PassCell!41}96.88 & \cellcolor{PassCell!36}85.00 & \cellcolor{PassCell!34}80.00 & \cellcolor{PassCell!35}82.50 & \cellcolor{PassCell!35}82.50 & \cellcolor{PassCell!29}68.75 & \cellcolor{PassCell!23}54.38 & \cellcolor{PassCell!27}65.00 & \cellcolor{PassCell!26}62.71 & \cellcolor{PassCell!34}80.69\hspace{0.35em}\textcolor{red!65!black}{\scriptsize $\triangle$-1.60} \\
 & Short GRPO & \cellcolor{PassCell!41}97.50 & \cellcolor{PassCell!41}97.50 & \cellcolor{PassCell!40}96.25 & \cellcolor{PassCell!41}97.08 & \cellcolor{PassCell!37}88.12 & \cellcolor{PassCell!35}83.12 & \cellcolor{PassCell!37}88.12 & \cellcolor{PassCell!36}86.46 & \cellcolor{PassCell!30}72.50 & \cellcolor{PassCell!24}56.25 & \cellcolor{PassCell!28}67.50 & \cellcolor{PassCell!27}65.42 & \cellcolor{PassCell!35}82.99\hspace{0.35em}\textcolor{green!55!black}{\scriptsize $\triangle$+0.69} \\
 & Mid OPD & \cellcolor{PassCell!41}98.12 & \cellcolor{PassCell!40}96.25 & \cellcolor{PassCell!40}95.62 & \cellcolor{PassCell!41}96.67 & \cellcolor{PassCell!36}85.62 & \cellcolor{PassCell!33}79.38 & \cellcolor{PassCell!35}83.12 & \cellcolor{PassCell!35}82.71 & \cellcolor{PassCell!29}68.75 & \cellcolor{PassCell!23}55.00 & \cellcolor{PassCell!28}66.88 & \cellcolor{PassCell!27}63.54 & \cellcolor{PassCell!34}80.97\hspace{0.35em}\textcolor{red!65!black}{\scriptsize $\triangle$-1.32} \\
 & Mid GRPO & \cellcolor{PassCell!41}97.50 & \cellcolor{PassCell!41}97.50 & \cellcolor{PassCell!40}96.25 & \cellcolor{PassCell!41}97.08 & \cellcolor{PassCell!36}86.25 & \cellcolor{PassCell!35}83.12 & \cellcolor{PassCell!36}86.25 & \cellcolor{PassCell!36}85.21 & \cellcolor{PassCell!29}68.12 & \cellcolor{PassCell!24}57.50 & \cellcolor{PassCell!28}66.88 & \cellcolor{PassCell!27}64.17 & \cellcolor{PassCell!35}82.15\hspace{0.35em}\textcolor{red!65!black}{\scriptsize $\triangle$-0.14} \\
 & Long OPD & \cellcolor{PassCell!41}97.50 & \cellcolor{PassCell!41}96.88 & \cellcolor{PassCell!40}95.62 & \cellcolor{PassCell!41}96.67 & \cellcolor{PassCell!36}85.00 & \cellcolor{PassCell!34}80.00 & \cellcolor{PassCell!36}86.88 & \cellcolor{PassCell!35}83.96 & \cellcolor{PassCell!29}68.12 & \cellcolor{PassCell!23}54.38 & \cellcolor{PassCell!27}64.38 & \cellcolor{PassCell!26}62.29 & \cellcolor{PassCell!34}80.97\hspace{0.35em}\textcolor{red!65!black}{\scriptsize $\triangle$-1.32} \\
 & Long GRPO & \cellcolor{PassCell!41}97.50 & \cellcolor{PassCell!41}97.50 & \cellcolor{PassCell!40}95.62 & \cellcolor{PassCell!41}96.88 & \cellcolor{PassCell!36}86.88 & \cellcolor{PassCell!34}81.88 & \cellcolor{PassCell!38}89.38 & \cellcolor{PassCell!36}86.04 & \cellcolor{PassCell!30}71.88 & \cellcolor{PassCell!25}58.75 & \cellcolor{PassCell!29}69.38 & \cellcolor{PassCell!28}66.67 & \cellcolor{PassCell!35}83.19\hspace{0.35em}\textcolor{green!55!black}{\scriptsize $\triangle$+0.90} \\
\cmidrule(lr){1-15}
\multirow{7}{*}{4 Opt.:4 Sub.} & Instruct & \cellcolor{PassCell!40}95.62 & \cellcolor{PassCell!35}83.75 & \cellcolor{PassCell!37}88.75 & \cellcolor{PassCell!38}89.38 & \cellcolor{PassCell!19}45.62 & \cellcolor{PassCell!16}38.75 & \cellcolor{PassCell!11}25.62 & \cellcolor{PassCell!15}36.67 & \cellcolor{PassCell!4}9.38 & \cellcolor{PassCell!1}2.50 & \cellcolor{PassCell!3}6.25 & \cellcolor{PassCell!3}6.04 & \cellcolor{PassCell!18}44.03\hspace{0.35em}\textcolor{green!55!black}{\scriptsize $\triangle$+0.00} \\
 & Short OPD & \cellcolor{PassCell!40}95.62 & \cellcolor{PassCell!33}78.12 & \cellcolor{PassCell!35}84.38 & \cellcolor{PassCell!36}86.04 & \cellcolor{PassCell!22}51.25 & \cellcolor{PassCell!17}40.00 & \cellcolor{PassCell!13}31.25 & \cellcolor{PassCell!17}40.83 & \cellcolor{PassCell!8}20.00 & \cellcolor{PassCell!4}8.75 & \cellcolor{PassCell!8}18.75 & \cellcolor{PassCell!7}15.83 & \cellcolor{PassCell!20}47.57\hspace{0.35em}\textcolor{green!55!black}{\scriptsize $\triangle$+3.54} \\
 & Short GRPO & \cellcolor{PassCell!41}96.88 & \cellcolor{PassCell!36}86.25 & \cellcolor{PassCell!39}92.50 & \cellcolor{PassCell!39}91.88 & \cellcolor{PassCell!27}63.12 & \cellcolor{PassCell!20}47.50 & \cellcolor{PassCell!17}40.00 & \cellcolor{PassCell!21}50.21 & \cellcolor{PassCell!10}25.00 & \cellcolor{PassCell!4}9.38 & \cellcolor{PassCell!9}20.62 & \cellcolor{PassCell!8}18.33 & \cellcolor{PassCell!22}53.47\hspace{0.35em}\textcolor{green!55!black}{\scriptsize $\triangle$+9.44} \\
 & Mid OPD & \cellcolor{PassCell!39}92.50 & \cellcolor{PassCell!33}77.50 & \cellcolor{PassCell!36}85.00 & \cellcolor{PassCell!36}85.00 & \cellcolor{PassCell!22}51.25 & \cellcolor{PassCell!15}36.88 & \cellcolor{PassCell!14}34.38 & \cellcolor{PassCell!17}40.83 & \cellcolor{PassCell!7}17.50 & \cellcolor{PassCell!4}10.00 & \cellcolor{PassCell!7}15.62 & \cellcolor{PassCell!6}14.38 & \cellcolor{PassCell!20}46.74\hspace{0.35em}\textcolor{green!55!black}{\scriptsize $\triangle$+2.71} \\
 & Mid GRPO & \cellcolor{PassCell!41}96.88 & \cellcolor{PassCell!36}86.25 & \cellcolor{PassCell!39}91.88 & \cellcolor{PassCell!38}91.67 & \cellcolor{PassCell!28}66.25 & \cellcolor{PassCell!20}46.88 & \cellcolor{PassCell!19}45.62 & \cellcolor{PassCell!22}52.92 & \cellcolor{PassCell!12}27.50 & \cellcolor{PassCell!6}13.12 & \cellcolor{PassCell!10}23.12 & \cellcolor{PassCell!9}21.25 & \cellcolor{PassCell!23}55.28\hspace{0.35em}\textcolor{green!55!black}{\scriptsize $\triangle$+11.25} \\
 & Long OPD & \cellcolor{PassCell!39}93.75 & \cellcolor{PassCell!32}75.62 & \cellcolor{PassCell!36}85.00 & \cellcolor{PassCell!36}84.79 & \cellcolor{PassCell!20}47.50 & \cellcolor{PassCell!17}39.38 & \cellcolor{PassCell!14}34.38 & \cellcolor{PassCell!17}40.42 & \cellcolor{PassCell!7}17.50 & \cellcolor{PassCell!3}7.50 & \cellcolor{PassCell!7}16.88 & \cellcolor{PassCell!6}13.96 & \cellcolor{PassCell!19}46.39\hspace{0.35em}\textcolor{green!55!black}{\scriptsize $\triangle$+2.36} \\
 & Long GRPO & \cellcolor{PassCell!41}96.88 & \cellcolor{PassCell!36}86.88 & \cellcolor{PassCell!38}90.62 & \cellcolor{PassCell!38}91.46 & \cellcolor{PassCell!26}61.88 & \cellcolor{PassCell!20}46.88 & \cellcolor{PassCell!18}43.75 & \cellcolor{PassCell!21}50.83 & \cellcolor{PassCell!14}32.50 & \cellcolor{PassCell!5}11.88 & \cellcolor{PassCell!9}21.25 & \cellcolor{PassCell!9}21.88 & \cellcolor{PassCell!23}54.72\hspace{0.35em}\textcolor{green!55!black}{\scriptsize $\triangle$+10.69} \\
\cmidrule(lr){1-15}
\multirow{7}{*}{4 Opt.:8 Sub.} & Instruct & \cellcolor{PassCell!31}73.12 & \cellcolor{PassCell!25}60.00 & \cellcolor{PassCell!27}65.00 & \cellcolor{PassCell!28}66.04 & \cellcolor{PassCell!1}1.25 & \cellcolor{PassCell!1}1.25 & \cellcolor{PassCell!0}0.00 & \cellcolor{PassCell!0}0.83 & \cellcolor{PassCell!0}0.00 & \cellcolor{PassCell!0}0.00 & \cellcolor{PassCell!0}0.00 & \cellcolor{PassCell!0}0.00 & \cellcolor{PassCell!9}22.29\hspace{0.35em}\textcolor{green!55!black}{\scriptsize $\triangle$+0.00} \\
 & Short OPD & \cellcolor{PassCell!37}87.50 & \cellcolor{PassCell!29}68.12 & \cellcolor{PassCell!33}78.12 & \cellcolor{PassCell!33}77.92 & \cellcolor{PassCell!15}35.00 & \cellcolor{PassCell!12}27.50 & \cellcolor{PassCell!9}21.88 & \cellcolor{PassCell!12}28.12 & \cellcolor{PassCell!5}12.50 & \cellcolor{PassCell!2}5.00 & \cellcolor{PassCell!3}7.50 & \cellcolor{PassCell!4}8.33 & \cellcolor{PassCell!16}38.12\hspace{0.35em}\textcolor{green!55!black}{\scriptsize $\triangle$+15.83} \\
 & Short GRPO & \cellcolor{PassCell!38}90.62 & \cellcolor{PassCell!33}78.12 & \cellcolor{PassCell!34}80.62 & \cellcolor{PassCell!35}83.12 & \cellcolor{PassCell!12}28.12 & \cellcolor{PassCell!10}25.00 & \cellcolor{PassCell!5}11.25 & \cellcolor{PassCell!9}21.46 & \cellcolor{PassCell!1}1.88 & \cellcolor{PassCell!1}2.50 & \cellcolor{PassCell!1}1.25 & \cellcolor{PassCell!1}1.88 & \cellcolor{PassCell!15}35.49\hspace{0.35em}\textcolor{green!55!black}{\scriptsize $\triangle$+13.19} \\
 & Mid OPD & \cellcolor{PassCell!37}88.75 & \cellcolor{PassCell!30}71.25 & \cellcolor{PassCell!33}78.75 & \cellcolor{PassCell!33}79.58 & \cellcolor{PassCell!17}41.25 & \cellcolor{PassCell!11}26.25 & \cellcolor{PassCell!9}20.62 & \cellcolor{PassCell!12}29.38 & \cellcolor{PassCell!5}12.50 & \cellcolor{PassCell!2}5.62 & \cellcolor{PassCell!2}5.00 & \cellcolor{PassCell!3}7.71 & \cellcolor{PassCell!16}38.89\hspace{0.35em}\textcolor{green!55!black}{\scriptsize $\triangle$+16.60} \\
 & Mid GRPO & \cellcolor{PassCell!30}71.25 & \cellcolor{PassCell!24}58.12 & \cellcolor{PassCell!28}66.25 & \cellcolor{PassCell!27}65.21 & \cellcolor{PassCell!0}0.62 & \cellcolor{PassCell!0}0.62 & \cellcolor{PassCell!0}0.00 & \cellcolor{PassCell!0}0.42 & \cellcolor{PassCell!0}0.00 & \cellcolor{PassCell!0}0.00 & \cellcolor{PassCell!0}0.00 & \cellcolor{PassCell!0}0.00 & \cellcolor{PassCell!9}21.88\hspace{0.35em}\textcolor{red!65!black}{\scriptsize $\triangle$-0.42} \\
 & Long OPD & \cellcolor{PassCell!38}89.38 & \cellcolor{PassCell!29}70.00 & \cellcolor{PassCell!32}76.25 & \cellcolor{PassCell!33}78.54 & \cellcolor{PassCell!15}36.25 & \cellcolor{PassCell!12}27.50 & \cellcolor{PassCell!9}21.88 & \cellcolor{PassCell!12}28.54 & \cellcolor{PassCell!5}11.25 & \cellcolor{PassCell!2}5.62 & \cellcolor{PassCell!3}6.25 & \cellcolor{PassCell!3}7.71 & \cellcolor{PassCell!16}38.26\hspace{0.35em}\textcolor{green!55!black}{\scriptsize $\triangle$+15.97} \\
 & Long GRPO & \cellcolor{PassCell!30}70.62 & \cellcolor{PassCell!25}59.38 & \cellcolor{PassCell!26}62.50 & \cellcolor{PassCell!27}64.17 & \cellcolor{PassCell!0}0.62 & \cellcolor{PassCell!1}1.25 & \cellcolor{PassCell!0}0.00 & \cellcolor{PassCell!0}0.62 & \cellcolor{PassCell!0}0.00 & \cellcolor{PassCell!0}0.00 & \cellcolor{PassCell!0}0.00 & \cellcolor{PassCell!0}0.00 & \cellcolor{PassCell!9}21.60\hspace{0.35em}\textcolor{red!65!black}{\scriptsize $\triangle$-0.69} \\
\bottomrule
\end{tabular}
}
\end{table}

\begin{figure}[!htbp]
    \centering
    \includegraphics[width=0.97\textwidth]{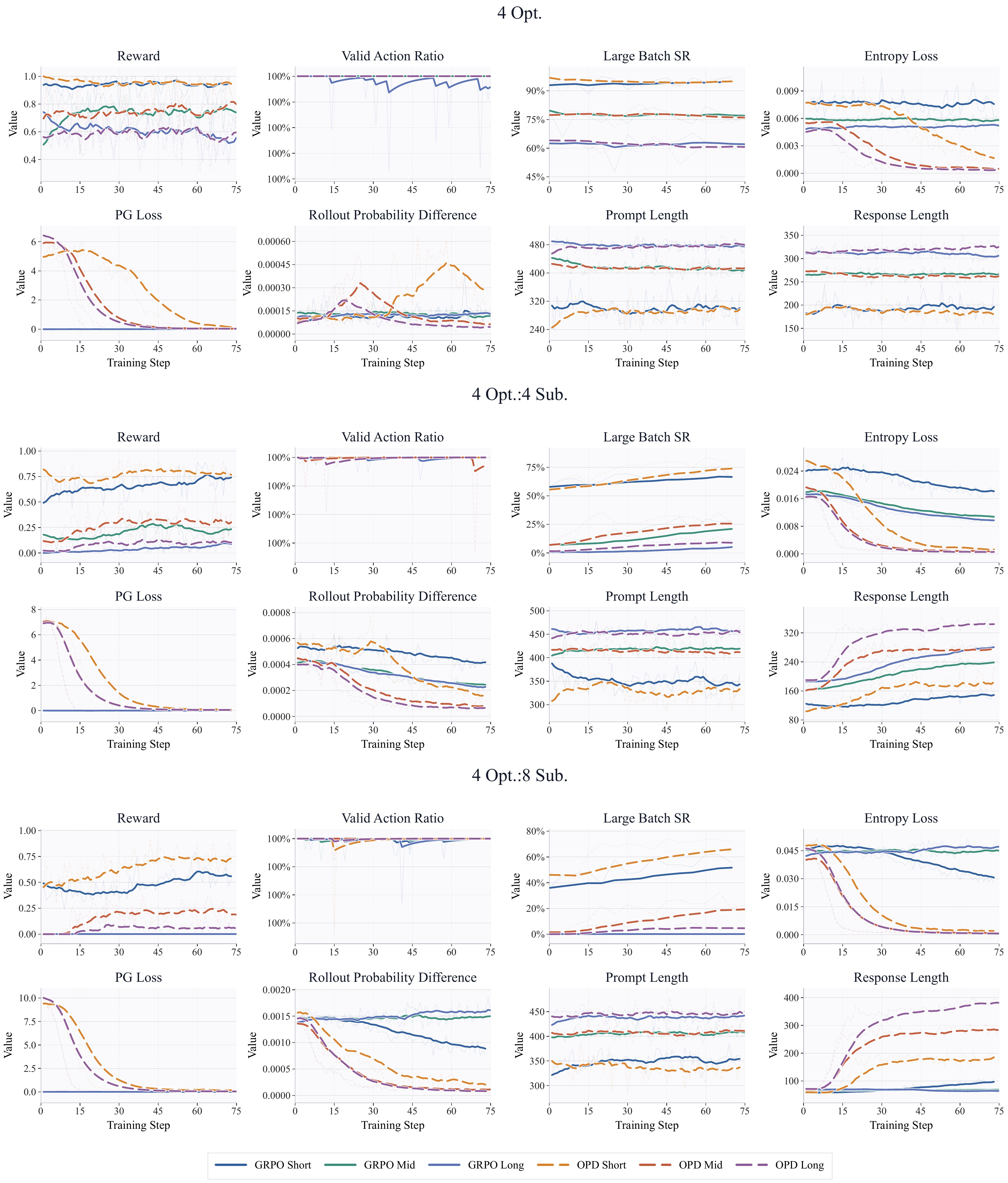}
    \caption{Training dynamics across optimization settings.
Comparison of multi-turn GRPO and OPD under 4 Opt., 4 Opt.:4 Sub., and 4 Opt.:8 Sub. settings. }
    \label{fig:multi_reward_curves} 
\end{figure}

\paragraph{Results.} 
(1) \textbf{GRPO and OPD have applicable boundaries based on RL task horizon and data quality (non-optimal action ratio),} as shown in Figure~\ref{fig:reward_text_bar}, Table~\ref{tab:pattern_base_instruct_opd_grpo}, and Figure~\ref{fig:multi_reward_curves}. There are three specific regions. \textbf{Region A is the unnecessary region, where planning knowledge tokens are less suitable to RL and are better acquired through SFT. Different planning patterns achieve comparable performance, making RL unnecessary.} For GRPO and OPD, even if the teacher performance is 100\%, high mutual information planning knowledge tokens are difficult to improve via RL and hard to generalize across samples. These tokens need to be injected via SFT. Moreover, the planning knowledge in this region is highly sample-specific and difficult to transfer across instances. The parameter updates induced by RL optimization are insufficient to fully encode such fine-grained planning knowledge into the model, even with high-quality teacher supervision, making SFT a more effective approach for injecting these patterns. \textbf{Region B is the effective region. Low mutual information patterns generalize well through RL.} Low mutual information planning pattern tokens are easy to stimulate via RL and can generalize across samples. This is essentially a process of selecting planning patterns from low-quality pretraining data that are statistically more likely to achieve higher final rewards. \textbf{Region C is the unsupported region for long horizon tasks. Long horizon tasks suffer from incorrect credit assignment due to sparse outcome rewards and therefore require fine-grained reward methods.} In long-horizon tasks, when both non-optimal action ratio and RL task horizon increase, correct and incorrect actions are mixed in very long trajectories. ORM based GRPO will assign rewards incorrectly, failing to optimize long horizon agent tasks. In contrast, fine grained reward methods like OPD provide a promising approach.
(2) \textbf{In the effective region, GRPO and OPD show similar performance improvements. However, OPD reduces entropy and performs worse than the approximate upper bound of GRPO's pass@8 performance,} as shown in Figure~\ref{fig:reward_text_bar}. In 4 Opt.:4 Sub., OPD leads in pass@1, while GRPO achieves a higher ceiling with larger k. In 4 Opt.:8 Sub., GRPO helps short-horizon but offers little gain on long-horizon tasks. Meanwhile, OPD shows better pass@1 and pass@8. This indicates that GRPO generates the best planning pattern based on the environment, showing stronger potential. However, OPD suppresses diversity and directly follows the planning pattern of the teacher. Therefore, it is clear that the correctness of the teacher planning pattern is extremely important. 
(3) \textbf{Pass@k may not always provide an accurate characterization of the upper bound of model capability in multi-turn long-horizon planning scenarios,} as shown in Figure~\ref{fig:reward_text_bar} and Table~\ref{tab:pattern_base_instruct_opd_grpo}. This observation is motivated by the fact that the pass@k performance before RL training is nearly zero in these long-horizon settings, while RL training can lead to substantial improvements. Such a near-zero pass@k score does not necessarily imply that the base model lacks the ability to generate successful trajectories, as the necessary knowledge or solution patterns may exist in the pretrained model but are rarely sampled. Instead, it suggests that pass@k may gradually become a less accurate estimator of the model capability upper bound as the planning horizon becomes increasingly long.
(4) \textbf{RL shift CoT length toward high-reward patterns in pre-training data.} After training, the output length of model will change, which is related to the data distribution in pretraining. If the better planning pattern is longer, the trained output length will increase. If the better planning pattern is shorter, the trained output length may decrease.

\begin{guidancebox}{Takeaway}
\textbf{Planning knowledge is better suited for SFT (unnecessary region).} When planning knowledge has high mutual information, different samples may require different optimization directions. The limited update magnitude of RL makes it difficult to inject such knowledge effectively.

\textbf{Planning patterns are better suited for RL (effective region).} When planning patterns have low mutual information, different samples share similar optimization directions. RL can stimulate these patterns and improve generalization across samples that contain similar patterns in the pretraining corpus with high expected rewards.

\textbf{Multi-turn long-horizon tasks need fine grained credit assignment like OPD (GRPO unsupported region).} Outcome based rewards struggle when correct and incorrect actions mix in long trajectories, making fine grained reward based methods necessary. OPD demonstrates a larger effective region than GRPO with an ideal teacher.

\textbf{OPD follows teacher patterns to improve performance, but it reduces entropy and limits its upper bound.} In the effective region, OPD performs similarly to GRPO but scales worse with more sampling times. 

\textbf{Although pass@k exceeds the base model with large k in long-horizon tasks, we do not think that RL improves the model's capability ceiling.} Instead, we argue that pass@k is distorted in this setting because pre-training data may already contain the correct trajectories.

\textbf{RL shifts CoT length toward high-reward patterns in the pretraining data.} The output length increases depending on whether the better planning patterns are longer.
\end{guidancebox}

\subsubsection{Gradient Direction Analysis}

\paragraph{Top-k Subspace for SVD.}

\begin{figure}[h]
    \centering
    \includegraphics[width=\textwidth]{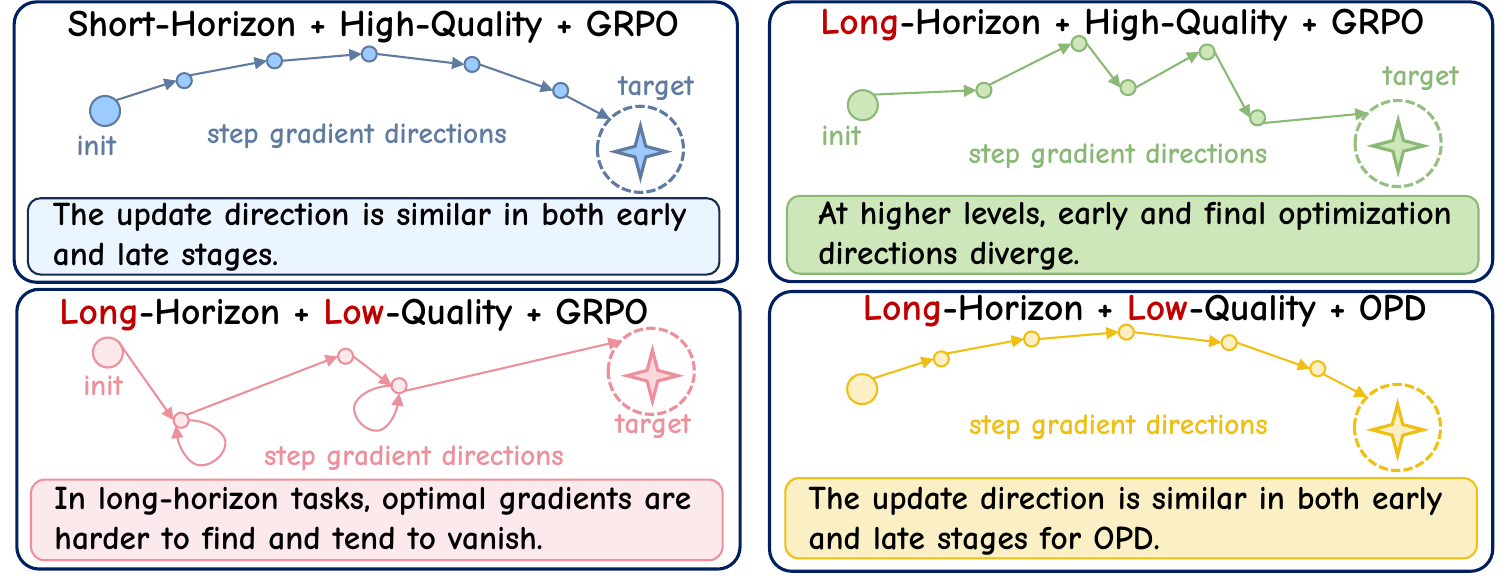}
    \caption{Comparison of step gradient directions between GRPO and OPD under varying optimization conditions.}
    \label{fig:SVD} 
\end{figure}

\begin{figure}[h]
    \centering
    \includegraphics[width=\textwidth]{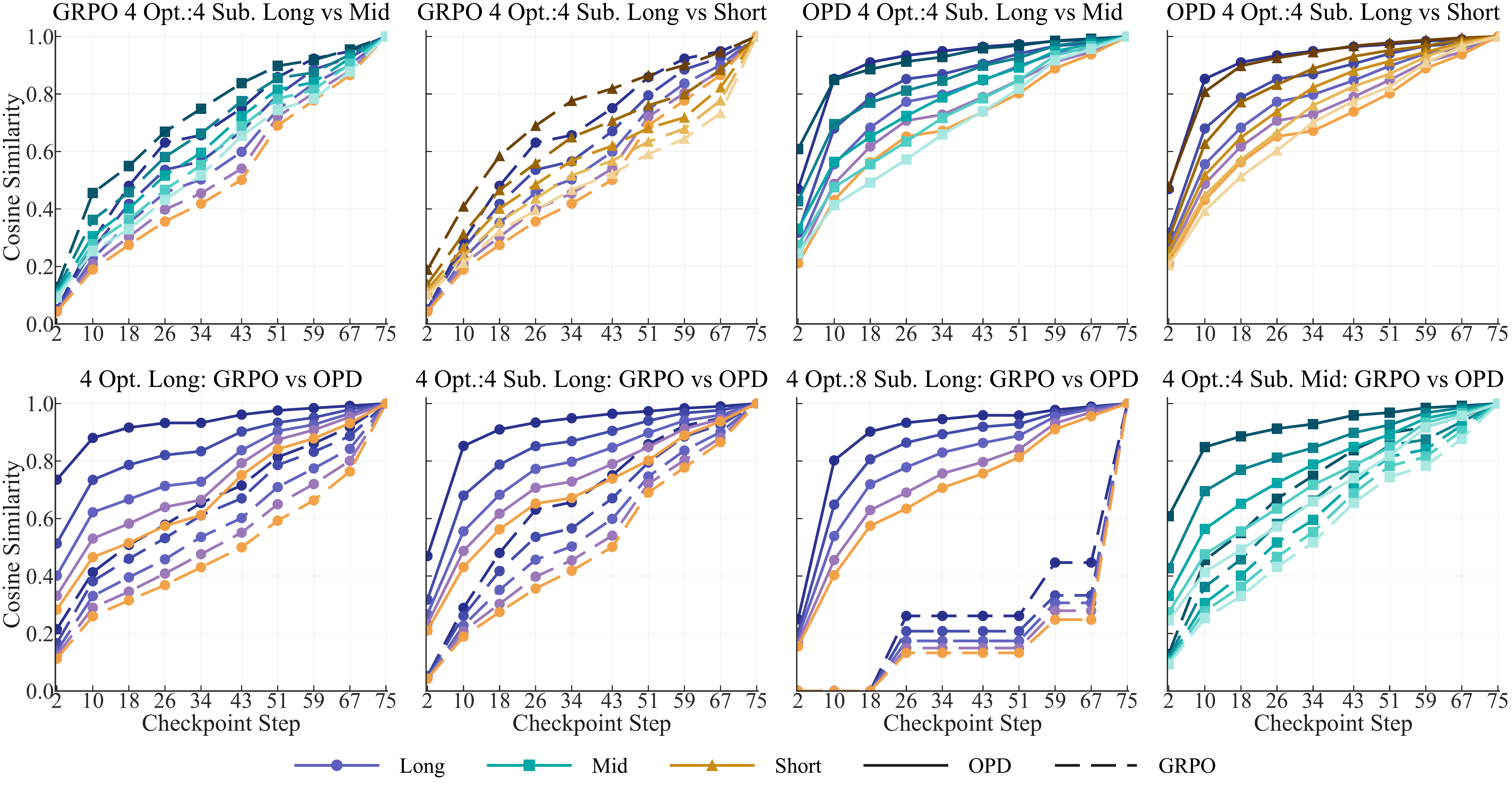}
    \caption{Evolution of gradient directions. Top-5 SVD direction similarity between intermediate and final checkpoints.}
    \label{fig:para_direction_top5} 
    \vspace{-13pt}
\end{figure}

Beyond measuring task performance, we analyze whether the model already follows the final parameter-update direction in the early stages of training. \citet{cai2026learning} analyzes the gradient direction. We further show how the update direction differs under two dimensions: pre-training data quality and post-training horizon. Specifically, we compare the dominant update directions of early and intermediate checkpoints with those of the final checkpoint.
Let $W_{\ell}^{(t)}$ be the weight matrix of layer $\ell$ at checkpoint $t$, and let $b$ be the base checkpoint.
We first define the update matrix
\begin{equation}
\Delta W_{\ell}^{(t)} = W_{\ell}^{(t)} - W_{\ell}^{(b)}.
\end{equation}
For each $\Delta W_{\ell}^{(t)}$, we compute SVD:
\begin{equation}
\Delta W_{\ell}^{(t)} = U_{\ell}^{(t)} \Sigma_{\ell}^{(t)} \left(V_{\ell}^{(t)}\right)^{\top}.
\end{equation}
We keep the first $k$ left singular vectors,
\begin{equation}
U_{\ell,k}^{(t)} = \big[u_{\ell,1}^{(t)}, u_{\ell,2}^{(t)}, \dots, u_{\ell,k}^{(t)}\big].
\end{equation}
Given a target final checkpoint $f$, we compare checkpoint $t$ with $f$ by cosine similarity of matched singular vectors:
\begin{equation}
s_{\ell,i}(t,f) = \left| \frac{\left(u_{\ell,i}^{(t)}\right)^\top u_{\ell,i}^{(f)}}
{\left\|u_{\ell,i}^{(t)}\right\|_2 \, \left\|u_{\ell,i}^{(f)}\right\|_2} \right|,
\quad i=1,\dots,k.
\end{equation}
We use absolute value because SVD vectors can flip sign.
The layer-level Top-$k$ score is
\begin{equation}
s_{\ell}^{(k)}(t,f) = \frac{1}{k}\sum_{i=1}^{k} s_{\ell,i}(t,f).
\end{equation}
Finally, we average over all valid layers $\mathcal{L}$:
\begin{equation}
S^{(k)}(t,f)
=
\frac{1}{|\mathcal{L}|}
\sum_{\ell \in \mathcal{L}} s_{\ell}^{(k)}(t,f),
\quad
S^{(k)}(t,f)\in[0,1].
\end{equation}
A larger value means the update direction at checkpoint $t$ is closer to the final update direction.

\paragraph{Experimental Setup.} (1) \textbf{Horizon length.} Under the 4 Opt.:4 Sub setting, we evaluate both the GRPO and OPD algorithms. For each configuration, we conduct pairwise comparisons across the short, middle, and long levels. (2) \textbf{Algorithm comparison.} We fix the difficulty at the long-horizon and utilize three settings: 4 Opt., 4 Opt.:4 Sub, and 4 Opt.:8 Sub. This part primarily compares the parameter update directions between the GRPO and OPD algorithms under long-horizon conditions. (3) \textbf{Top-k Subspace.} We select $k=5$ for the top-$k$ subspace analysis. For every layer, we compute the parameters of both the FFN and Attention modules. The maximum number of training steps is set to 75. We select step 1 as the initial checkpoint and uniformly sample 10 intermediate checkpoints for evaluation.

\paragraph{Results.} As shown in Figure~\ref{fig:SVD} and Figure~\ref{fig:para_direction_top5}.
(1) \textbf{Coarse credit assignment leads to noisy GRPO gradient direction in long horizons.} The first row display the performance of the GRPO algorithm across different horizon levels. The plotted curves visually demonstrate that as the task horizon increases, the numerical values remain lower. This directly indicates that as the horizon becomes longer, the coarse granularity of credit assignment causes the parameter update direction of GRPO to become inconsistent.
(2) \textbf{Vanishing gradients with increasing planning horizon and non-optimal actions.} Observing the subfigures in the second row, specific GRPO curves begin exactly at zero and remain completely flat along the x-axis during the initial checkpoint steps. This distinct flat trajectory at zero provides evidence that continuously increasing the number of non-optimal actions triggers a vanishing gradient issue. Because of this issue, the GRPO gradients fail to undergo any updates.
(3) \textbf{Superior directional consistency of OPD.} The second row of subfigures directly compares the GRPO and OPD algorithms across multiple configurations. The charts illustrate that the GRPO curves only reach higher values under restricted simple conditions. This proves that the update direction of GRPO is not universally consistent, relies heavily on specific applicability conditions, and is unsuitable for real long horizon scenarios. In contrast, the OPD curves maintain a relatively high level across all evaluated settings. This confirms that OPD maintains a much more consistent parameter update direction.

\begin{guidancebox}{Takeaway}
\textbf{OPD exhibits stronger directional stability than GRPO under long-horizon and low-quality pre-training corpus settings.} While GRPO shows inconsistent or even vanishing update directions as task horizon and non-optimal actions increase, OPD maintains consistently high subspace alignment, suggesting more reliable credit assignment in ideal teacher condition.
\end{guidancebox}

\subsection{Impact of Planning Knowledge}
\label{sec:planning_knowledge}
\subsubsection{Accuracy Analysis}
\begin{figure}[h]
    \centering
    \includegraphics[width=\textwidth]{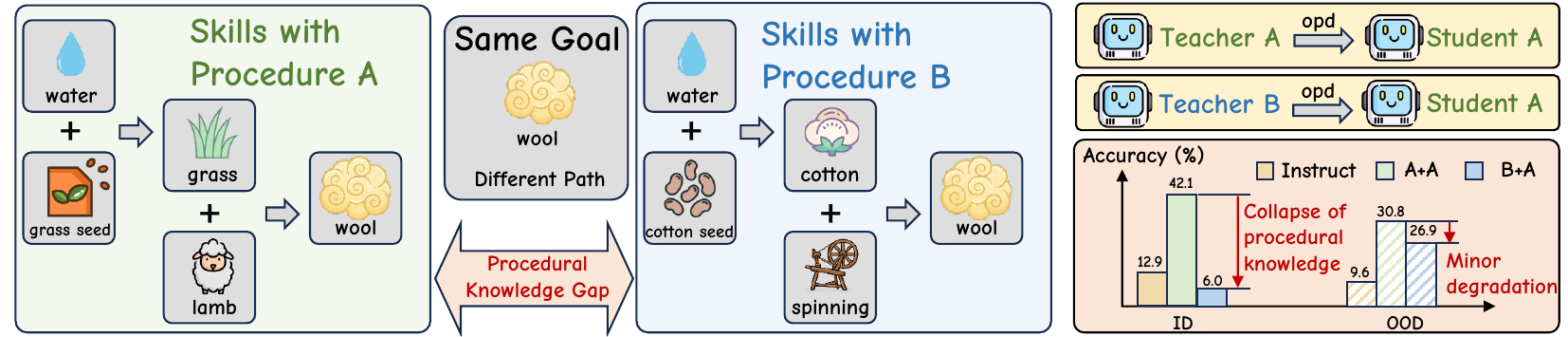}
    \caption{The procedural knowledge gap during teacher-student distillation. Distilling a different procedural path for the same goal via the OPD method (teacher B to student A) leads to a substantial collapse of procedural knowledge and severely damages ID performance, while mildly affecting OOD generalization.}
    \label{fig:gap} 
\end{figure}
\paragraph{Knowledge Gap in OPD.} In real-world environments, there are multiple reasonable and effective paths to complete a task (e.g., Recipe A and Recipe B). This multi-path characteristic exposes a core issue in knowledge distillation: when the student and teacher models form divergent procedural knowledge based on different pre-training corpora, can this knowledge be effectively distilled? To this end, by comparing the mismatched setting (student with recipe A + teacher with recipe B) with the aligned setting (student with recipe A + teacher with recipe A), we investigate whether two completely different sets of procedural knowledge can successfully bridge this significant gap and be distilled. Specifically, we explore the scenario where the student model conducts procedural knowledge world modeling based on Recipe A, while the teacher model provides procedural knowledge world modeling based on Recipe B.
As shown in Figure \ref{fig:gap}, we illustrate this gap using a specific crafting example where both procedures share the same final goal of producing wool. Procedure A creates wool using water, grass seed, and a lamb. Procedure B achieves the identical goal using water, cotton seed, and spinning. When Student A, who is pre-trained on Procedure A, receives distillation from Teacher B via the OPD method, a severe collapse of procedural knowledge occurs. This structural mismatch leads to a significant drop in in-distribution accuracy compared to the aligned setting where Teacher A is used.
\paragraph{Task Setup.} (1) \textbf{Models.} For the base model, we use 4 Opt.:4 Sub setting. For the teacher models, we train the teacher models separately using Recipe A and Recipe B, respectively. Although these two recipes use different raw materials and input materials for the same target, both can successfully make the target. (2) \textbf{Training.} We train the models on the middle-horizon RL training set. The subset uses FA domain.
\begin{figure}[h]
    \centering
    \includegraphics[width=\textwidth]{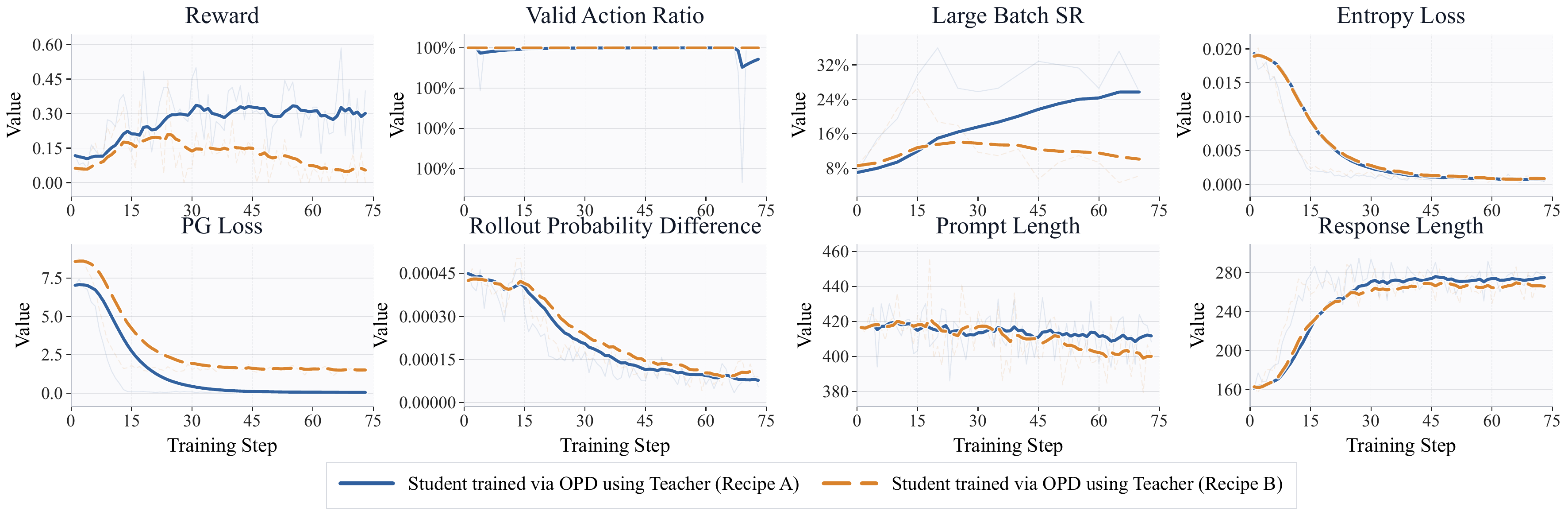}
    \caption{Comparison of training dynamics under two OPD teacher recipes.}
    \label{fig:multi_reward_gap} 
\end{figure}

\begin{table}[h]
\centering
\footnotesize
\setlength{\tabcolsep}{2.2pt}
\caption{Knowledge gap comparison for the 4 Opt.:4 Sub pattern mixture in RL. The final $\Delta (B - A)$ row reports Student + Teacher B minus Student + Teacher A.}
\label{tab:pretrain_gap_4opt_4sub}
\resizebox{0.97\columnwidth}{!}{
\begin{tabular}{llccccccccccccc}
\toprule
\multirow{2}{*}{Pattern Mix} & \multirow{2}{*}{Model} & \multicolumn{4}{c}{Short} & \multicolumn{4}{c}{Middle} & \multicolumn{4}{c}{Long} & \multicolumn{1}{c}{Avg} \\
\cmidrule(lr){3-6} \cmidrule(lr){7-10} \cmidrule(lr){11-14} \cmidrule(lr){15-15}
 &  & FA & LF & EA & Mean & FA & LF & EA & Mean & FA & LF & EA & Mean & S/M/L Mean ($\triangle$) \\
\midrule
\rowcolor{BlockHeader}
\multicolumn{15}{c}{\textbf{avg@8}} \\
\midrule
\multirow{6}{*}{4 Opt.:4 Sub.} & Instruct Model & \cellcolor{AvgCell!32}75.00 & \cellcolor{AvgCell!26}61.41 & \cellcolor{AvgCell!29}67.97 & \cellcolor{AvgCell!29}68.12 & \cellcolor{AvgCell!5}12.97 & \cellcolor{AvgCell!4}9.69 & \cellcolor{AvgCell!2}5.31 & \cellcolor{AvgCell!4}9.32 & \cellcolor{AvgCell!1}1.56 & \cellcolor{AvgCell!0}0.39 & \cellcolor{AvgCell!0}0.86 & \cellcolor{AvgCell!0}0.94 & \cellcolor{AvgCell!11}26.13\hspace{0.35em}\textcolor{green!55!black}{\scriptsize $\triangle$+0.00} \\
 & Teacher A & \cellcolor{AvgCell!42}100.00 & -- & -- & \cellcolor{AvgCell!42}100.00 & \cellcolor{AvgCell!42}100.00 & -- & -- & \cellcolor{AvgCell!42}100.00 & \cellcolor{AvgCell!42}100.00 & -- & -- & \cellcolor{AvgCell!42}100.00 & \cellcolor{AvgCell!42}100.00\hspace{0.35em}\textcolor{green!55!black}{\scriptsize $\triangle$+73.87} \\
 & Teacher B & \cellcolor{AvgCell!42}100.00 & -- & -- & \cellcolor{AvgCell!42}100.00 & \cellcolor{AvgCell!42}100.00 & -- & -- & \cellcolor{AvgCell!42}100.00 & \cellcolor{AvgCell!42}100.00 & -- & -- & \cellcolor{AvgCell!42}100.00 & \cellcolor{AvgCell!42}100.00\hspace{0.35em}\textcolor{green!55!black}{\scriptsize $\triangle$+73.87} \\
 & Student + Teacher A & \cellcolor{AvgCell!38}91.25 & \cellcolor{AvgCell!30}71.56 & \cellcolor{AvgCell!34}79.77 & \cellcolor{AvgCell!34}80.86 & \cellcolor{AvgCell!18}42.19 & \cellcolor{AvgCell!13}30.86 & \cellcolor{AvgCell!11}26.48 & \cellcolor{AvgCell!14}33.18 & \cellcolor{AvgCell!5}12.89 & \cellcolor{AvgCell!3}6.41 & \cellcolor{AvgCell!5}10.94 & \cellcolor{AvgCell!4}10.08 & \cellcolor{AvgCell!17}41.37\hspace{0.35em}\textcolor{green!55!black}{\scriptsize $\triangle$+15.24} \\
 & Student + Teacher B & \cellcolor{AvgCell!30}72.19 & \cellcolor{AvgCell!27}63.91 & \cellcolor{AvgCell!33}78.75 & \cellcolor{AvgCell!30}71.61 & \cellcolor{AvgCell!3}6.09 & \cellcolor{AvgCell!11}26.95 & \cellcolor{AvgCell!9}20.78 & \cellcolor{AvgCell!8}17.94 & \cellcolor{AvgCell!1}2.27 & \cellcolor{AvgCell!2}5.31 & \cellcolor{AvgCell!3}7.27 & \cellcolor{AvgCell!2}4.95 & \cellcolor{AvgCell!13}31.50\hspace{0.35em}\textcolor{green!55!black}{\scriptsize $\triangle$+5.37} \\
 & $\Delta$ (B - A) & \cellcolor{DeltaCell!8}\textcolor{red!65!black}{-19.06} & \cellcolor{DeltaCell!3}\textcolor{red!65!black}{-7.66} & \cellcolor{DeltaCell!0}\textcolor{red!65!black}{-1.02} & \cellcolor{DeltaCell!4}\textcolor{red!65!black}{-9.24} & \cellcolor{DeltaCell!15}\textcolor{red!65!black}{-36.09} & \cellcolor{DeltaCell!2}\textcolor{red!65!black}{-3.91} & \cellcolor{DeltaCell!2}\textcolor{red!65!black}{-5.70} & \cellcolor{DeltaCell!6}\textcolor{red!65!black}{-15.23} & \cellcolor{DeltaCell!4}\textcolor{red!65!black}{-10.62} & \cellcolor{DeltaCell!0}\textcolor{red!65!black}{-1.09} & \cellcolor{DeltaCell!2}\textcolor{red!65!black}{-3.67} & \cellcolor{DeltaCell!2}\textcolor{red!65!black}{-5.13} & \cellcolor{DeltaCell!4}\textcolor{red!65!black}{-9.87} \\
\midrule
\rowcolor{BlockHeader}
\multicolumn{15}{c}{\textbf{pass@8}} \\
\midrule
\multirow{6}{*}{4 Opt.:4 Sub.} & Instruct Model & \cellcolor{PassCell!40}95.62 & \cellcolor{PassCell!35}83.75 & \cellcolor{PassCell!37}88.75 & \cellcolor{PassCell!38}89.38 & \cellcolor{PassCell!19}45.62 & \cellcolor{PassCell!16}38.75 & \cellcolor{PassCell!11}25.62 & \cellcolor{PassCell!15}36.67 & \cellcolor{PassCell!4}9.38 & \cellcolor{PassCell!1}2.50 & \cellcolor{PassCell!3}6.25 & \cellcolor{PassCell!3}6.04 & \cellcolor{PassCell!18}44.03\hspace{0.35em}\textcolor{green!55!black}{\scriptsize $\triangle$+0.00} \\
 & Teacher A & \cellcolor{PassCell!42}100.00 & -- & -- & \cellcolor{PassCell!42}100.00 & \cellcolor{PassCell!42}100.00 & -- & -- & \cellcolor{PassCell!42}100.00 & \cellcolor{PassCell!42}100.00 & -- & -- & \cellcolor{PassCell!42}100.00 & \cellcolor{PassCell!42}100.00\hspace{0.35em}\textcolor{green!55!black}{\scriptsize $\triangle$+55.97} \\
 & Teacher B & \cellcolor{PassCell!42}100.00 & -- & -- & \cellcolor{PassCell!42}100.00 & \cellcolor{PassCell!42}100.00 & -- & -- & \cellcolor{PassCell!42}100.00 & \cellcolor{PassCell!42}100.00 & -- & -- & \cellcolor{PassCell!42}100.00 & \cellcolor{PassCell!42}100.00\hspace{0.35em}\textcolor{green!55!black}{\scriptsize $\triangle$+55.97} \\
 & Student + Teacher A & \cellcolor{PassCell!39}92.50 & \cellcolor{PassCell!33}77.50 & \cellcolor{PassCell!36}85.00 & \cellcolor{PassCell!36}85.00 & \cellcolor{PassCell!22}51.25 & \cellcolor{PassCell!15}36.88 & \cellcolor{PassCell!14}34.38 & \cellcolor{PassCell!17}40.83 & \cellcolor{PassCell!7}17.50 & \cellcolor{PassCell!4}10.00 & \cellcolor{PassCell!7}15.62 & \cellcolor{PassCell!6}14.38 & \cellcolor{PassCell!20}46.74\hspace{0.35em}\textcolor{green!55!black}{\scriptsize $\triangle$+2.71} \\
 & Student + Teacher B & \cellcolor{PassCell!32}75.62 & \cellcolor{PassCell!31}73.75 & \cellcolor{PassCell!36}85.00 & \cellcolor{PassCell!33}78.12 & \cellcolor{PassCell!3}7.50 & \cellcolor{PassCell!14}34.38 & \cellcolor{PassCell!12}29.38 & \cellcolor{PassCell!10}23.75 & \cellcolor{PassCell!2}3.75 & \cellcolor{PassCell!4}8.75 & \cellcolor{PassCell!5}11.25 & \cellcolor{PassCell!3}7.92 & \cellcolor{PassCell!15}36.60\hspace{0.35em}\textcolor{red!65!black}{\scriptsize $\triangle$-7.43} \\
 & $\Delta$ (B - A) & \cellcolor{DeltaCell!7}\textcolor{red!65!black}{-16.88} & \cellcolor{DeltaCell!2}\textcolor{red!65!black}{-3.75} & \cellcolor{DeltaCell!0}\textcolor{green!55!black}{+0.00} & \cellcolor{DeltaCell!3}\textcolor{red!65!black}{-6.88} & \cellcolor{DeltaCell!18}\textcolor{red!65!black}{-43.75} & \cellcolor{DeltaCell!1}\textcolor{red!65!black}{-2.50} & \cellcolor{DeltaCell!2}\textcolor{red!65!black}{-5.00} & \cellcolor{DeltaCell!7}\textcolor{red!65!black}{-17.08} & \cellcolor{DeltaCell!6}\textcolor{red!65!black}{-13.75} & \cellcolor{DeltaCell!1}\textcolor{red!65!black}{-1.25} & \cellcolor{DeltaCell!2}\textcolor{red!65!black}{-4.38} & \cellcolor{DeltaCell!3}\textcolor{red!65!black}{-6.46} & \cellcolor{DeltaCell!4}\textcolor{red!65!black}{-10.14} \\
\bottomrule
\end{tabular}
}
\end{table}

\paragraph{Results.} As shown in Figure~\ref{fig:multi_reward_gap} and Table~\ref{tab:pretrain_gap_4opt_4sub}. 
(1) \textbf{Knowledge mismatch blocks distillation even with perfect performance teachers.} Teacher B achieves a perfect score of 100.00 in all tasks. However, training a student with a different recipe fails completely. The Student + Teacher B model gets an overall pass@8 score of only 36.60. This is worse than the base instruct model score of 44.03. This shows that a highly capable teacher cannot improve the student if their training recipes do not match. 
(2) \textbf{Mismatched procedural distillation collapses ID planning knowledge while mildly affecting OOD generalization.} When the teacher and student acquire different but both valid procedural paths for the same goal, forcing distillation of the mismatched path causes a substantial collapse of the student's original ID planning knowledge. This disrupts the planning knowledge tokens for FA world modeling, leading to severe ID performance degradation, while LF and EA are less affected because their corresponding world modeling is not directly overwritten. For example, in the Middle difficulty FA task, Student + Teacher B reduces avg@8 and pass@8 by 36.09 and 43.75 points compared with Student + Teacher A.
(3) \textbf{Large knowledge gap causes training collapse in later stages.} Figure~\ref{fig:multi_reward_gap} shows that the performance goes up in the early stage because the student aligns the planning pattern. In the later stage, the main optimization direction is to align the planning knowledge. However, the knowledge gap between the two recipes is too big, which finally leads to a complete training collapse.

\begin{guidancebox}{Takeaway}
\textbf{OPD requires aligned procedural knowledge between student and teacher.} OPD cannot effectively distill knowledge from a perfect teacher when the student and teacher internalize different procedural paths for the same goal. Successful distillation requires similar internal world modeling between the teacher and student.

\textbf{Mismatched procedural distillation collapses ID planning knowledge while mildly affecting OOD generalization.} When teacher and student follow different valid procedures, distilling the mismatched path overwrites the student’s existing planning knowledge tokens.

\textbf{OPD distills patterns early but fails on mismatched knowledge late.} In the early stage, OPD successfully distills better planning patterns to improve performance. In the later stage, forcing OPD to distill completely different procedural knowledge leads to a training collapse.
\end{guidancebox}

\subsubsection{KL Dynamic Analysis}

\begin{figure}[h]
    \centering
    \includegraphics[width=\textwidth]{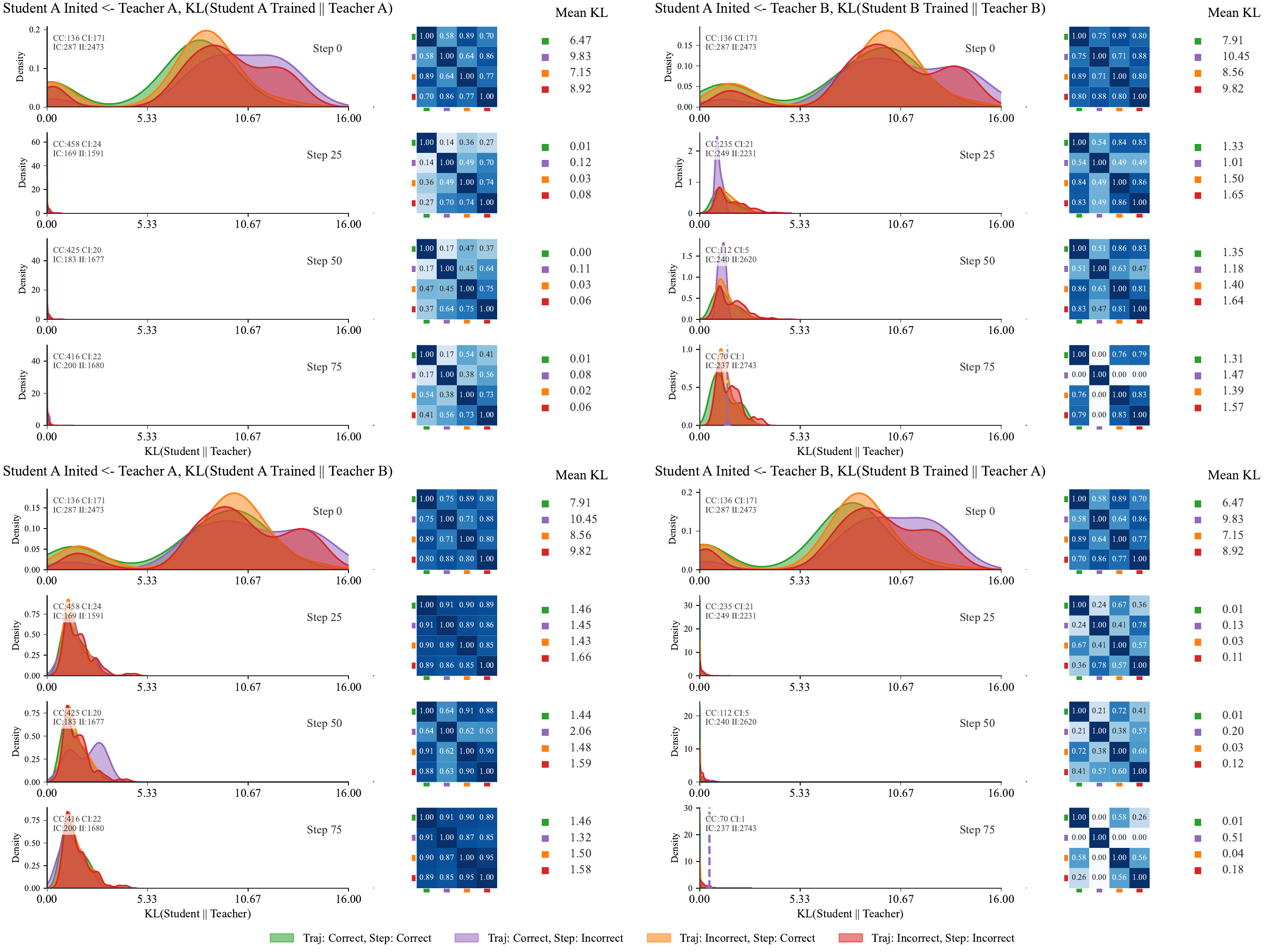}
    \caption{Token-level KL divergence distributions between students and matched or crossed teachers across training steps. Samples are grouped by trajectory-level and step-level correctness.}
    \label{fig:student_teacher_kl_correlation} 
\end{figure}

\paragraph{KL Dynamic and Overlap.}
We use KL divergence to measure how close a trained student is to one of the two teacher models at the token distribution level. Let $T_A$ and $T_B$ denote the two teacher models. All student checkpoints start from the same A recipe. We use $\sigma\in\{A,B\}$ to denote which teacher is used to train the student, and $\gamma\in\{A,B\}$ to denote which teacher is used for KL comparison. The student trained with teacher $T_\sigma$ is denoted as $\pi_{\theta_\sigma}$. A trajectory $\tau \sim \pi_{\theta_\sigma}$ is rolled out by this student. On this fixed trajectory, the student policy $p^\sigma_{k,t}$ and the comparison teacher policy $q^\gamma_{k,t}$ evaluate next-token probabilities at decoding step $t$ within turn $k$, conditioned on the same prefix. Let $\mathcal{V}_{k,t}$ be the top-$k$ token subset, and let $\ell^{p,\sigma}_{k,t}(v)$ and $\ell^{q,\gamma}_{k,t}(v)$ be the stored log probabilities for token $v$.
\begin{equation}
\begin{aligned}
p^{\sigma}_{k,t}(v) &= \exp\left(\ell^{p,\sigma}_{k,t}(v)\right), \quad
q^{\gamma}_{k,t}(v) = \exp\left(\ell^{q,\gamma}_{k,t}(v)\right), \\
p^{\sigma}_{k,t,\mathrm{tail}} &= 1-\sum_{v\in\mathcal{V}_{k,t}}p^{\sigma}_{k,t}(v), \quad
q^{\gamma}_{k,t,\mathrm{tail}} = 1-\sum_{v\in\mathcal{V}_{k,t}}q^{\gamma}_{k,t}(v).
\end{aligned}
\end{equation}

For each token position, we compute the KL from the student distribution to the selected teacher distribution, and average it over the whole trajectory. We use four comparison settings: setting 1 uses the student trained with $T_A$ and compares it with $T_A$; setting 2 uses the student trained with $T_B$ and compares it with $T_B$; setting 3 uses the student trained with $T_A$ and compares it with $T_B$; setting 4 uses the student trained with $T_B$ and compares it with $T_A$.
\begin{equation}
\mathrm{KL}^{\sigma,\gamma}(\tau)
=
\frac{1}{\sum_{k=1}^{K}T_k}
\sum_{k=1}^{K}
\sum_{t=1}^{T_k}
\sum_{v\in\mathcal{V}_{k,t}\cup\{\mathrm{tail}\}}
P^{\sigma}_{k,t}(v)
\log
\frac{P^{\sigma}_{k,t}(v)}{Q^{\gamma}_{k,t}(v)}.
\end{equation}
We further split trajectories into four data types according to trajectory correctness and step correctness: correct trajectory with correct step, correct trajectory with incorrect step, incorrect trajectory with correct step, and incorrect trajectory with incorrect step. Let $\mathcal{R}_{c}$ be the set of trajectories in data type $c$. For each training step, comparison setting $(\sigma,\gamma)$, and data type $c$, we report the mean KL. To compare two data types $c$ and $c'$, we estimate their KL density functions $f_c(x)$ and $f_{c'}(x)$ with Gaussian kernel density estimation, where $x$ is a trajectory-level KL value, and compute their overlap.
\begin{equation}
\overline{\mathrm{KL}}^{\sigma,\gamma}_{c}
=
\frac{1}{|\mathcal{R}_{c}|}
\sum_{\tau\in\mathcal{R}_{c}}
\mathrm{KL}^{\sigma,\gamma}(\tau),
\quad
\mathrm{Overlap}(c,c')
=
\int \min\left(f_c(x),f_{c'}(x)\right)\,dx.
\end{equation}

\paragraph{Task Setup.}
(1) \textbf{Data and model.} We use the two models trained in the Section~\ref{sec:planning_knowledge} to perform inference on the FA test set with middle difficulty. The rollout process uses two models pretrained from recipe A and trained later by teacher A and teacher B respectively. Then we calculate the reverse KL divergence. We also divide the rollout data into four types based on trajectory correctness and single step correctness to obtain the distribution, overlap rate, and mean of the KL divergence.
(2) \textbf{Distribution recording and divergence calculation.} During the KL divergence calculation stage, the models perform a rollout independently and record the probability distribution. Subsequently, teacher A and teacher B record the output probability distributions on these rollout trajectories and calculate the reverse KL divergence based on this.
(3) \textbf{Data classification and statistical settings.} We divide the rollout data into four types based on trajectory correctness and single step correctness to obtain the distribution, overlap rate, and mean of the KL divergence. At the same time, we set the K value of top K to 100 and calculate the RL training dynamics at four points, which are 0, 25, 50, and 75.

\begin{figure}[t]
    \centering
    \includegraphics[width=\textwidth]{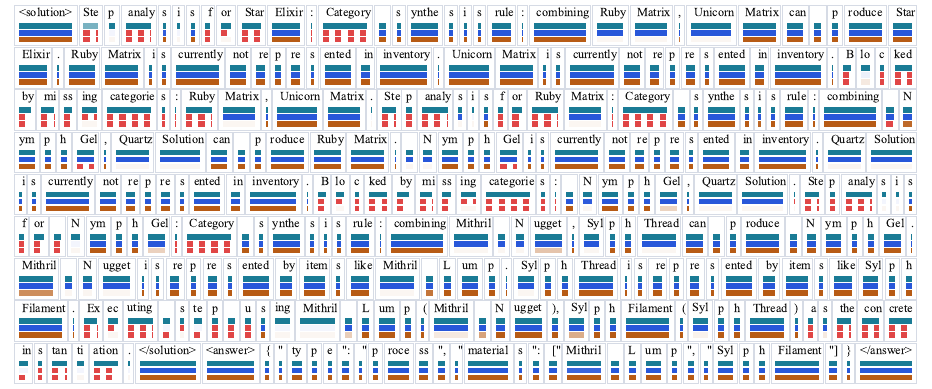}
    \caption{Token-level probability heatmap on a student A rollout. This figure shows an early-stage RL optimization, mainly focusing on optimizing the tokens corresponding to planning patterns that have high cross-sample generalization. For each token generated by student A, the top cyan, middle blue, and bottom orange bars show the probabilities assigned by student A, teacher A, and teacher B, respectively. Darker colors indicate higher probabilities. Red striped bars indicate that the generated token is absent from the corresponding teacher’s top-k predictions.}
    \label{fig:token_probability_heatbars_step0_row0} 
    \vspace{-10pt}
\end{figure}

\begin{figure}[t]
    \centering
    \includegraphics[width=\textwidth]{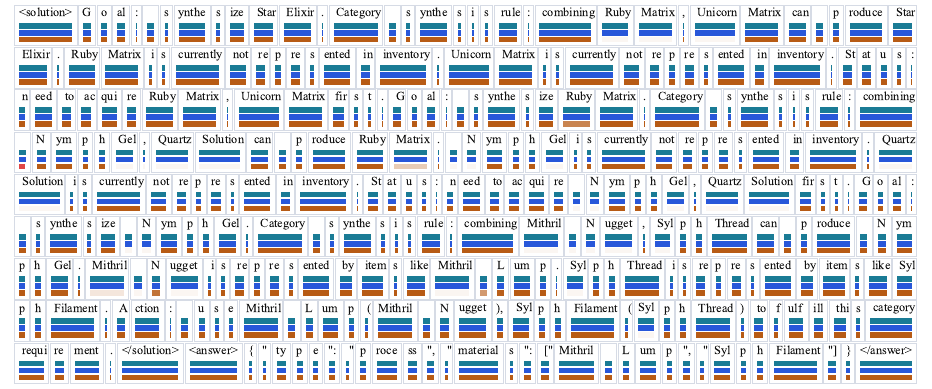}
    \caption{Token-level probability heatmap on a Student A rollout for step 75. This figure shows a later-stage RL optimization, which mainly focuses on optimizing sample-specific planning knowledge. However, the optimization capacity of RL is insufficient to inject highly specialized planning knowledge.}
    \label{fig:token_probability_heatbars_step75_row0} 
    \vspace{-10pt}
\end{figure}

\paragraph{Results.} As shown in Figure~\ref{fig:student_teacher_kl_correlation} and Figure~\ref{fig:token_probability_heatbars_step0_row0} and Figure~\ref{fig:token_probability_heatbars_step75_row0}. 
(1)\textbf{Need for error correction data.} The proportion of the four data types shows that the number of correct trajectories with incorrect steps is extremely small. This indicates a need to add more error correction data. Gold standard data alone fails to stimulate the recovery ability from error state prefixes.
(2)\textbf{Optimization of planning knowledge.} As shown in the first group of figures, the KL divergence on the left side is successfully optimized down. In contrast, after optimizing the planning patterns, the KL divergence on the right side becomes difficult to decrease further. Planning knowledge has large token mutual information and requires point by point optimization. The reinforcement learning update magnitude is too small and is insufficient to transfer massive amounts of knowledge to the domain of teacher B.
(3)\textbf{Reward correlation and knowledge disruption.} Based on the average KL values, the mean KL divergence for many correct steps is larger than that for incorrect steps. The correlation between the mean KL divergence and the environment reward is smaller for mismatched planning knowledge compared to matched planning knowledge. Completing a single sample involves multiple reasoning paths. If the teacher fails to cover these paths, the model treats other correct paths as incorrect ones. This process destroys the original knowledge of the model.
(4)\textbf{Transfer failure in later stages.} As shown in Figure~\ref{fig:token_probability_heatbars_step0_row0} and Figure~\ref{fig:token_probability_heatbars_step75_row0}, both models optimize the planning pattern in the early stages. In the later stages, teacher B primarily optimizes planning knowledge. However, planning knowledge varies greatly across different samples. The reinforcement learning update fails to transfer from knowledge A to knowledge B.

\begin{guidancebox}{Takeaway}
\textbf{Incorporate error correction data.} Relying solely on gold standard data is insufficient. Training pipelines need error state prefixes to teach models how to recover from mistakes.

\textbf{RL updates are too small to support sample-specific large-scale knowledge distillation.} Standard RL updates are too small for the point by point optimization required to distill massive planning knowledge.

\textbf{The same target can be reached by multiple different but correct reasoning paths.} If the teacher misses them, OPD may assign wrong probabilities to valid paths and hurt the student's existing knowledge via KL loss.

\textbf{Early pattern learning improves RL rewards, but failure to learn task-specific tokens may cause later degradation.} Models easily optimize general planning patterns early in training, but they struggle to distill high variance planning knowledge in later stages.
\end{guidancebox}

\section{Planning Ability Integration through Multi-Teacher On-Policy Agentic Distillation}
\label{sec:multi_opd}

\subsection{Definitions and Research Problems}

\textbf{Multi-Teacher On-Policy Agentic Distillation (MOPD)} is a method that distills and combines the abilities of multiple teacher models into one student model. Existing technical route mainly follow two paradigms, as shown in Table~\ref{tab:mopd_paradigms}. \textbf{(1) Cross domain capability integration.} This paradigm trains the student model in different environments based on the student model. After training, teacher models from different environments are merged to integrate the abilities of multiple experts~\citep{xiao2026mimo,xu2026deepseek}. \textbf{(2) Cross stage capability retention.} This paradigm uses curriculum learning to train the student model. The model learns from reasoning ability to agentic ability gradually. In the final stage, the final model from each stage is used as a teacher model and then merged to recover abilities that are forgotten in earlier stages~\citep{zeng2026glm,yang2026nemotron}.

\begin{table}[h]
    \centering
    \footnotesize
    \setlength{\tabcolsep}{4pt}
    \renewcommand{\arraystretch}{1.15} 
    \caption{Comparison of MOPD paradigms across different foundation models.}
    \label{tab:mopd_paradigms}
    \begin{tabularx}{\textwidth}{l X X X X}
        \toprule
        \textbf{Foundation Models} & \textbf{Training Paradigm} & \textbf{Student} & \textbf{Teacher} & \textbf{Primary Motivation} \\
        \midrule
        \makecell[tl]{\textbf{MiMo-V2-Flash} \\ \citep{xiao2026mimo}} & 
        Pre-training $\rightarrow$ \newline Domain training $\rightarrow$ \newline  \textcolor{red!70!black}{MOPD} & 
        SFT Model & Domain-specialized Model & Cross domain capability integration \\
        \midrule
        \makecell[tl]{\textbf{GLM-5} \\ \citep{zeng2026glm}} & Pre-training $\rightarrow$ \newline Mid-training $\rightarrow$ \newline Cascading RL $\rightarrow$ \newline \textcolor{red!70!black}{MOPD} & Last RL checkpoint & Previous RL checkpoint & Cross stage capability retention \\
        \midrule
        \makecell[tl]{\textbf{Nemotron-Cascade 2} \\ \citep{yang2026nemotron}} & Pre-training  $\rightarrow$ \newline Multi-domain RL $\rightarrow$ \newline \textcolor{red!70!black}{MOPD} $\rightarrow$ \newline Cascading RL & Middle RL checkpoint  & Previous RL checkpoint & Cross stage capability retention \\
        \midrule
        \makecell[tl]{\textbf{DeepSeek-V4} \\ \citep{xu2026deepseek}} & Pre-training $\rightarrow$ \newline Domain training $\rightarrow$ \newline \textcolor{red!70!black}{MOPD} & SFT Model & Domain-specialized Model & Cross domain capability integration \\
        \bottomrule
    \end{tabularx}
\end{table}

Multi-teacher agent-OPD can generally be categorized into two paradigms: mixture and cascaded. Given that the mixture approach incurs infrastructure overhead (e.g., simultaneous deployment and inference of multiple teacher models), we focus on the cascaded form. Under this sequential training paradigm, the student policy learns from one teacher at a time. For the $m$-th teacher ($m \in \{1, \dots, M\}$), the objective provides dense, step-by-step supervision by aligning the student's predictions with those of the specific teacher:
\begin{equation}
    \mathcal{L}_{\text{Agent-OPD}}^{(m)}(\theta) = \mathbb{E}_{\tau \sim \pi_\theta} \left[ \sum_{k=1}^{K} \sum_{t=1}^{T_k} D_{\text{KL}}(p_{k,t} \parallel q_{k,t}^{(m)}) \right],
\end{equation}
where $q_{k,t}^{(m)} \triangleq \pi_{\text{teacher}}^{(m)}(\cdot \mid H_k, \hat{Y}_{k,<t})$ denotes the next-token distribution provided by the $m$-th teacher. The overall training proceeds by iteratively minimizing $\mathcal{L}_{\text{Agent-OPD}}^{(m)}(\theta)$ in sequence from $m=1$ to $M$.

Although MOPD is widely used in existing foundation models, the training mechanisms and conditions for MOPD remain unclear. Therefore, we urgently need to analyze the MOPD algorithm deeply to explore its working conditions and boundaries. Thus, we propose three progressive research questions:

\begin{itemize}
    \item \textbf{Q1: Generalization Mode.} Under what conditions can MOPD effectively achieve cross-environment generalization?
    \item \textbf{Q2: Continual Learning Mode.} Under what conditions can MOPD address limited cross-environment generalization through continual learning without conflicts?
    \item \textbf{Q3: \textcolor{red!70!black}{Conflict Mode.}} Under what conditions does MOPD lose cross-environment generalization and suffer severe conflicts among environments?
\end{itemize}

\begin{figure}[h]
    \centering
    \includegraphics[width=\textwidth]{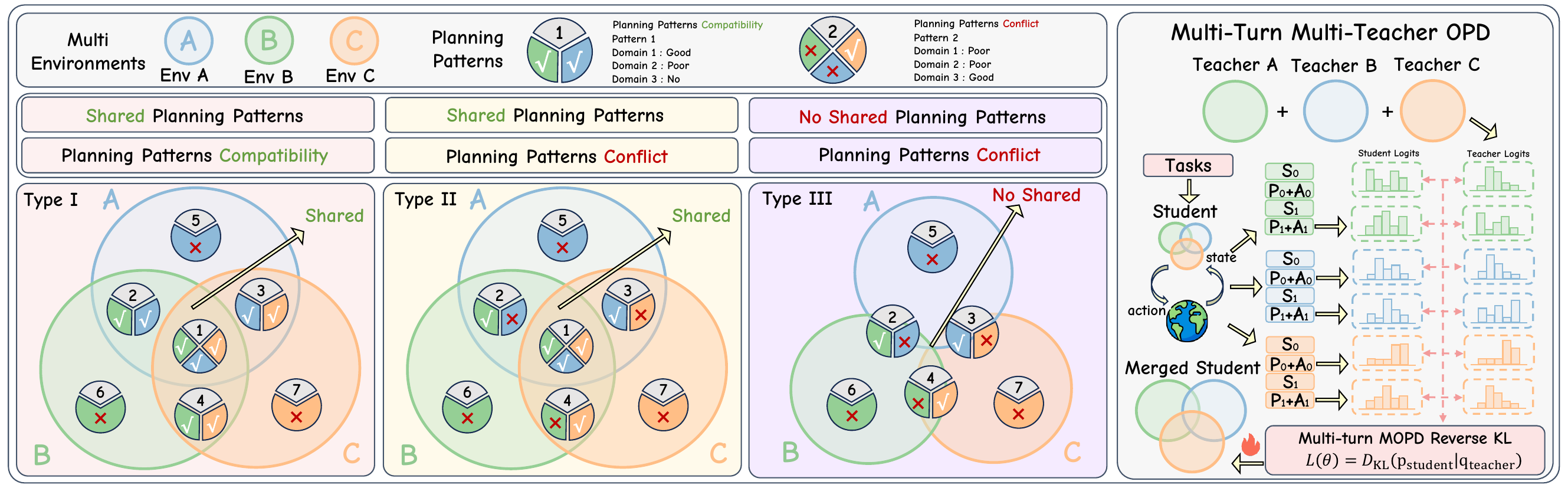}
    \caption{Multi-Turn MOPD. Analysis framework of non-shared and conflicting planning patterns in different environments for multi-teacher on-policy distillation.}
    \label{fig:MOPD} 
\end{figure}

To analyze the three problems above, we propose a unified framework. As shown in Figure~\ref{fig:MOPD}, we use two dimensions for our analysis: (1) whether there is a shared area of planning patterns, and (2) whether the planning patterns are compatible with different environments. Based on these two dimensions, we classify the relationship between planning patterns and environments into three typical coverage patterns: Type I (Shared and Compatibility Planning Patterns), Type II (Shared and Conflict Planning Patterns), and Type III (No Shared and Conflict Planning Patterns). These structural relationships determine how the planning patterns distribute between the student and multiple teachers across different environments.

\subsection{Shared and Compatibility Multi Teachers On-Policy Agentic Distillation}

To answer the \textbf{Q1}, we focus on the \textbf{Type I} category: \textbf{Shared and Compatibility Planning Patterns}. In this scenario, different environments share common planning patterns that are entirely compatible and free of cross-domain conflicts. This implies that the student and the multiple teachers possess an intersecting and consistent understanding of the tasks across different domains. To simulate this structural relationship and empirically evaluate how MOPD achieves effective cross-environment generalization under these conditions, we design the following experimental setup.

\begin{table}[!h]
\centering
\footnotesize
\setlength{\tabcolsep}{2.2pt}
\caption{Ensemble main table. Rows are grouped by pattern mixture. The teacher group reports Teacher FA, Teacher LF, and Teacher EA. The MOPD order is FA->LF->EA. Two global blocks are reported: avg@8 and pass@8. Domain abbreviations: FA = Fantasy Alchemy; LF = Livestock Farming; EA = Electronic Assembly.}
\label{tab:ensemble_main}
\resizebox{0.97\columnwidth}{!}{
\begin{tabular}{llccccccccccccc}
\toprule
\multirow{2}{*}{Pattern Mix} & \multirow{2}{*}{Model} & \multicolumn{4}{c}{Short} & \multicolumn{4}{c}{Mid} & \multicolumn{4}{c}{Long} & \multicolumn{1}{c}{Avg} \\
\cmidrule(lr){3-6} \cmidrule(lr){7-10} \cmidrule(lr){11-14} \cmidrule(lr){15-15}
 &  & FA & LF & EA & Mean & FA & LF & EA & Mean & FA & LF & EA & Mean & S/M/L Mean \\
\midrule
\rowcolor{EnsMainBlockHeader}
\multicolumn{15}{c}{\textbf{avg@8}} \\
\midrule
\multirow{3}{*}{Teacher} & Teacher FA & \cellcolor{EnsMainAvgCell!42}100.00 & \cellcolor{EnsMainAvgCell!0}0.00 & \cellcolor{EnsMainAvgCell!0}0.00 & \cellcolor{EnsMainAvgCell!14}33.33 & \cellcolor{EnsMainAvgCell!42}100.00 & \cellcolor{EnsMainAvgCell!0}0.00 & \cellcolor{EnsMainAvgCell!0}0.00 & \cellcolor{EnsMainAvgCell!14}33.33 & \cellcolor{EnsMainAvgCell!42}100.00 & \cellcolor{EnsMainAvgCell!0}0.00 & \cellcolor{EnsMainAvgCell!0}0.00 & \cellcolor{EnsMainAvgCell!14}33.33 & \cellcolor{EnsMainAvgCell!14}33.33 \\
 & Teacher LF & \cellcolor{EnsMainAvgCell!0}0.00 & \cellcolor{EnsMainAvgCell!42}100.00 & \cellcolor{EnsMainAvgCell!0}0.00 & \cellcolor{EnsMainAvgCell!14}33.33 & \cellcolor{EnsMainAvgCell!0}0.00 & \cellcolor{EnsMainAvgCell!42}100.00 & \cellcolor{EnsMainAvgCell!0}0.00 & \cellcolor{EnsMainAvgCell!14}33.33 & \cellcolor{EnsMainAvgCell!0}0.00 & \cellcolor{EnsMainAvgCell!42}100.00 & \cellcolor{EnsMainAvgCell!0}0.00 & \cellcolor{EnsMainAvgCell!14}33.33 & \cellcolor{EnsMainAvgCell!14}33.33 \\
 & Teacher EA & \cellcolor{EnsMainAvgCell!0}0.00 & \cellcolor{EnsMainAvgCell!0}0.00 & \cellcolor{EnsMainAvgCell!42}100.00 & \cellcolor{EnsMainAvgCell!14}33.33 & \cellcolor{EnsMainAvgCell!0}0.00 & \cellcolor{EnsMainAvgCell!0}0.00 & \cellcolor{EnsMainAvgCell!42}100.00 & \cellcolor{EnsMainAvgCell!14}33.33 & \cellcolor{EnsMainAvgCell!0}0.00 & \cellcolor{EnsMainAvgCell!0}0.00 & \cellcolor{EnsMainAvgCell!42}100.00 & \cellcolor{EnsMainAvgCell!14}33.33 & \cellcolor{EnsMainAvgCell!14}33.33 \\
\cmidrule(lr){1-15}
\multirow{4}{*}{4 Opt.:4 Sub.(a)} & Instruct & \cellcolor{EnsMainAvgCell!36}86.25 & \cellcolor{EnsMainAvgCell!34}80.70 & \cellcolor{EnsMainAvgCell!33}79.45 & \cellcolor{EnsMainAvgCell!34}82.14 & \cellcolor{EnsMainAvgCell!18}41.95 & \cellcolor{EnsMainAvgCell!15}34.92 & \cellcolor{EnsMainAvgCell!13}30.08 & \cellcolor{EnsMainAvgCell!15}35.65 & \cellcolor{EnsMainAvgCell!6}13.83 & \cellcolor{EnsMainAvgCell!4}8.83 & \cellcolor{EnsMainAvgCell!4}8.91 & \cellcolor{EnsMainAvgCell!4}10.52 & \cellcolor{EnsMainAvgCell!18}42.77 \\
 & After FA & \cellcolor{EnsMainAvgCell!39}92.50 & \cellcolor{EnsMainAvgCell!37}88.59 & \cellcolor{EnsMainAvgCell!36}86.41 & \cellcolor{EnsMainAvgCell!37}89.17 & \cellcolor{EnsMainAvgCell!26}61.64 & \cellcolor{EnsMainAvgCell!22}52.58 & \cellcolor{EnsMainAvgCell!20}48.59 & \cellcolor{EnsMainAvgCell!23}54.27 & \cellcolor{EnsMainAvgCell!13}31.64 & \cellcolor{EnsMainAvgCell!10}24.38 & \cellcolor{EnsMainAvgCell!9}22.19 & \cellcolor{EnsMainAvgCell!11}26.07 & \cellcolor{EnsMainAvgCell!24}56.50 \\
 & After LF & \cellcolor{EnsMainAvgCell!38}90.55 & \cellcolor{EnsMainAvgCell!39}92.81 & \cellcolor{EnsMainAvgCell!36}85.55 & \cellcolor{EnsMainAvgCell!38}89.64 & \cellcolor{EnsMainAvgCell!25}60.55 & \cellcolor{EnsMainAvgCell!25}58.36 & \cellcolor{EnsMainAvgCell!20}48.75 & \cellcolor{EnsMainAvgCell!23}55.89 & \cellcolor{EnsMainAvgCell!12}28.75 & \cellcolor{EnsMainAvgCell!12}27.66 & \cellcolor{EnsMainAvgCell!12}28.44 & \cellcolor{EnsMainAvgCell!12}28.28 & \cellcolor{EnsMainAvgCell!24}57.93 \\
 & After EA & \cellcolor{EnsMainAvgCell!37}88.59 & \cellcolor{EnsMainAvgCell!39}92.73 & \cellcolor{EnsMainAvgCell!36}85.08 & \cellcolor{EnsMainAvgCell!37}88.80 & \cellcolor{EnsMainAvgCell!22}51.95 & \cellcolor{EnsMainAvgCell!20}47.42 & \cellcolor{EnsMainAvgCell!20}47.81 & \cellcolor{EnsMainAvgCell!21}49.06 & \cellcolor{EnsMainAvgCell!9}22.50 & \cellcolor{EnsMainAvgCell!8}18.83 & \cellcolor{EnsMainAvgCell!12}27.81 & \cellcolor{EnsMainAvgCell!10}23.05 & \cellcolor{EnsMainAvgCell!23}53.64 \\
\cmidrule(lr){1-15}
\multirow{4}{*}{4 Opt.:4 Sub.(b)} & Instruct & \cellcolor{EnsMainAvgCell!32}77.19 & \cellcolor{EnsMainAvgCell!26}62.66 & \cellcolor{EnsMainAvgCell!29}68.44 & \cellcolor{EnsMainAvgCell!29}69.43 & \cellcolor{EnsMainAvgCell!6}14.38 & \cellcolor{EnsMainAvgCell!4}9.53 & \cellcolor{EnsMainAvgCell!2}5.23 & \cellcolor{EnsMainAvgCell!4}9.71 & \cellcolor{EnsMainAvgCell!0}1.17 & \cellcolor{EnsMainAvgCell!0}0.23 & \cellcolor{EnsMainAvgCell!0}0.70 & \cellcolor{EnsMainAvgCell!0}0.70 & \cellcolor{EnsMainAvgCell!11}26.61 \\
 & After FA & \cellcolor{EnsMainAvgCell!38}91.02 & \cellcolor{EnsMainAvgCell!31}73.59 & \cellcolor{EnsMainAvgCell!34}81.25 & \cellcolor{EnsMainAvgCell!34}81.95 & \cellcolor{EnsMainAvgCell!19}45.08 & \cellcolor{EnsMainAvgCell!16}36.95 & \cellcolor{EnsMainAvgCell!11}25.16 & \cellcolor{EnsMainAvgCell!15}35.73 & \cellcolor{EnsMainAvgCell!6}15.47 & \cellcolor{EnsMainAvgCell!2}5.39 & \cellcolor{EnsMainAvgCell!5}12.27 & \cellcolor{EnsMainAvgCell!5}11.04 & \cellcolor{EnsMainAvgCell!18}42.91 \\
 & After LF & \cellcolor{EnsMainAvgCell!38}90.00 & \cellcolor{EnsMainAvgCell!31}73.05 & \cellcolor{EnsMainAvgCell!34}80.62 & \cellcolor{EnsMainAvgCell!34}81.22 & \cellcolor{EnsMainAvgCell!19}45.16 & \cellcolor{EnsMainAvgCell!14}33.36 & \cellcolor{EnsMainAvgCell!11}26.02 & \cellcolor{EnsMainAvgCell!15}34.84 & \cellcolor{EnsMainAvgCell!6}15.23 & \cellcolor{EnsMainAvgCell!3}6.88 & \cellcolor{EnsMainAvgCell!6}13.44 & \cellcolor{EnsMainAvgCell!5}11.85 & \cellcolor{EnsMainAvgCell!18}42.64 \\
 & After EA & \cellcolor{EnsMainAvgCell!36}85.00 & \cellcolor{EnsMainAvgCell!29}69.69 & \cellcolor{EnsMainAvgCell!33}79.53 & \cellcolor{EnsMainAvgCell!33}78.07 & \cellcolor{EnsMainAvgCell!16}37.42 & \cellcolor{EnsMainAvgCell!13}30.62 & \cellcolor{EnsMainAvgCell!9}21.56 & \cellcolor{EnsMainAvgCell!13}29.87 & \cellcolor{EnsMainAvgCell!4}10.00 & \cellcolor{EnsMainAvgCell!3}6.33 & \cellcolor{EnsMainAvgCell!5}13.05 & \cellcolor{EnsMainAvgCell!4}9.79 & \cellcolor{EnsMainAvgCell!16}39.24 \\
\cmidrule(lr){1-15}
\multirow{4}{*}{4 Opt.:8 Sub.} & Instruct & \cellcolor{EnsMainAvgCell!22}51.64 & \cellcolor{EnsMainAvgCell!17}40.55 & \cellcolor{EnsMainAvgCell!18}42.27 & \cellcolor{EnsMainAvgCell!19}44.82 & \cellcolor{EnsMainAvgCell!0}0.00 & \cellcolor{EnsMainAvgCell!0}0.08 & \cellcolor{EnsMainAvgCell!0}0.00 & \cellcolor{EnsMainAvgCell!0}0.03 & \cellcolor{EnsMainAvgCell!0}0.00 & \cellcolor{EnsMainAvgCell!0}0.00 & \cellcolor{EnsMainAvgCell!0}0.00 & \cellcolor{EnsMainAvgCell!0}0.00 & \cellcolor{EnsMainAvgCell!6}14.95 \\
 & After FA & \cellcolor{EnsMainAvgCell!35}84.06 & \cellcolor{EnsMainAvgCell!26}62.58 & \cellcolor{EnsMainAvgCell!30}72.58 & \cellcolor{EnsMainAvgCell!31}73.07 & \cellcolor{EnsMainAvgCell!13}30.23 & \cellcolor{EnsMainAvgCell!9}21.72 & \cellcolor{EnsMainAvgCell!7}17.03 & \cellcolor{EnsMainAvgCell!10}22.99 & \cellcolor{EnsMainAvgCell!3}7.66 & \cellcolor{EnsMainAvgCell!2}3.67 & \cellcolor{EnsMainAvgCell!2}5.16 & \cellcolor{EnsMainAvgCell!2}5.49 & \cellcolor{EnsMainAvgCell!14}33.85 \\
 & After LF & \cellcolor{EnsMainAvgCell!35}83.83 & \cellcolor{EnsMainAvgCell!26}61.09 & \cellcolor{EnsMainAvgCell!30}71.48 & \cellcolor{EnsMainAvgCell!30}72.14 & \cellcolor{EnsMainAvgCell!12}28.83 & \cellcolor{EnsMainAvgCell!9}21.17 & \cellcolor{EnsMainAvgCell!8}18.12 & \cellcolor{EnsMainAvgCell!10}22.71 & \cellcolor{EnsMainAvgCell!3}7.81 & \cellcolor{EnsMainAvgCell!1}3.36 & \cellcolor{EnsMainAvgCell!2}5.00 & \cellcolor{EnsMainAvgCell!2}5.39 & \cellcolor{EnsMainAvgCell!14}33.41 \\
 & After EA & \cellcolor{EnsMainAvgCell!34}81.80 & \cellcolor{EnsMainAvgCell!25}60.55 & \cellcolor{EnsMainAvgCell!28}66.56 & \cellcolor{EnsMainAvgCell!29}69.64 & \cellcolor{EnsMainAvgCell!11}26.33 & \cellcolor{EnsMainAvgCell!7}16.88 & \cellcolor{EnsMainAvgCell!7}15.55 & \cellcolor{EnsMainAvgCell!8}19.58 & \cellcolor{EnsMainAvgCell!3}6.41 & \cellcolor{EnsMainAvgCell!1}1.88 & \cellcolor{EnsMainAvgCell!1}3.05 & \cellcolor{EnsMainAvgCell!2}3.78 & \cellcolor{EnsMainAvgCell!13}31.00 \\
\midrule
\rowcolor{EnsMainBlockHeader}
\multicolumn{15}{c}{\textbf{pass@8}} \\
\midrule
\multirow{3}{*}{Teacher} & Teacher FA & \cellcolor{EnsMainPassCell!42}100.00 & \cellcolor{EnsMainPassCell!0}0.00 & \cellcolor{EnsMainPassCell!0}0.00 & \cellcolor{EnsMainPassCell!14}33.33 & \cellcolor{EnsMainPassCell!42}100.00 & \cellcolor{EnsMainPassCell!0}0.00 & \cellcolor{EnsMainPassCell!0}0.00 & \cellcolor{EnsMainPassCell!14}33.33 & \cellcolor{EnsMainPassCell!42}100.00 & \cellcolor{EnsMainPassCell!0}0.00 & \cellcolor{EnsMainPassCell!0}0.00 & \cellcolor{EnsMainPassCell!14}33.33 & \cellcolor{EnsMainPassCell!14}33.33 \\
 & Teacher LF & \cellcolor{EnsMainPassCell!0}0.00 & \cellcolor{EnsMainPassCell!42}100.00 & \cellcolor{EnsMainPassCell!0}0.00 & \cellcolor{EnsMainPassCell!14}33.33 & \cellcolor{EnsMainPassCell!0}0.00 & \cellcolor{EnsMainPassCell!42}100.00 & \cellcolor{EnsMainPassCell!0}0.00 & \cellcolor{EnsMainPassCell!14}33.33 & \cellcolor{EnsMainPassCell!0}0.00 & \cellcolor{EnsMainPassCell!42}100.00 & \cellcolor{EnsMainPassCell!0}0.00 & \cellcolor{EnsMainPassCell!14}33.33 & \cellcolor{EnsMainPassCell!14}33.33 \\
 & Teacher EA & \cellcolor{EnsMainPassCell!0}0.00 & \cellcolor{EnsMainPassCell!0}0.00 & \cellcolor{EnsMainPassCell!42}100.00 & \cellcolor{EnsMainPassCell!14}33.33 & \cellcolor{EnsMainPassCell!0}0.00 & \cellcolor{EnsMainPassCell!0}0.00 & \cellcolor{EnsMainPassCell!42}100.00 & \cellcolor{EnsMainPassCell!14}33.33 & \cellcolor{EnsMainPassCell!0}0.00 & \cellcolor{EnsMainPassCell!0}0.00 & \cellcolor{EnsMainPassCell!42}100.00 & \cellcolor{EnsMainPassCell!14}33.33 & \cellcolor{EnsMainPassCell!14}33.33 \\
\cmidrule(lr){1-15}
\multirow{4}{*}{4 Opt.:4 Sub.(a)} & Instruct & \cellcolor{EnsMainPassCell!41}98.12 & \cellcolor{EnsMainPassCell!40}96.25 & \cellcolor{EnsMainPassCell!39}92.50 & \cellcolor{EnsMainPassCell!40}95.62 & \cellcolor{EnsMainPassCell!32}75.62 & \cellcolor{EnsMainPassCell!28}66.88 & \cellcolor{EnsMainPassCell!27}63.12 & \cellcolor{EnsMainPassCell!29}68.54 & \cellcolor{EnsMainPassCell!17}40.62 & \cellcolor{EnsMainPassCell!14}32.50 & \cellcolor{EnsMainPassCell!13}30.62 & \cellcolor{EnsMainPassCell!15}34.58 & \cellcolor{EnsMainPassCell!28}66.25 \\
 & After FA & \cellcolor{EnsMainPassCell!40}95.00 & \cellcolor{EnsMainPassCell!40}95.62 & \cellcolor{EnsMainPassCell!39}93.12 & \cellcolor{EnsMainPassCell!40}94.58 & \cellcolor{EnsMainPassCell!29}68.12 & \cellcolor{EnsMainPassCell!28}66.88 & \cellcolor{EnsMainPassCell!28}65.62 & \cellcolor{EnsMainPassCell!28}66.88 & \cellcolor{EnsMainPassCell!17}41.25 & \cellcolor{EnsMainPassCell!17}40.62 & \cellcolor{EnsMainPassCell!17}40.62 & \cellcolor{EnsMainPassCell!17}40.83 & \cellcolor{EnsMainPassCell!28}67.43 \\
 & After LF & \cellcolor{EnsMainPassCell!40}95.62 & \cellcolor{EnsMainPassCell!40}95.00 & \cellcolor{EnsMainPassCell!38}90.00 & \cellcolor{EnsMainPassCell!39}93.54 & \cellcolor{EnsMainPassCell!29}68.75 & \cellcolor{EnsMainPassCell!28}65.62 & \cellcolor{EnsMainPassCell!24}56.88 & \cellcolor{EnsMainPassCell!27}63.75 & \cellcolor{EnsMainPassCell!17}39.38 & \cellcolor{EnsMainPassCell!16}38.12 & \cellcolor{EnsMainPassCell!15}36.25 & \cellcolor{EnsMainPassCell!16}37.92 & \cellcolor{EnsMainPassCell!27}65.07 \\
 & After EA & \cellcolor{EnsMainPassCell!39}91.88 & \cellcolor{EnsMainPassCell!40}95.62 & \cellcolor{EnsMainPassCell!38}90.62 & \cellcolor{EnsMainPassCell!39}92.71 & \cellcolor{EnsMainPassCell!25}60.62 & \cellcolor{EnsMainPassCell!25}58.75 & \cellcolor{EnsMainPassCell!24}58.12 & \cellcolor{EnsMainPassCell!25}59.17 & \cellcolor{EnsMainPassCell!14}32.50 & \cellcolor{EnsMainPassCell!12}28.75 & \cellcolor{EnsMainPassCell!18}41.88 & \cellcolor{EnsMainPassCell!14}34.38 & \cellcolor{EnsMainPassCell!26}62.08 \\
\cmidrule(lr){1-15}
\multirow{4}{*}{4 Opt.:4 Sub.(b)} & Instruct & \cellcolor{EnsMainPassCell!40}95.62 & \cellcolor{EnsMainPassCell!36}86.88 & \cellcolor{EnsMainPassCell!38}89.38 & \cellcolor{EnsMainPassCell!38}90.62 & \cellcolor{EnsMainPassCell!22}51.25 & \cellcolor{EnsMainPassCell!16}38.12 & \cellcolor{EnsMainPassCell!12}28.12 & \cellcolor{EnsMainPassCell!16}39.17 & \cellcolor{EnsMainPassCell!3}8.12 & \cellcolor{EnsMainPassCell!1}1.88 & \cellcolor{EnsMainPassCell!2}5.00 & \cellcolor{EnsMainPassCell!2}5.00 & \cellcolor{EnsMainPassCell!19}44.93 \\
 & After FA & \cellcolor{EnsMainPassCell!39}91.88 & \cellcolor{EnsMainPassCell!34}80.00 & \cellcolor{EnsMainPassCell!36}85.62 & \cellcolor{EnsMainPassCell!36}85.83 & \cellcolor{EnsMainPassCell!21}50.62 & \cellcolor{EnsMainPassCell!19}44.38 & \cellcolor{EnsMainPassCell!14}33.75 & \cellcolor{EnsMainPassCell!18}42.92 & \cellcolor{EnsMainPassCell!9}21.25 & \cellcolor{EnsMainPassCell!4}10.00 & \cellcolor{EnsMainPassCell!8}18.75 & \cellcolor{EnsMainPassCell!7}16.67 & \cellcolor{EnsMainPassCell!20}48.47 \\
 & After LF & \cellcolor{EnsMainPassCell!39}91.88 & \cellcolor{EnsMainPassCell!33}78.75 & \cellcolor{EnsMainPassCell!37}88.12 & \cellcolor{EnsMainPassCell!36}86.25 & \cellcolor{EnsMainPassCell!22}51.88 & \cellcolor{EnsMainPassCell!18}41.88 & \cellcolor{EnsMainPassCell!15}36.25 & \cellcolor{EnsMainPassCell!18}43.33 & \cellcolor{EnsMainPassCell!9}21.88 & \cellcolor{EnsMainPassCell!4}10.00 & \cellcolor{EnsMainPassCell!9}21.25 & \cellcolor{EnsMainPassCell!7}17.71 & \cellcolor{EnsMainPassCell!21}49.10 \\
 & After EA & \cellcolor{EnsMainPassCell!37}88.12 & \cellcolor{EnsMainPassCell!32}76.25 & \cellcolor{EnsMainPassCell!36}85.00 & \cellcolor{EnsMainPassCell!35}83.12 & \cellcolor{EnsMainPassCell!19}45.00 & \cellcolor{EnsMainPassCell!17}40.00 & \cellcolor{EnsMainPassCell!14}32.50 & \cellcolor{EnsMainPassCell!16}39.17 & \cellcolor{EnsMainPassCell!7}16.25 & \cellcolor{EnsMainPassCell!4}8.75 & \cellcolor{EnsMainPassCell!9}20.62 & \cellcolor{EnsMainPassCell!6}15.21 & \cellcolor{EnsMainPassCell!19}45.83 \\
\cmidrule(lr){1-15}
\multirow{4}{*}{4 Opt.:8 Sub.} & Instruct & \cellcolor{EnsMainPassCell!30}70.62 & \cellcolor{EnsMainPassCell!26}61.25 & \cellcolor{EnsMainPassCell!28}67.50 & \cellcolor{EnsMainPassCell!28}66.46 & \cellcolor{EnsMainPassCell!0}0.00 & \cellcolor{EnsMainPassCell!0}0.62 & \cellcolor{EnsMainPassCell!0}0.00 & \cellcolor{EnsMainPassCell!0}0.21 & \cellcolor{EnsMainPassCell!0}0.00 & \cellcolor{EnsMainPassCell!0}0.00 & \cellcolor{EnsMainPassCell!0}0.00 & \cellcolor{EnsMainPassCell!0}0.00 & \cellcolor{EnsMainPassCell!9}22.22 \\
 & After FA & \cellcolor{EnsMainPassCell!38}89.38 & \cellcolor{EnsMainPassCell!29}68.75 & \cellcolor{EnsMainPassCell!33}78.75 & \cellcolor{EnsMainPassCell!33}78.96 & \cellcolor{EnsMainPassCell!16}37.50 & \cellcolor{EnsMainPassCell!12}29.38 & \cellcolor{EnsMainPassCell!11}26.88 & \cellcolor{EnsMainPassCell!13}31.25 & \cellcolor{EnsMainPassCell!5}12.50 & \cellcolor{EnsMainPassCell!2}5.00 & \cellcolor{EnsMainPassCell!3}7.50 & \cellcolor{EnsMainPassCell!4}8.33 & \cellcolor{EnsMainPassCell!17}39.51 \\
 & After LF & \cellcolor{EnsMainPassCell!37}88.12 & \cellcolor{EnsMainPassCell!29}68.75 & \cellcolor{EnsMainPassCell!33}78.75 & \cellcolor{EnsMainPassCell!33}78.54 & \cellcolor{EnsMainPassCell!14}34.38 & \cellcolor{EnsMainPassCell!12}28.75 & \cellcolor{EnsMainPassCell!11}26.25 & \cellcolor{EnsMainPassCell!13}29.79 & \cellcolor{EnsMainPassCell!4}10.62 & \cellcolor{EnsMainPassCell!3}6.88 & \cellcolor{EnsMainPassCell!3}8.12 & \cellcolor{EnsMainPassCell!4}8.54 & \cellcolor{EnsMainPassCell!16}38.96 \\
 & After EA & \cellcolor{EnsMainPassCell!37}87.50 & \cellcolor{EnsMainPassCell!28}66.88 & \cellcolor{EnsMainPassCell!31}73.12 & \cellcolor{EnsMainPassCell!32}75.83 & \cellcolor{EnsMainPassCell!15}35.00 & \cellcolor{EnsMainPassCell!11}25.62 & \cellcolor{EnsMainPassCell!9}21.88 & \cellcolor{EnsMainPassCell!12}27.50 & \cellcolor{EnsMainPassCell!4}10.00 & \cellcolor{EnsMainPassCell!2}4.38 & \cellcolor{EnsMainPassCell!3}6.25 & \cellcolor{EnsMainPassCell!3}6.88 & \cellcolor{EnsMainPassCell!15}36.74 \\
\bottomrule
\end{tabular}
}
\end{table}

\paragraph{Task Setup.}
(1) \textbf{Teacher models.} The experiment includes three domains: fantasy alchemy, livestock farming, and electronic assembly. We train an expert teacher model for each domain, so we get three domain expert teachers in total. Each teacher model learns from the gold trajectories in the pretraining and posttraining datasets of its domain. To bring in different planning supervision signals, we design four planning patterns for internal world models. Three expert teachers share one planning template. We assign the other templates to only some teachers. We distribute three planning patterns equally for each teacher. Therefore, all three teachers learn the correct trajectory supervision, but they see different planning template distributions and have varied planning patterns.
(2) \textbf{Student models.} The student models use 4 Opt.:4 Sub. and 4 Opt.:8 Sub. as base models. The 4 Opt.:4 Sub. setting has two versions: 4 Opt.:4 Sub.(a) and 4 Opt.:4 Sub.(b). For the student SFT data, we build two group settings on the same three domains: electronic assembly, fantasy alchemy, and livestock farming. Both settings build on pretraining trajectories and include two kinds of supervision signals: normal planning samples from gold actions, and shortcut style distractor samples with wrong actions. Both settings use four kinds of normal planning templates and four kinds of distractor planning templates. We make the sample size of each template type equal to keep an overall balance between normal samples and distractor samples. The main difference between the two settings is the domain and template distribution of normal planning samples. Form a uses a mixed distribution: the three domains share one normal planning template, and the other normal planning templates only appear in two domains. Therefore, the subsets of planning patterns seen by different domains are not exactly the same. Form b removes this difference: every normal planning template appears in all three domains with equal sampling. Thus, Form a focuses on the unequal planning pattern exposure among domain experts. Form b makes sure each domain covers all planning templates, while keeping the same domain scope and equal distractor construction.
(3) \textbf{Training configurations.} To ensure a strict comparison, the three student models use the exact same amount of pretraining data and the same training parameters. The student models go through OPD one by one in the order of FA, LF, and EA. We train the OPD on the posttraining dataset with a middle difficulty level.

\paragraph{Table Results.} As shown in Table~\ref{tab:ensemble_main}. 
(1) \textbf{Shared and compatible planning patterns are a basic condition for cross environment generalization.} It shows that although teacher models only work in their own domains (for example, Teacher FA has an Avg@8 of 100.00 in the FA domain but 0.00 in LF and EA), student models achieve significant performance improvements in unseen domains after distillation from a single domain teacher. Take 4 Opt.:4 Sub.(a) as an example. In Middle difficulty, for in-domain environments, the avg@8 of the base Instruct model in the FA domain increases from 41.95 to 61.64. For out of domain environments, the avg@8 of the base Instruct model in the LF domain is 34.92. However, after it receives distillation only from the FA domain teacher, its performance in the LF domain jumps directly to 52.58. This indicates that when different environments shared and compatible planning patterns, MOPD breaks domain barriers and allows the model to transfer abilities learned in one environment to other environments.
(2) \textbf{Cross environment generalization does not directly depend on the percentage of optimal actions in the data.} We compare the 4 Opt.:4 Sub. model with the 4 Opt.:8 Sub. model that introduces more interfering options. The Instruct model of the latter has very low base performance because a large number of low quality trajectories affect it. Its average avg@8 in Middle difficulty is only 0.03. However, after MOPD training, such as in the After FA setting, the performance of the 4 Opt.:8 Sub. model recovers greatly in all domains. Specifically, its average avg@8 in Middle difficulty successfully increases to 22.99. This shows that as long as shared planning patterns exist at the bottom level of the model, MOPD effectively extracts them and achieves cross environment generalization. This is true even if a large amount of low quality data covers these high quality patterns during the SFT stage. The generalization effect does not strictly depend on the initial percentage of high quality data.

\begin{figure}[!h]
    \centering
    \includegraphics[width=\textwidth]{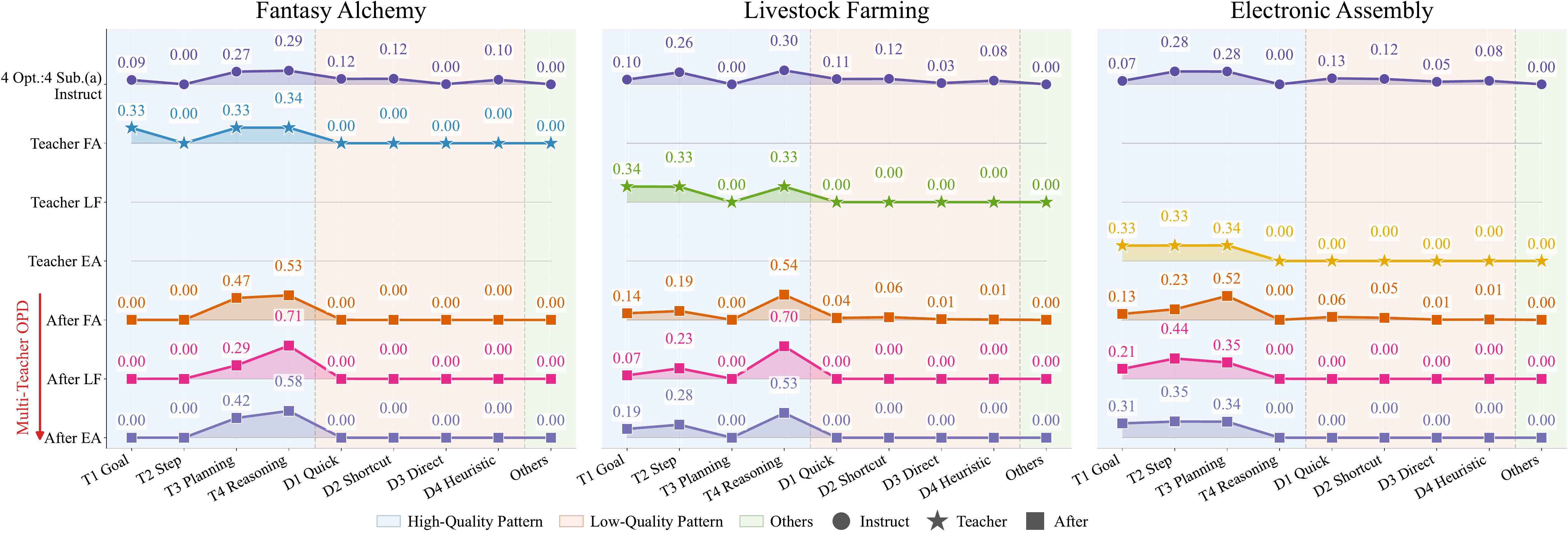}
    \caption{Planning pattern distributions for 4 Opt.:4 Sub. (a) before and after multi-teacher OPD.}
    \label{fig:part_3_general_pattern_44a_second_token_ridgeline} 
\end{figure}

\begin{figure}[!h]
    \centering
    \includegraphics[width=\textwidth]{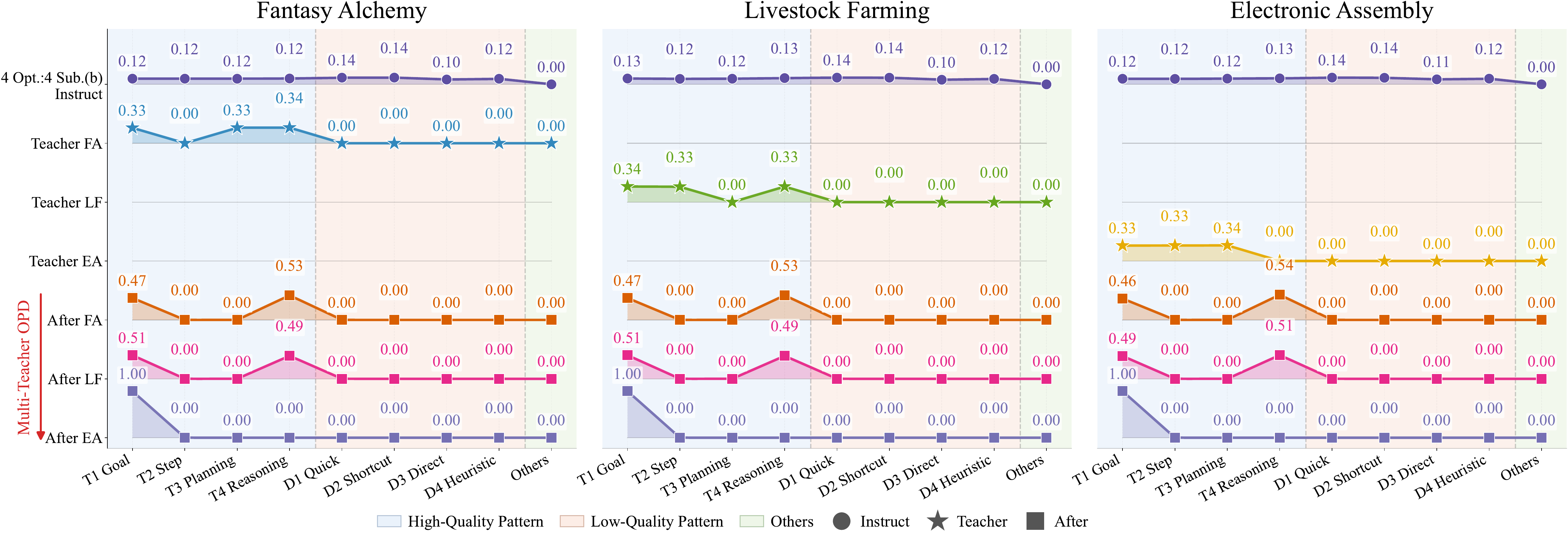}
    \caption{Planning pattern distributions for 4 Opt.:4 Sub. (b) before and after multi-teacher OPD.}
    \label{fig:part_3_general_pattern_44b_second_token_ridgeline} 
\end{figure}

\begin{figure}[!h]
    \centering
    \includegraphics[width=\textwidth]{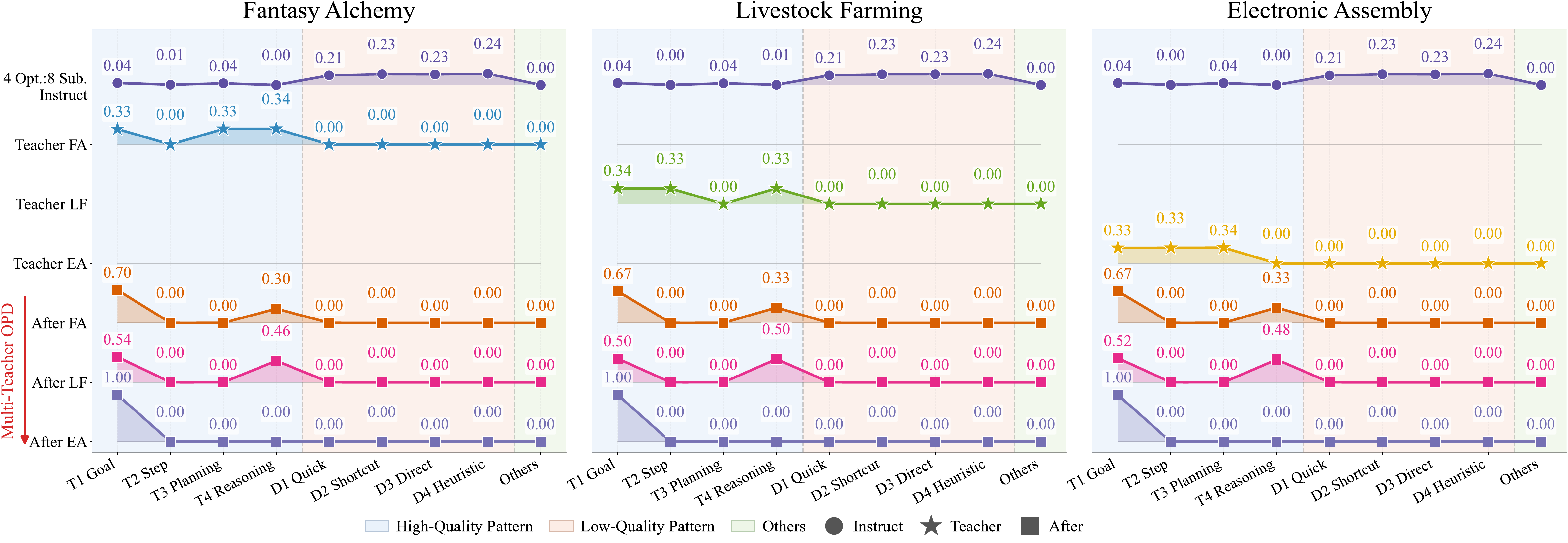}
    \caption{Planning pattern distributions for 4 Opt.:8 Sub. before and after multi-teacher OPD.}
    \label{fig:part_3_general_pattern_48_second_token_ridgeline} 
\end{figure}

\paragraph{Figure Results.} As shown in Figure~\ref{fig:part_3_general_pattern_44a_second_token_ridgeline}, Figure~\ref{fig:part_3_general_pattern_44b_second_token_ridgeline}, and Figure~\ref{fig:part_3_general_pattern_48_second_token_ridgeline}. 
(1) \textbf{Consistent with previous studies, the process is mode seeking rather than mode covering.} As the distribution change graph in Figure~\ref{fig:part_3_general_pattern_44b_second_token_ridgeline} shows, the student model in the Instruct stage presents a relatively uniform distribution across various high quality planning patterns. However, after MOPD training, the probability distribution of the model does not cover all teacher preferences. Instead, it quickly shrinks to a portion of the shared patterns. For example, in Figure~\ref{fig:part_3_general_pattern_44b_second_token_ridgeline}, the trained model strongly focuses on the "T1 Goal" and "T4 Reasoning" patterns in all three environments, and it does not cover the T3 pattern of the teacher. This matches the mode seeking phenomenon of the reverse KL divergence in existing OPD.
(2) \textbf{MOPD is to capture the common distribution among multiple teachers as much as possible.} As shown in Figure~\ref{fig:part_3_general_pattern_44b_second_token_ridgeline} and Figure~\ref{fig:part_3_general_pattern_48_second_token_ridgeline}. Sequential distillation drives a consistent shift in the planning patterns of the student model across all domains. For example, when the student model completes the distillation from the last teacher (EA), its planning pattern distributions in Fantasy Alchemy, Livestock Farming, and Electronic Assembly all converge to the single "T1 Goal" template with a ratio of 1.00 at the same time. This shows that sequential MOPD does not cause separate local adaptations in different domains. Instead, it drives the model to find and lock onto a globally unified common distribution among the teacher strategies of each domain.
(3) \textbf{Training on one environment affects the planning patterns of other environments that share the same distribution.} As shown in Figure~\ref{fig:part_3_general_pattern_44b_second_token_ridgeline}, the experimental results in the first two columns reveal that the planning pattern trained in a single environment shows the same changes in other environments. For example, combining the data in the After FA row, when the model finishes training in the Fantasy Alchemy environment, the percentage of the T2 Step pattern in both the local FA environment and the unseen Livestock Farming (LF) environment shows the exact same decline. Specifically, both values drop from the baseline of 0.12 to 0.00. At the same time, the T4 Reasoning pattern in these two environments activates at the same time across domains, and both accurately reach 0.53. This indicates that as long as different environments share a consistent distribution basis in their underlying task structures, optimizing for a single environment directly causes cross domain generalization for the planning patterns in other environments.
(4) \textbf{However, MOPD fails to trigger planning patterns that do not exist in the model's pre-training data for other environments.} This limitation is very clear in the T3 Planning changes in Figure~\ref{fig:part_3_general_pattern_44a_second_token_ridgeline}. As the results in the After FA row show, when we introduce specific optimization in the Fantasy Alchemy environment, the model successfully increases the percentage of the T3 Planning pattern in the local FA environment from 0.27 in the Instruct stage to 0.47. However, when we apply this successful training signal from the FA environment to the Livestock Farming (LF) environment, it fails to trigger cross environment generalization for this planning pattern. This is because this planning pattern does not exist in the pretraining data of Livestock Farming. This proves that OPD cross environment generalization cannot artificially create or trigger specific planning patterns that are missing in the pretraining data of the target environment.

\begin{guidancebox}{Takeaway}

\textbf{Shared and compatible planning patterns enable cross-environment generalization.} When multiple environments share compatible planning patterns, MOPD can transfer the capability learned from one domain teacher to other domains, even if each teacher itself is only effective in its own environment.

\textbf{MOPD performs mode seeking over the shared teacher distribution.} Sequential multi-teacher OPD does not simply cover all planning patterns from all teachers. Instead, it tends to collapse toward a subset of shared high-quality planning modes, revealing that the core mechanism of MOPD is to identify and amplify the shared distribution among teachers.

\textbf{Cross-environment generalization is bounded by the student's existing pattern support.} MOPD can activate and strengthen planning patterns that already exist in the student's underlying distribution, even when they are diluted by low-quality SFT data. However, it cannot reliably create planning patterns that are absent from the target environment's pretraining or SFT support.

\end{guidancebox}

\subsection{Non-Shared Conflicting Multi Teachers On-Policy Agentic Distillation}

\begin{table}[!h]
\centering
\footnotesize
\setlength{\tabcolsep}{2.2pt}
\caption{Ensemble main table for No Shared Planning Patterns and Planning Patterns Conflict.}
\label{tab:ensemble_main_2}
\resizebox{0.97\columnwidth}{!}{
\begin{tabular}{llccccccccccccc}
\toprule
\multirow{2}{*}{Pattern Mix} & \multirow{2}{*}{Model} & \multicolumn{4}{c}{Short} & \multicolumn{4}{c}{Mid} & \multicolumn{4}{c}{Long} & \multicolumn{1}{c}{Avg} \\
\cmidrule(lr){3-6} \cmidrule(lr){7-10} \cmidrule(lr){11-14} \cmidrule(lr){15-15}
 &  & FA & LF & EA & Mean & FA & LF & EA & Mean & FA & LF & EA & Mean & S/M/L Mean \\
\midrule
\rowcolor{EnsMainTwoBlockHeader}
\multicolumn{15}{c}{\textbf{avg@8}} \\
\midrule
\multirow{3}{*}{Teacher} & Teacher FA & \cellcolor{EnsMainTwoAvgCell!42}100.00 & \cellcolor{EnsMainTwoAvgCell!0}0.00 & \cellcolor{EnsMainTwoAvgCell!0}0.00 & \cellcolor{EnsMainTwoAvgCell!14}33.33 & \cellcolor{EnsMainTwoAvgCell!42}100.00 & \cellcolor{EnsMainTwoAvgCell!0}0.00 & \cellcolor{EnsMainTwoAvgCell!0}0.00 & \cellcolor{EnsMainTwoAvgCell!14}33.33 & \cellcolor{EnsMainTwoAvgCell!42}100.00 & \cellcolor{EnsMainTwoAvgCell!0}0.00 & \cellcolor{EnsMainTwoAvgCell!0}0.00 & \cellcolor{EnsMainTwoAvgCell!14}33.33 & \cellcolor{EnsMainTwoAvgCell!14}33.33 \\
 & Teacher LF & \cellcolor{EnsMainTwoAvgCell!0}0.00 & \cellcolor{EnsMainTwoAvgCell!42}100.00 & \cellcolor{EnsMainTwoAvgCell!0}0.00 & \cellcolor{EnsMainTwoAvgCell!14}33.33 & \cellcolor{EnsMainTwoAvgCell!0}0.00 & \cellcolor{EnsMainTwoAvgCell!42}100.00 & \cellcolor{EnsMainTwoAvgCell!0}0.00 & \cellcolor{EnsMainTwoAvgCell!14}33.33 & \cellcolor{EnsMainTwoAvgCell!0}0.00 & \cellcolor{EnsMainTwoAvgCell!42}100.00 & \cellcolor{EnsMainTwoAvgCell!0}0.00 & \cellcolor{EnsMainTwoAvgCell!14}33.33 & \cellcolor{EnsMainTwoAvgCell!14}33.33 \\
 & Teacher EA & \cellcolor{EnsMainTwoAvgCell!0}0.00 & \cellcolor{EnsMainTwoAvgCell!0}0.00 & \cellcolor{EnsMainTwoAvgCell!42}100.00 & \cellcolor{EnsMainTwoAvgCell!14}33.33 & \cellcolor{EnsMainTwoAvgCell!0}0.00 & \cellcolor{EnsMainTwoAvgCell!0}0.00 & \cellcolor{EnsMainTwoAvgCell!42}100.00 & \cellcolor{EnsMainTwoAvgCell!14}33.33 & \cellcolor{EnsMainTwoAvgCell!0}0.00 & \cellcolor{EnsMainTwoAvgCell!0}0.00 & \cellcolor{EnsMainTwoAvgCell!42}100.00 & \cellcolor{EnsMainTwoAvgCell!14}33.33 & \cellcolor{EnsMainTwoAvgCell!14}33.33 \\
\cmidrule(lr){1-15}
\multirow{4}{*}{4 Opt.:4 Sub.(c)} & Instruct & \cellcolor{EnsMainTwoAvgCell!23}54.77 & \cellcolor{EnsMainTwoAvgCell!21}49.53 & \cellcolor{EnsMainTwoAvgCell!23}55.31 & \cellcolor{EnsMainTwoAvgCell!22}53.20 & \cellcolor{EnsMainTwoAvgCell!0}0.31 & \cellcolor{EnsMainTwoAvgCell!0}0.23 & \cellcolor{EnsMainTwoAvgCell!0}0.62 & \cellcolor{EnsMainTwoAvgCell!0}0.39 & \cellcolor{EnsMainTwoAvgCell!0}0.00 & \cellcolor{EnsMainTwoAvgCell!0}0.00 & \cellcolor{EnsMainTwoAvgCell!0}0.00 & \cellcolor{EnsMainTwoAvgCell!0}0.00 & \cellcolor{EnsMainTwoAvgCell!8}17.86 \\
 & After FA & \cellcolor{EnsMainTwoAvgCell!40}95.23 & \cellcolor{EnsMainTwoAvgCell!27}64.61 & \cellcolor{EnsMainTwoAvgCell!15}35.47 & \cellcolor{EnsMainTwoAvgCell!27}65.10 & \cellcolor{EnsMainTwoAvgCell!30}70.62 & \cellcolor{EnsMainTwoAvgCell!1}2.89 & \cellcolor{EnsMainTwoAvgCell!0}0.00 & \cellcolor{EnsMainTwoAvgCell!10}24.51 & \cellcolor{EnsMainTwoAvgCell!23}55.16 & \cellcolor{EnsMainTwoAvgCell!0}0.00 & \cellcolor{EnsMainTwoAvgCell!0}0.00 & \cellcolor{EnsMainTwoAvgCell!8}18.39 & \cellcolor{EnsMainTwoAvgCell!15}36.00 \\
 & After LF & \cellcolor{EnsMainTwoAvgCell!39}93.75 & \cellcolor{EnsMainTwoAvgCell!37}88.28 & \cellcolor{EnsMainTwoAvgCell!8}19.14 & \cellcolor{EnsMainTwoAvgCell!28}67.06 & \cellcolor{EnsMainTwoAvgCell!27}63.28 & \cellcolor{EnsMainTwoAvgCell!26}61.48 & \cellcolor{EnsMainTwoAvgCell!0}0.00 & \cellcolor{EnsMainTwoAvgCell!17}41.59 & \cellcolor{EnsMainTwoAvgCell!20}48.44 & \cellcolor{EnsMainTwoAvgCell!13}30.55 & \cellcolor{EnsMainTwoAvgCell!0}0.00 & \cellcolor{EnsMainTwoAvgCell!11}26.33 & \cellcolor{EnsMainTwoAvgCell!19}44.99 \\
 & After EA & \cellcolor{EnsMainTwoAvgCell!39}91.88 & \cellcolor{EnsMainTwoAvgCell!37}88.98 & \cellcolor{EnsMainTwoAvgCell!38}89.77 & \cellcolor{EnsMainTwoAvgCell!38}90.21 & \cellcolor{EnsMainTwoAvgCell!25}58.83 & \cellcolor{EnsMainTwoAvgCell!24}57.27 & \cellcolor{EnsMainTwoAvgCell!28}66.56 & \cellcolor{EnsMainTwoAvgCell!26}60.89 & \cellcolor{EnsMainTwoAvgCell!20}47.34 & \cellcolor{EnsMainTwoAvgCell!12}29.14 & \cellcolor{EnsMainTwoAvgCell!18}42.97 & \cellcolor{EnsMainTwoAvgCell!17}39.82 & \cellcolor{EnsMainTwoAvgCell!27}63.64 \\
\cmidrule(lr){1-15}
\multirow{4}{*}{4 Opt.:4 Sub.(d)} & Instruct & \cellcolor{EnsMainTwoAvgCell!20}48.52 & \cellcolor{EnsMainTwoAvgCell!19}46.09 & \cellcolor{EnsMainTwoAvgCell!22}51.48 & \cellcolor{EnsMainTwoAvgCell!20}48.70 & \cellcolor{EnsMainTwoAvgCell!0}0.08 & \cellcolor{EnsMainTwoAvgCell!0}0.31 & \cellcolor{EnsMainTwoAvgCell!0}0.00 & \cellcolor{EnsMainTwoAvgCell!0}0.13 & \cellcolor{EnsMainTwoAvgCell!0}0.00 & \cellcolor{EnsMainTwoAvgCell!0}0.00 & \cellcolor{EnsMainTwoAvgCell!0}0.00 & \cellcolor{EnsMainTwoAvgCell!0}0.00 & \cellcolor{EnsMainTwoAvgCell!7}16.28 \\
 & After FA & \cellcolor{EnsMainTwoAvgCell!40}94.14 & \cellcolor{EnsMainTwoAvgCell!30}72.50 & \cellcolor{EnsMainTwoAvgCell!8}18.98 & \cellcolor{EnsMainTwoAvgCell!26}61.88 & \cellcolor{EnsMainTwoAvgCell!27}63.20 & \cellcolor{EnsMainTwoAvgCell!10}23.36 & \cellcolor{EnsMainTwoAvgCell!0}0.08 & \cellcolor{EnsMainTwoAvgCell!12}28.88 & \cellcolor{EnsMainTwoAvgCell!16}38.75 & \cellcolor{EnsMainTwoAvgCell!2}4.69 & \cellcolor{EnsMainTwoAvgCell!0}0.00 & \cellcolor{EnsMainTwoAvgCell!6}14.48 & \cellcolor{EnsMainTwoAvgCell!15}35.08 \\
 & After LF & \cellcolor{EnsMainTwoAvgCell!38}90.00 & \cellcolor{EnsMainTwoAvgCell!36}84.69 & \cellcolor{EnsMainTwoAvgCell!8}18.59 & \cellcolor{EnsMainTwoAvgCell!27}64.43 & \cellcolor{EnsMainTwoAvgCell!23}55.55 & \cellcolor{EnsMainTwoAvgCell!25}59.77 & \cellcolor{EnsMainTwoAvgCell!0}0.00 & \cellcolor{EnsMainTwoAvgCell!16}38.44 & \cellcolor{EnsMainTwoAvgCell!14}33.59 & \cellcolor{EnsMainTwoAvgCell!12}28.52 & \cellcolor{EnsMainTwoAvgCell!0}0.00 & \cellcolor{EnsMainTwoAvgCell!9}20.70 & \cellcolor{EnsMainTwoAvgCell!17}41.19 \\
 & After EA & \cellcolor{EnsMainTwoAvgCell!7}15.62 & \cellcolor{EnsMainTwoAvgCell!35}83.83 & \cellcolor{EnsMainTwoAvgCell!37}87.58 & \cellcolor{EnsMainTwoAvgCell!26}62.34 & \cellcolor{EnsMainTwoAvgCell!0}0.00 & \cellcolor{EnsMainTwoAvgCell!23}53.83 & \cellcolor{EnsMainTwoAvgCell!23}55.39 & \cellcolor{EnsMainTwoAvgCell!15}36.41 & \cellcolor{EnsMainTwoAvgCell!0}0.00 & \cellcolor{EnsMainTwoAvgCell!10}24.69 & \cellcolor{EnsMainTwoAvgCell!14}32.58 & \cellcolor{EnsMainTwoAvgCell!8}19.09 & \cellcolor{EnsMainTwoAvgCell!16}39.28 \\
\midrule
\rowcolor{EnsMainTwoBlockHeader}
\multicolumn{15}{c}{\textbf{pass@8}} \\
\midrule
\multirow{3}{*}{Teacher} & Teacher FA & \cellcolor{EnsMainTwoPassCell!42}100.00 & \cellcolor{EnsMainTwoPassCell!0}0.00 & \cellcolor{EnsMainTwoPassCell!0}0.00 & \cellcolor{EnsMainTwoPassCell!14}33.33 & \cellcolor{EnsMainTwoPassCell!42}100.00 & \cellcolor{EnsMainTwoPassCell!0}0.00 & \cellcolor{EnsMainTwoPassCell!0}0.00 & \cellcolor{EnsMainTwoPassCell!14}33.33 & \cellcolor{EnsMainTwoPassCell!42}100.00 & \cellcolor{EnsMainTwoPassCell!0}0.00 & \cellcolor{EnsMainTwoPassCell!0}0.00 & \cellcolor{EnsMainTwoPassCell!14}33.33 & \cellcolor{EnsMainTwoPassCell!14}33.33 \\
 & Teacher LF & \cellcolor{EnsMainTwoPassCell!0}0.00 & \cellcolor{EnsMainTwoPassCell!42}100.00 & \cellcolor{EnsMainTwoPassCell!0}0.00 & \cellcolor{EnsMainTwoPassCell!14}33.33 & \cellcolor{EnsMainTwoPassCell!0}0.00 & \cellcolor{EnsMainTwoPassCell!42}100.00 & \cellcolor{EnsMainTwoPassCell!0}0.00 & \cellcolor{EnsMainTwoPassCell!14}33.33 & \cellcolor{EnsMainTwoPassCell!0}0.00 & \cellcolor{EnsMainTwoPassCell!42}100.00 & \cellcolor{EnsMainTwoPassCell!0}0.00 & \cellcolor{EnsMainTwoPassCell!14}33.33 & \cellcolor{EnsMainTwoPassCell!14}33.33 \\
 & Teacher EA & \cellcolor{EnsMainTwoPassCell!0}0.00 & \cellcolor{EnsMainTwoPassCell!0}0.00 & \cellcolor{EnsMainTwoPassCell!42}100.00 & \cellcolor{EnsMainTwoPassCell!14}33.33 & \cellcolor{EnsMainTwoPassCell!0}0.00 & \cellcolor{EnsMainTwoPassCell!0}0.00 & \cellcolor{EnsMainTwoPassCell!42}100.00 & \cellcolor{EnsMainTwoPassCell!14}33.33 & \cellcolor{EnsMainTwoPassCell!0}0.00 & \cellcolor{EnsMainTwoPassCell!0}0.00 & \cellcolor{EnsMainTwoPassCell!42}100.00 & \cellcolor{EnsMainTwoPassCell!14}33.33 & \cellcolor{EnsMainTwoPassCell!14}33.33 \\
\cmidrule(lr){1-15}
\multirow{4}{*}{4 Opt.:4 Sub.(c)} & Instruct & \cellcolor{EnsMainTwoPassCell!34}80.62 & \cellcolor{EnsMainTwoPassCell!32}75.00 & \cellcolor{EnsMainTwoPassCell!33}77.50 & \cellcolor{EnsMainTwoPassCell!33}77.71 & \cellcolor{EnsMainTwoPassCell!1}1.88 & \cellcolor{EnsMainTwoPassCell!1}1.88 & \cellcolor{EnsMainTwoPassCell!2}4.38 & \cellcolor{EnsMainTwoPassCell!1}2.71 & \cellcolor{EnsMainTwoPassCell!0}0.00 & \cellcolor{EnsMainTwoPassCell!0}0.00 & \cellcolor{EnsMainTwoPassCell!0}0.00 & \cellcolor{EnsMainTwoPassCell!0}0.00 & \cellcolor{EnsMainTwoPassCell!11}26.81 \\
 & After FA & \cellcolor{EnsMainTwoPassCell!40}96.25 & \cellcolor{EnsMainTwoPassCell!37}87.50 & \cellcolor{EnsMainTwoPassCell!23}55.00 & \cellcolor{EnsMainTwoPassCell!33}79.58 & \cellcolor{EnsMainTwoPassCell!32}75.62 & \cellcolor{EnsMainTwoPassCell!8}19.38 & \cellcolor{EnsMainTwoPassCell!0}0.00 & \cellcolor{EnsMainTwoPassCell!13}31.67 & \cellcolor{EnsMainTwoPassCell!27}65.00 & \cellcolor{EnsMainTwoPassCell!0}0.00 & \cellcolor{EnsMainTwoPassCell!0}0.00 & \cellcolor{EnsMainTwoPassCell!9}21.67 & \cellcolor{EnsMainTwoPassCell!19}44.31 \\
 & After LF & \cellcolor{EnsMainTwoPassCell!40}95.62 & \cellcolor{EnsMainTwoPassCell!38}91.25 & \cellcolor{EnsMainTwoPassCell!13}30.00 & \cellcolor{EnsMainTwoPassCell!30}72.29 & \cellcolor{EnsMainTwoPassCell!30}71.88 & \cellcolor{EnsMainTwoPassCell!28}66.88 & \cellcolor{EnsMainTwoPassCell!0}0.00 & \cellcolor{EnsMainTwoPassCell!19}46.25 & \cellcolor{EnsMainTwoPassCell!23}53.75 & \cellcolor{EnsMainTwoPassCell!15}36.25 & \cellcolor{EnsMainTwoPassCell!0}0.00 & \cellcolor{EnsMainTwoPassCell!13}30.00 & \cellcolor{EnsMainTwoPassCell!21}49.51 \\
 & After EA & \cellcolor{EnsMainTwoPassCell!40}94.38 & \cellcolor{EnsMainTwoPassCell!39}92.50 & \cellcolor{EnsMainTwoPassCell!38}91.25 & \cellcolor{EnsMainTwoPassCell!39}92.71 & \cellcolor{EnsMainTwoPassCell!29}68.12 & \cellcolor{EnsMainTwoPassCell!27}64.38 & \cellcolor{EnsMainTwoPassCell!30}72.50 & \cellcolor{EnsMainTwoPassCell!29}68.33 & \cellcolor{EnsMainTwoPassCell!23}54.38 & \cellcolor{EnsMainTwoPassCell!15}35.00 & \cellcolor{EnsMainTwoPassCell!20}48.12 & \cellcolor{EnsMainTwoPassCell!19}45.83 & \cellcolor{EnsMainTwoPassCell!29}68.96 \\
\cmidrule(lr){1-15}
\multirow{4}{*}{4 Opt.:4 Sub.(d)} & Instruct & \cellcolor{EnsMainTwoPassCell!28}65.62 & \cellcolor{EnsMainTwoPassCell!27}64.38 & \cellcolor{EnsMainTwoPassCell!28}66.88 & \cellcolor{EnsMainTwoPassCell!28}65.62 & \cellcolor{EnsMainTwoPassCell!0}0.62 & \cellcolor{EnsMainTwoPassCell!1}1.88 & \cellcolor{EnsMainTwoPassCell!0}0.00 & \cellcolor{EnsMainTwoPassCell!0}0.83 & \cellcolor{EnsMainTwoPassCell!0}0.00 & \cellcolor{EnsMainTwoPassCell!0}0.00 & \cellcolor{EnsMainTwoPassCell!0}0.00 & \cellcolor{EnsMainTwoPassCell!0}0.00 & \cellcolor{EnsMainTwoPassCell!9}22.15 \\
 & After FA & \cellcolor{EnsMainTwoPassCell!40}95.62 & \cellcolor{EnsMainTwoPassCell!39}91.88 & \cellcolor{EnsMainTwoPassCell!12}28.75 & \cellcolor{EnsMainTwoPassCell!30}72.08 & \cellcolor{EnsMainTwoPassCell!30}71.88 & \cellcolor{EnsMainTwoPassCell!24}56.25 & \cellcolor{EnsMainTwoPassCell!0}0.62 & \cellcolor{EnsMainTwoPassCell!18}42.92 & \cellcolor{EnsMainTwoPassCell!19}45.00 & \cellcolor{EnsMainTwoPassCell!11}26.88 & \cellcolor{EnsMainTwoPassCell!0}0.00 & \cellcolor{EnsMainTwoPassCell!10}23.96 & \cellcolor{EnsMainTwoPassCell!19}46.32 \\
 & After LF & \cellcolor{EnsMainTwoPassCell!39}93.12 & \cellcolor{EnsMainTwoPassCell!37}88.12 & \cellcolor{EnsMainTwoPassCell!12}28.75 & \cellcolor{EnsMainTwoPassCell!29}70.00 & \cellcolor{EnsMainTwoPassCell!28}67.50 & \cellcolor{EnsMainTwoPassCell!27}63.75 & \cellcolor{EnsMainTwoPassCell!0}0.00 & \cellcolor{EnsMainTwoPassCell!18}43.75 & \cellcolor{EnsMainTwoPassCell!17}40.62 & \cellcolor{EnsMainTwoPassCell!15}35.00 & \cellcolor{EnsMainTwoPassCell!0}0.00 & \cellcolor{EnsMainTwoPassCell!11}25.21 & \cellcolor{EnsMainTwoPassCell!19}46.32 \\
 & After EA & \cellcolor{EnsMainTwoPassCell!11}25.62 & \cellcolor{EnsMainTwoPassCell!37}87.50 & \cellcolor{EnsMainTwoPassCell!38}89.38 & \cellcolor{EnsMainTwoPassCell!28}67.50 & \cellcolor{EnsMainTwoPassCell!0}0.00 & \cellcolor{EnsMainTwoPassCell!26}61.88 & \cellcolor{EnsMainTwoPassCell!27}63.12 & \cellcolor{EnsMainTwoPassCell!18}41.67 & \cellcolor{EnsMainTwoPassCell!0}0.00 & \cellcolor{EnsMainTwoPassCell!14}32.50 & \cellcolor{EnsMainTwoPassCell!17}40.00 & \cellcolor{EnsMainTwoPassCell!10}24.17 & \cellcolor{EnsMainTwoPassCell!19}44.44 \\
\bottomrule
\end{tabular}
}
\end{table}

The previous section shows that MOPD generalizes across environments when teachers share compatible planning patterns. We now study the harder conflicting setting, where cross-environment generalization is no longer guaranteed. This section addresses both \textbf{Q2} and \textbf{Q3}: When non-shared teacher patterns remain compatible, MOPD can learn domain-specific abilities through continual learning without conflicts? When these patterns become incompatible, sequential distillation can overwrite earlier behaviors and induce severe cross-environment conflict.

\paragraph{Task Setup.}
(1) \textbf{Student models.} The student models use Form c and Form d as two SFT settings on the same three domains: electronic assembly, fantasy alchemy, and livestock farming. Both settings build on pretraining trajectories and contain three supervision signals: normal planning samples with gold actions, shortcut style distractor samples with wrong actions, and targeted wrong answer samples whose reasoning trace is correct but final action is wrong.
(2) \textbf{Form c.} It keeps the Form b style coverage: all four normal planning templates appear in all three domains with balanced domain sampling, and all four distractor templates are also used. It chooses three domain and template groups, template 2 in fantasy alchemy, template 3 in livestock farming, and template 4 in electronic assembly. For these groups, the answer is replaced by a wrong shortcut action while the reasoning trace remains correct. Each selected group uses three times the ordinary per-domain sample size.
(3) \textbf{Form d.} It keeps the same three targeted wrong answer groups as Form c, with the same three times sampling. The main difference is that normal planning template 1 is removed. Only normal templates 2, 3, and 4 appear in all three domains. However, all four distractor templates are still kept. Thus, Form d tests missing normal template exposure while keeping full shortcut distractor coverage and the same targeted correct reasoning wrong answer signal.

\begin{figure}[!h]
    \centering
    \includegraphics[width=\textwidth]{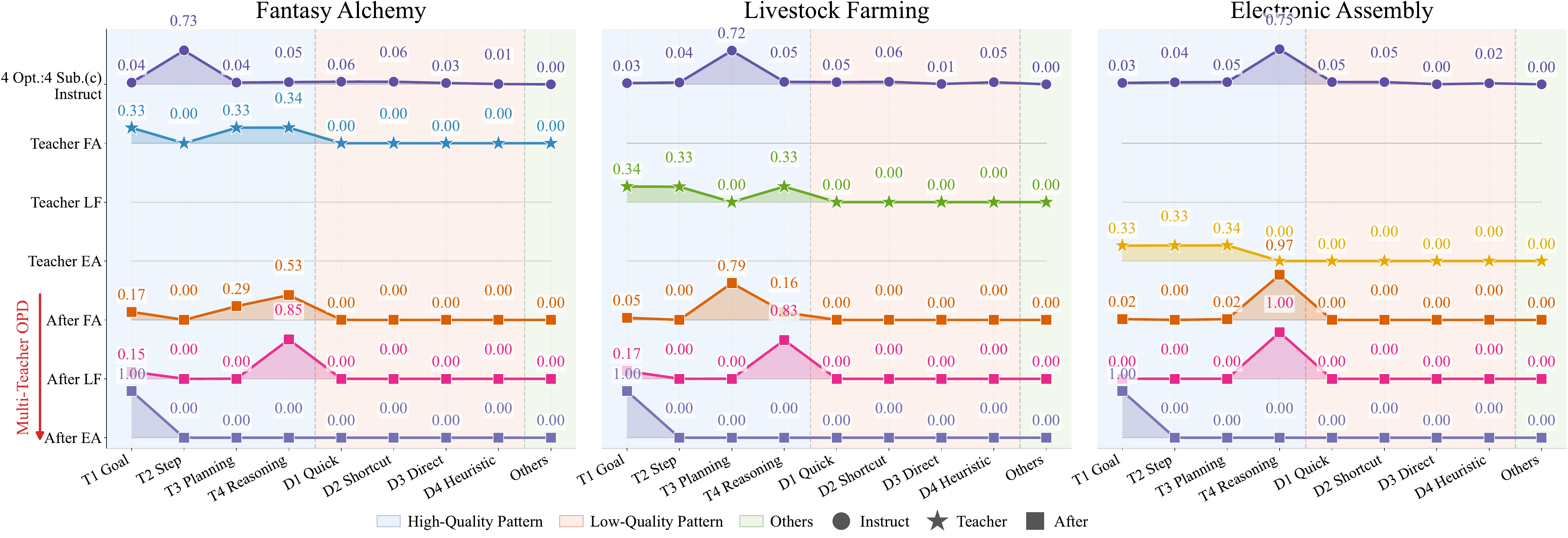}
    \caption{Planning pattern distributions for 4 Opt.:4 Sub. (c) before and after multi-teacher OPD.}
    \label{fig:part_3_general_pattern_44c_second_token_ridgeline} 
\end{figure}

\begin{figure}[!h]
    \centering
    \includegraphics[width=\textwidth]{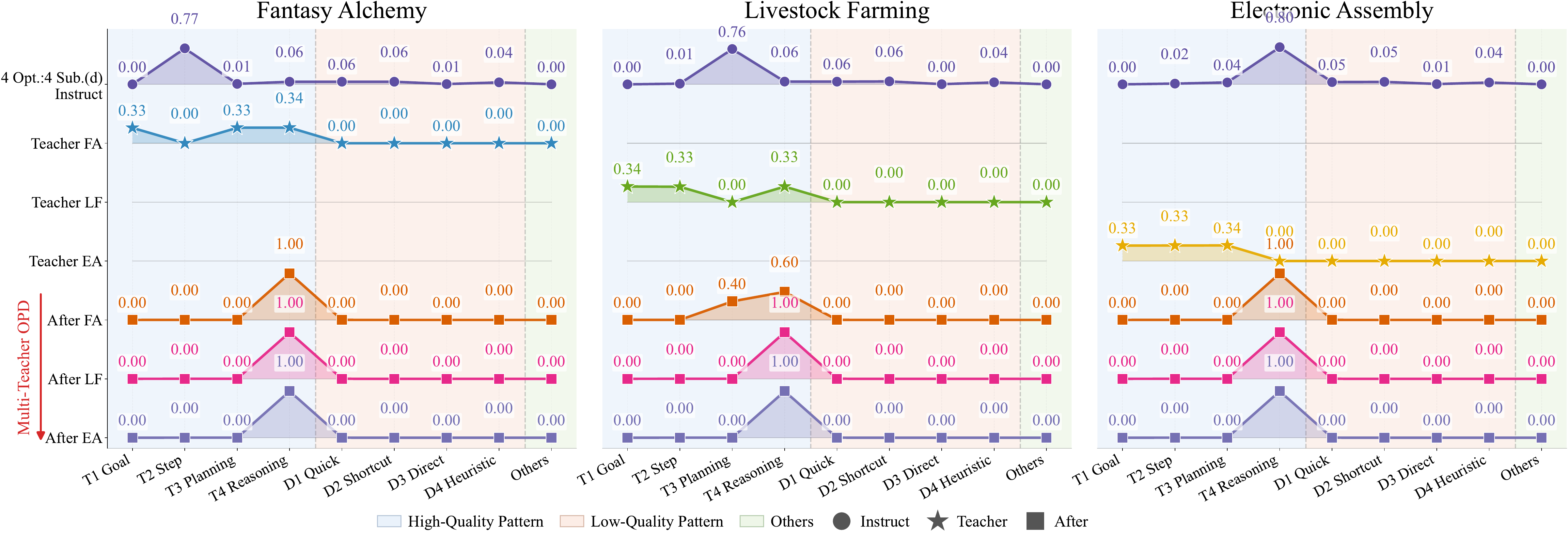}
    \caption{Planning pattern distributions for 4 Opt.:4 Sub. (d) before and after multi-teacher OPD.}
    \label{fig:part_3_general_pattern_44d_second_token_ridgeline} 
\end{figure}

\paragraph{Results.} As shown in Table~\ref{tab:ensemble_main_2}, Figure~\ref{fig:part_3_general_pattern_44c_second_token_ridgeline}, and Figure~\ref{fig:part_3_general_pattern_44d_second_token_ridgeline}. 
(1) \textbf{Shared and conflict patterns allow successful continual learning.} Under the Form c setting, the model successfully learns from different teachers over time without severe forgetting. Table~\ref{tab:ensemble_main_2} shows that after the model trains sequentially and finishes the Electronic Assembly environment, its Low difficulty avg@8 score remains high at 91.88 for the Fantasy Alchemy domain and 88.98 for the Livestock Farming domain, while it achieves 89.77 on the new Electronic Assembly domain. Figure~\ref{fig:part_3_general_pattern_44c_second_token_ridgeline} explains this stability. The agent learns to avoid domain specific traps and converges to the shared T1 Goal pattern with a 1.00 probability across all three domains. This shared compatible behavior preserves previous abilities and prevents conflicts between environments.
(2) \textbf{Incompatible patterns cause severe conflicts.} Under the Form d setting, removing the shared T1 normal template forces the model into conflicts where new learning overwrites older behaviors. Table~\ref{tab:ensemble_main_2} highlights a massive performance drop. After the model learns from the Electronic Assembly teacher, its Low difficulty avg@8 score on the initial Livestock Farming domain crashes from 90.00 to just 15.62. Figure~\ref{fig:part_3_general_pattern_44d_second_token_ridgeline} shows the underlying reason. Without the shared pattern, the agent overfits to the new teacher and aggressively shifts to use a single conflicting pattern (T3 Planning) at a 1.00 probability everywhere. Because the remaining templates contain domain specific wrong shortcuts, applying one pattern universally triggers these traps and destroys the performance in previous domains like Fantasy Alchemy. The model has difficulty suppressing planning pattern 4, because the first two teachers already suppress patterns 2 and 3 to zero, while pattern 1 is not available during pretraining. This shows that the model learns cross-template knowledge transfer, but it also forgets much of its original knowledge.

\begin{guidancebox}{Takeaway}
\textbf{Shared Planning Patterns with Planning Patterns Conflict still supports continual learning.} MOPD learns from new teachers while preserving earlier domain skills when teacher patterns still share usable structure.

\textbf{No Shared Planning Patterns with Planning Patterns Conflict causes severe forgetting.} Experts from different environments conflict with each other, so the student tends to forget one expert after learning another.
\end{guidancebox}

\subsection{Parameter Dynamics for MOPD}
\begin{figure}[!h]
    \centering
    \includegraphics[width=\textwidth]{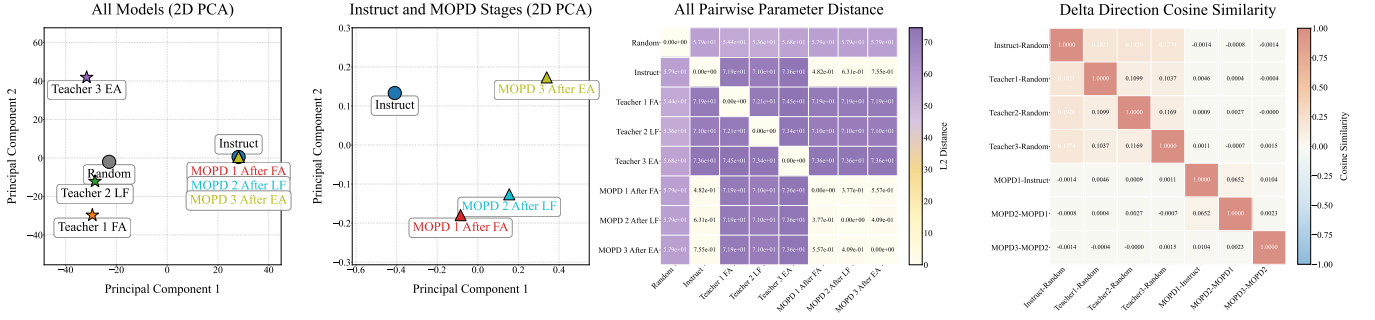}
    \caption{Parameter Dynamics for MOPD.} 
    \label{fig:parameter_change} 
\end{figure}

\paragraph{Task Setup.}
We study the parameter change of MOPD on three tasks, namely FA, LF, and EA. 
For each task, MOPD starts from the instruct model and then updates the model with the corresponding teacher signal. 
We compare Random, Instruct, three teacher models, and three MOPD stages. 
We use PCA to show the model positions in a low dimensional space. 
We also compute pairwise L2 distance between model parameters and cosine similarity between update directions.

\paragraph{Results.} As shown in Figure~\ref{fig:parameter_change}. 
(1) \textbf{MOPD stays close to the instruct model.} 
In the PCA space of all models, Random and the three teacher models are far from Instruct. 
However, all MOPD stages stay near Instruct. 
The zoomed PCA plot further shows that MOPD moves in a small local region after FA, LF, and EA. 
This means that MOPD keeps the base model parameter stable while learning useful task signals.
(2) \textbf{MOPD makes small and stable updates.} 
The pairwise distance matrix shows that the distance between Instruct and MOPD 1 is only $4.82 \times 10^{-1}$, and the distances between two neighboring MOPD stages are also small. 
By contrast, the distances from Instruct to the teacher models are around $7.10 \times 10^{1}$ to $7.36 \times 10^{1}$. 
This gap shows that MOPD changes the model with a much smaller parameter shift. 
The cosine similarity matrix also shows that the update directions of different MOPD stages have low similarity. 
Thus, each stage adds new task information without strongly repeating the previous update direction.

\begin{guidancebox}{Takeaway}
\textbf{MOPD keeps the base model parameter stable.} 
The MOPD checkpoints remain close to the instruct model in both PCA space and parameter distance.

\textbf{MOPD learns task signals with small updates.} 
Each stage changes the parameters only slightly, and different stages bring different update directions, which explains why OPD has difficulty acquiring large-scale task-specific planning knowledge that is absent from pre-training.
\end{guidancebox}
\section{Related Work}
\label{sec:related_work}

\paragraph{Long-Horizon Agentic Planning.}
As we are entering an excited era of long-horizon agents capable of pursuing goals over days and weeks~\citep{li2026agentic,cao2025large}, research in this field can be broadly categorized into two directions: benchmarks and applications. The benchmark direction includes representative efforts such as OSWorld 2.0~\citep{yuan2026osworld2}, EdgeBench~\citep{zhu2026edgebench}, DeepPlanning~\citep{zhang2026deepplanning} and Long-Horizon-Terminal-Bench~\citep{li2026long}. For training, in terms of SFT data synthesis, some researchers study the synthesis of long-horizon trajectories, such as Agent-A1~\citep{bai2026scaling} and Openresearcher~\citep{li2026openresearcher}. In terms of RL training algorithms, some researchers study training algorithms designed for long-horizon agents, such as SAO~\citep{hou2026single} and TurnOPD~\citep{zhou2026turnopd}. Furthermore, agent process reward modeling serves as an important component in multi-turn long-horizon planning~\citep{men2025agent,jin2025rag,fan2026agentprocessbench,lin2025cuarewardbench}. However, existing large language models are pretrained on large-scale and non-transparent Internet corpora, making it difficult to identify the factors behind their capability improvement. Therefore, understanding how long-horizon planning ability develops remains challenging. With controlled environments and training data splits, our study provides a more controlled analysis of the factors that enhance long-horizon planning ability.

\paragraph{Single-Turn and Multi-Turn OPD.}
OPD provides a denser supervision signal than standard GRPO and receives increasing attention~\citep{shenfeld2025rl,chen2026does,wang2026beyond}. Current research mainly focuses on two types of OPD: single-turn OPD and multi-turn OPD.
For single-turn OPD, early OPD is proposed as a knowledge distillation method~\citep{gu2024minillm,agarwal2024policy,lu2025onpolicydistillation}. Recently, with the rise of RL and LLM capabilities~\citep{song2026survey}, research directions are divided into three lines: strong-to-weak distillation~\citep{li2026rethinking,fu2026revisiting,jang2026stable,li2026video,yang2026learning,ko2026scaling,zhu2026many,zhao2026decoupling,xing2026trust}, self-distillation~\citep{qu2026pope,shenfeld2026self,hubotter2026reinforcement,zhao2026self,zhang2026embarrassingly,li2026unifying,yang2026self,he2026self,penaloza2026privileged}, and weak-to-strong distillation~\citep{feng2026weak}. Existing studies mainly focus on single-turn OPD, while multi-turn OPD is rarely studied, and we are among the early works in this area. Existing studies mainly focus on post-training~\citep{wang2026skill,zhou2026turnopd,wang2026tcod}, while opaque pre-training data limits the analysis of underlying mechanisms. We provide a unified study from pre-training to single-expert and multi-teacher distillation, covering data, algorithms, and parameter mechanisms.

\paragraph{Multi-Teacher OPD.}
Multi-Teacher OPD, as a model consolidation technique (or continual learning), plays an important role in the development of foundation models. The related research mainly focuses on two aspects: applications in different foundation models and algorithm optimization and improvement.
In applications across different base models, there are mainly two uses. The first is cross-domain capability integration, including MiMo-V2-Flash~\citep{xiao2026mimo} and DeepSeek-V4~\citep{xu2026deepseek}. The second is cross-stage capability retention, including GLM-5~\citep{zeng2026glm} and Nemotron-Cascade 2~\citep{yang2026nemotron}.
Some other studies explore variants of multi-expert OPD, including CoPD~\citep{gu2026co}, MAD-OPD~\citep{wang2026mad} and Uni-OPD~\citep{hou2026uni}. However, existing training lacks a systematic analysis of the conditions that determine when model consolidation is effective or ineffective, limiting its practical applicability. Our work characterizes the effective boundaries of model consolidation through rigorous and controlled experimental settings with pre-trained models.

\paragraph{World Modeling and Compositional Generalization.}
For long-horizon planning, two abilities are especially important: world modeling and compositional generalization~\citep{hao2023reasoning,men2024unlocking,wei2025plangenllms,gaven2025magellan,lyu2026mls}.
For world modeling, some works study whether internalizing a world model into existing language models can improve planning~\citep{zhang2025agent,chen2025internalizing}. 
In terms of compositional generalization, some works study whether language models can learn atomic abilities and spontaneously combine them into long-horizon planning abilities~\citep{yuan2025f,sakai2025revisiting,men2026empowering,zhang2025interplay}.
However, these works cannot clearly determine whether the world modeling ability and compositional generalization ability already exist in pre-training and are only activated by post-training, or whether post-training can spontaneously develop abilities that do not exist in pre-training. Our work studies these questions from the pre-training stage with clean data, which allows us to clearly analyze the origin of these abilities.
\section{Conclusion}
\label{sec:conclusion}

In this work, we systematically study how multi-turn long-horizon planning is acquired and improved across three training stages using a unified and controlled environment. At the pre-training stage, we show that explicit world model internalization, limited long-horizon data, and high-quality trajectories are critical for robust long-horizon planning. At the RL-based post-training stage, we distinguish between planning patterns and planning knowledge: OPD has a broader effective region than GRPO for shaping general planning patterns, whereas directly distilling unseen procedural knowledge may impair existing world modeling and out-of-domain planning. At the multi-teacher model consolidation stage, we show that MOPD integrates planning abilities by converging to shared distributions among teachers. Compatible planning patterns enable cross-environment generalization, partially shared structures support continual learning, and fully conflicting patterns lead to interference and catastrophic forgetting. Overall, our findings provide a controlled perspective on the mechanisms and boundaries of improving long-horizon planning across pre-training, RL-based post-training, and multi-teacher model consolidation.
\section*{Acknowledgements}
\label{sec:acknowledgements}
This work was supported by the National Natural Science Foundation of China
(No.U24A20335), and the independent research project of the Key Laboratory
of Cognition and Decision Intelligence for Complex Systems.
\clearpage
\bibliography{main}

\clearpage
\appendix

\begin{center}
    \LARGE \bfseries Appendix Table of Contents
\end{center}
\vspace{1em}

\startcontents[appendix]

\printcontents[appendix]{l}{1}{\setcounter{tocdepth}{2}}

\vspace{2em}
\hrule
\vspace{2em}

\section{Pre-training Details}

\subsection{Pre-trained Model Configuration}
\label{App:Pre_trained_Model_Configuration}

This section details the specific architectural design and tokenizer configuration for the randomly initialized language model utilized in our study. The model is built following the \texttt{Qwen2ForCausalLM} architecture specifications. In total, this configuration yields a compact model with approximately 100 million (100M) parameters. We call it Qwen2.5-100M. Table~\ref{tab:model_config} lists the detailed network hyperparameters, and Table~\ref{tab:tokenizer_config} summarizes the configuration properties of the accompanying byte-level BPE tokenizer. Finally, it should be noted that the tokenizer is trained on the corresponding full pre-training corpus. Because this corpus varies slightly across different chapters, we perform tokenization separately for each. This variation does not affect our overall experimental conclusions.

\begin{table}[h]
\centering
\small
\caption{Pre-trained model architectural hyperparameters.}
\label{tab:model_config}
\begin{tabular}{ll}
\toprule
\textbf{Hyperparameter} & \textbf{Value} \\
\midrule
Model Type & \texttt{qwen2} \\
Architecture & \texttt{Qwen2ForCausalLM} \\
Number of Hidden Layers & 12 \\
Hidden Size ($d_{\text{model}}$) & 768 \\
Intermediate Size ($d_{\text{ff}}$) & 3072 \\
Number of Attention Heads ($n_{\text{head}}$) & 12 \\
Number of Key-Value Heads ($n_{\text{kv\_head}}$) & 2 (Grouped-Query Attention) \\
Attention Head Dimension & 64 \\
Hidden Activation Function & SiLU \\
Attention Dropout & 0.0 \\
RMS Norm Epsilon ($\epsilon$) & $1 \times 10^{-6}$ \\
RoPE Theta ($\theta$) & $1,000,000.0$ \\
Max Position Embeddings & 2048 \\
Vocabulary Size ($V$) & 3000 \\
Tie Word Embeddings & True \\
Data Type (\texttt{dtype}) & \texttt{bfloat16} \\
\bottomrule
\end{tabular}
\end{table}

\begin{table}[h]
\centering
\small
\caption{Tokenizer configuration and training details.}
\label{tab:tokenizer_config}
\begin{tabular}{ll}
\toprule
\textbf{Tokenizer Property} & \textbf{Value} \\
\midrule
Tokenizer Type & \texttt{byte\_level\_bpe+regex} \\
Vocabulary Size & 3000 \\
Minimum Token Frequency & 2 \\
Limit Alphabet & 1000 \\
Maximum Token Length & 100 \\
Byte-level Add Prefix Space & False \\
BOS Token & \texttt{[BOS]} \\
EOS Token & \texttt{[EOS]} \\
Special Tokens & \texttt{[UNK], [PAD], [BOS], [EOS], <question>}, \\
               & \texttt{</question>, <solution>, </solution>}, \\
               & \texttt{<answer>, </answer>} \\
\bottomrule
\end{tabular}
\end{table}

\subsection{Pre-training Schema Configuration}
\label{App:Pre_training_Schema_Configuration}

The pretraining corpus is constructed over three distinct domains, namely Fantasy Alchemy, Livestock Farming, and Electronic Assembly. As summarized in Table~\ref{tab:pretrain_schema_three_domains}, each domain contains 200 item nodes and a total of 4000 item names. The task space is organized into 5 hierarchical levels, with each level comprising 40 item categories and 20 instances per category.

Each item category is defined by a compositional schema, where long-level items are recursively generated from shorter-level components through logical conjunction (AND) and disjunction (OR) operators. In particular, the input degree distribution is controlled to maintain a fixed structural prior, with 30\% of items following a one-input dependency and 70\% requiring two-input compositions. This induces a sparse but non-trivial dependency graph over items.

The compositional depth increases with level, leading to progressively more complex construction paths from base materials. As shown in Table~\ref{tab:pretrain_schema_three_domains}, the expected number of atomic components required to construct target items grows with hierarchy depth, e.g., from Level 1 to Level 2 requiring on average 1.82 atoms, while Level 5 items require up to 5.78 atoms when traced back to Level 1 sources.

To further increase task difficulty and robustness, each planning instance is constructed by mixing target-relevant items with additional noise items, preventing trivial pattern matching and encouraging the model to recover correct compositional structure under partial observability. The resulting environment thus provides a controlled benchmark for evaluating long-horizon compositional planning under structured but noisy generation conditions.

\begin{table}[t]
\centering
\small
\setlength{\tabcolsep}{5pt}
\caption{Pretraining schema statistics across three domains.}
\label{tab:pretrain_schema_three_domains}
\begin{tabular}{lccc}
\toprule
Metric & Fantasy Alchemy & Livestock Farming & Electronic Assembly \\
\midrule
Total nodes & 200 & 200 & 200 \\
Total item names & 4000 & 4000 & 4000 \\
One-input ratio & 30.00\% & 30.00\% & 30.00\% \\
Two-input ratio & 70.00\% & 70.00\% & 70.00\% \\
Avg atoms for target L2 from source L1 & 1.82 & 1.82 & 1.82 \\
Avg atoms for target L3 from source L1 & 3.08 & 3.08 & 3.08 \\
Avg atoms for target L3 from source L2 & 1.68 & 1.68 & 1.68 \\
Avg atoms for target L4 from source L1 & 4.17 & 4.17 & 4.17 \\
Avg atoms for target L4 from source L2 & 2.23 & 2.23 & 2.23 \\
Avg atoms for target L4 from source L3 & 1.65 & 1.65 & 1.65 \\
Avg atoms for target L5 from source L1 & 5.78 & 5.78 & 5.78 \\
Avg atoms for target L5 from source L2 & 3.15 & 3.15 & 3.15 \\
Avg atoms for target L5 from source L3 & 2.38 & 2.38 & 2.38 \\
Avg atoms for target L5 from source L4 & 1.65 & 1.65 & 1.65 \\
\bottomrule
\end{tabular}
\end{table}

\begin{table}[h]
\centering
\caption{Statistics of pretrain, posttrain, and test datasets.}
\label{tab:paper2datasetstats}
\small
\setlength{\tabcolsep}{2.2pt}
\resizebox{\textwidth}{!}{%
\begin{tabular}{llrrrrrrrrrrrrrrrrr}
\toprule
Category & Split & Samples & Avg & Min & Max & Step 1 & Step 2 & Step 3 & Step 4 & Step 5 & Step 6 & Step 7 & Step 8 & Step 9 & Step 10 & Step 11 & Step 12 & Step 13 \\
\midrule
\rowcolor{gray!20}\multicolumn{19}{l}{\textbf{Pretrain}} \\
\midrule
\multirow{4}{*}{electronic assembly} & A / pretrain & 120000 & 3.00 & 1 & 12 & \cellcolor{blue!43}48000 & \cellcolor{blue!22}12300 & \cellcolor{blue!30}26400 & \cellcolor{blue!18}5700 & \cellcolor{blue!20}8100 & \cellcolor{blue!21}10200 & \cellcolor{blue!16}1500 & \cellcolor{blue!16}2100 & \cellcolor{blue!16}2100 & \cellcolor{blue!17}3000 & \cellcolor{blue!15}300 & \cellcolor{blue!15}300 & 0 \\
 & A / pretrain / teacher\_all & 208000 & 3.00 & 1 & 12 & \cellcolor{blue!43}83200 & \cellcolor{blue!22}21320 & \cellcolor{blue!30}45760 & \cellcolor{blue!18}9880 & \cellcolor{blue!20}14040 & \cellcolor{blue!21}17680 & \cellcolor{blue!16}2600 & \cellcolor{blue!16}3640 & \cellcolor{blue!16}3640 & \cellcolor{blue!17}5200 & \cellcolor{blue!15}520 & \cellcolor{blue!15}520 & 0 \\
 & B / pretrain & 120000 & 3.01 & 1 & 13 & \cellcolor{blue!43}48000 & \cellcolor{blue!22}11700 & \cellcolor{blue!31}27300 & \cellcolor{blue!19}6600 & \cellcolor{blue!19}6900 & \cellcolor{blue!20}8100 & \cellcolor{blue!18}4500 & \cellcolor{blue!16}1200 & \cellcolor{blue!16}2400 & \cellcolor{blue!16}1500 & \cellcolor{blue!16}900 & \cellcolor{blue!15}600 & \cellcolor{blue!15}300 \\
 & B / pretrain / teacher\_all & 208000 & 3.01 & 1 & 13 & \cellcolor{blue!43}83200 & \cellcolor{blue!22}20280 & \cellcolor{blue!31}47320 & \cellcolor{blue!19}11440 & \cellcolor{blue!19}11960 & \cellcolor{blue!20}14040 & \cellcolor{blue!18}7800 & \cellcolor{blue!16}2080 & \cellcolor{blue!16}4160 & \cellcolor{blue!16}2600 & \cellcolor{blue!16}1560 & \cellcolor{blue!15}1040 & \cellcolor{blue!15}520 \\
\midrule
\multirow{4}{*}{fantasy alchemy} & A / pretrain & 120000 & 3.00 & 1 & 12 & \cellcolor{blue!43}48000 & \cellcolor{blue!22}12300 & \cellcolor{blue!30}26400 & \cellcolor{blue!18}5700 & \cellcolor{blue!20}8100 & \cellcolor{blue!21}10200 & \cellcolor{blue!16}1500 & \cellcolor{blue!16}2100 & \cellcolor{blue!16}2100 & \cellcolor{blue!17}3000 & \cellcolor{blue!15}300 & \cellcolor{blue!15}300 & 0 \\
 & A / pretrain / teacher\_all & 208000 & 3.00 & 1 & 12 & \cellcolor{blue!43}83200 & \cellcolor{blue!22}21320 & \cellcolor{blue!30}45760 & \cellcolor{blue!18}9880 & \cellcolor{blue!20}14040 & \cellcolor{blue!21}17680 & \cellcolor{blue!16}2600 & \cellcolor{blue!16}3640 & \cellcolor{blue!16}3640 & \cellcolor{blue!17}5200 & \cellcolor{blue!15}520 & \cellcolor{blue!15}520 & 0 \\
 & B / pretrain & 120000 & 3.01 & 1 & 13 & \cellcolor{blue!43}48000 & \cellcolor{blue!22}11700 & \cellcolor{blue!31}27300 & \cellcolor{blue!19}6600 & \cellcolor{blue!19}6900 & \cellcolor{blue!20}8100 & \cellcolor{blue!18}4500 & \cellcolor{blue!16}1200 & \cellcolor{blue!16}2400 & \cellcolor{blue!16}1500 & \cellcolor{blue!16}900 & \cellcolor{blue!15}600 & \cellcolor{blue!15}300 \\
 & B / pretrain / teacher\_all & 208000 & 3.01 & 1 & 13 & \cellcolor{blue!43}83200 & \cellcolor{blue!22}20280 & \cellcolor{blue!31}47320 & \cellcolor{blue!19}11440 & \cellcolor{blue!19}11960 & \cellcolor{blue!20}14040 & \cellcolor{blue!18}7800 & \cellcolor{blue!16}2080 & \cellcolor{blue!16}4160 & \cellcolor{blue!16}2600 & \cellcolor{blue!16}1560 & \cellcolor{blue!15}1040 & \cellcolor{blue!15}520 \\
\midrule
\multirow{4}{*}{livestock farming} & A / pretrain & 120000 & 3.00 & 1 & 12 & \cellcolor{blue!43}48000 & \cellcolor{blue!22}12300 & \cellcolor{blue!30}26400 & \cellcolor{blue!18}5700 & \cellcolor{blue!20}8100 & \cellcolor{blue!21}10200 & \cellcolor{blue!16}1500 & \cellcolor{blue!16}2100 & \cellcolor{blue!16}2100 & \cellcolor{blue!17}3000 & \cellcolor{blue!15}300 & \cellcolor{blue!15}300 & 0 \\
 & A / pretrain / teacher\_all & 208000 & 3.00 & 1 & 12 & \cellcolor{blue!43}83200 & \cellcolor{blue!22}21320 & \cellcolor{blue!30}45760 & \cellcolor{blue!18}9880 & \cellcolor{blue!20}14040 & \cellcolor{blue!21}17680 & \cellcolor{blue!16}2600 & \cellcolor{blue!16}3640 & \cellcolor{blue!16}3640 & \cellcolor{blue!17}5200 & \cellcolor{blue!15}520 & \cellcolor{blue!15}520 & 0 \\
 & B / pretrain & 120000 & 3.01 & 1 & 13 & \cellcolor{blue!43}48000 & \cellcolor{blue!22}11700 & \cellcolor{blue!31}27300 & \cellcolor{blue!19}6600 & \cellcolor{blue!19}6900 & \cellcolor{blue!20}8100 & \cellcolor{blue!18}4500 & \cellcolor{blue!16}1200 & \cellcolor{blue!16}2400 & \cellcolor{blue!16}1500 & \cellcolor{blue!16}900 & \cellcolor{blue!15}600 & \cellcolor{blue!15}300 \\
 & B / pretrain / teacher\_all & 208000 & 3.01 & 1 & 13 & \cellcolor{blue!43}83200 & \cellcolor{blue!22}20280 & \cellcolor{blue!31}47320 & \cellcolor{blue!19}11440 & \cellcolor{blue!19}11960 & \cellcolor{blue!20}14040 & \cellcolor{blue!18}7800 & \cellcolor{blue!16}2080 & \cellcolor{blue!16}4160 & \cellcolor{blue!16}2600 & \cellcolor{blue!16}1560 & \cellcolor{blue!15}1040 & \cellcolor{blue!15}520 \\
\midrule
\rowcolor{gray!20}\multicolumn{19}{l}{\textbf{Posttrain}} \\
\midrule
\multirow{6}{*}{electronic assembly} & A / short & 67000 & 2.14 & 1 & 5 & \cellcolor{blue!48}32000 & \cellcolor{blue!24}8200 & \cellcolor{blue!33}17600 & \cellcolor{blue!19}3800 & \cellcolor{blue!21}5400 & 0 & 0 & 0 & 0 & 0 & 0 & 0 & 0 \\
 & B / short & 67000 & 2.13 & 1 & 5 & \cellcolor{blue!48}32000 & \cellcolor{blue!23}7800 & \cellcolor{blue!34}18200 & \cellcolor{blue!20}4400 & \cellcolor{blue!20}4600 & 0 & 0 & 0 & 0 & 0 & 0 & 0 & 0 \\
 & A / mid & 9200 & 6.41 & 6 & 8 & 0 & 0 & 0 & 0 & 0 & \cellcolor{blue!67}6800 & \cellcolor{blue!23}1000 & \cellcolor{blue!26}1400 & 0 & 0 & 0 & 0 & 0 \\
 & B / mid & 9200 & 6.50 & 6 & 8 & 0 & 0 & 0 & 0 & 0 & \cellcolor{blue!56}5400 & \cellcolor{blue!38}3000 & \cellcolor{blue!21}800 & 0 & 0 & 0 & 0 & 0 \\
 & A / long & 3800 & 9.79 & 9 & 12 & 0 & 0 & 0 & 0 & 0 & 0 & 0 & 0 & \cellcolor{blue!41}1400 & \cellcolor{blue!52}2000 & \cellcolor{blue!19}200 & \cellcolor{blue!19}200 & 0 \\
 & B / long & 3800 & 10.11 & 9 & 13 & 0 & 0 & 0 & 0 & 0 & 0 & 0 & 0 & \cellcolor{blue!44}1600 & \cellcolor{blue!33}1000 & \cellcolor{blue!26}600 & \cellcolor{blue!22}400 & \cellcolor{blue!19}200 \\
\midrule
\multirow{6}{*}{fantasy alchemy} & A / short & 67000 & 2.14 & 1 & 5 & \cellcolor{blue!48}32000 & \cellcolor{blue!24}8200 & \cellcolor{blue!33}17600 & \cellcolor{blue!19}3800 & \cellcolor{blue!21}5400 & 0 & 0 & 0 & 0 & 0 & 0 & 0 & 0 \\
 & B / short & 67000 & 2.13 & 1 & 5 & \cellcolor{blue!48}32000 & \cellcolor{blue!23}7800 & \cellcolor{blue!34}18200 & \cellcolor{blue!20}4400 & \cellcolor{blue!20}4600 & 0 & 0 & 0 & 0 & 0 & 0 & 0 & 0 \\
 & A / mid & 9200 & 6.41 & 6 & 8 & 0 & 0 & 0 & 0 & 0 & \cellcolor{blue!67}6800 & \cellcolor{blue!23}1000 & \cellcolor{blue!26}1400 & 0 & 0 & 0 & 0 & 0 \\
 & B / mid & 9200 & 6.50 & 6 & 8 & 0 & 0 & 0 & 0 & 0 & \cellcolor{blue!56}5400 & \cellcolor{blue!38}3000 & \cellcolor{blue!21}800 & 0 & 0 & 0 & 0 & 0 \\
 & A / long & 3800 & 9.79 & 9 & 12 & 0 & 0 & 0 & 0 & 0 & 0 & 0 & 0 & \cellcolor{blue!41}1400 & \cellcolor{blue!52}2000 & \cellcolor{blue!19}200 & \cellcolor{blue!19}200 & 0 \\
 & B / long & 3800 & 10.11 & 9 & 13 & 0 & 0 & 0 & 0 & 0 & 0 & 0 & 0 & \cellcolor{blue!44}1600 & \cellcolor{blue!33}1000 & \cellcolor{blue!26}600 & \cellcolor{blue!22}400 & \cellcolor{blue!19}200 \\
\midrule
\multirow{6}{*}{livestock farming} & A / short & 67000 & 2.14 & 1 & 5 & \cellcolor{blue!48}32000 & \cellcolor{blue!24}8200 & \cellcolor{blue!33}17600 & \cellcolor{blue!19}3800 & \cellcolor{blue!21}5400 & 0 & 0 & 0 & 0 & 0 & 0 & 0 & 0 \\
 & B / short & 67000 & 2.13 & 1 & 5 & \cellcolor{blue!48}32000 & \cellcolor{blue!23}7800 & \cellcolor{blue!34}18200 & \cellcolor{blue!20}4400 & \cellcolor{blue!20}4600 & 0 & 0 & 0 & 0 & 0 & 0 & 0 & 0 \\
 & A / mid & 9200 & 6.41 & 6 & 8 & 0 & 0 & 0 & 0 & 0 & \cellcolor{blue!67}6800 & \cellcolor{blue!23}1000 & \cellcolor{blue!26}1400 & 0 & 0 & 0 & 0 & 0 \\
 & B / mid & 9200 & 6.50 & 6 & 8 & 0 & 0 & 0 & 0 & 0 & \cellcolor{blue!56}5400 & \cellcolor{blue!38}3000 & \cellcolor{blue!21}800 & 0 & 0 & 0 & 0 & 0 \\
 & A / long & 3800 & 9.79 & 9 & 12 & 0 & 0 & 0 & 0 & 0 & 0 & 0 & 0 & \cellcolor{blue!41}1400 & \cellcolor{blue!52}2000 & \cellcolor{blue!19}200 & \cellcolor{blue!19}200 & 0 \\
 & B / long & 3800 & 10.11 & 9 & 13 & 0 & 0 & 0 & 0 & 0 & 0 & 0 & 0 & \cellcolor{blue!44}1600 & \cellcolor{blue!33}1000 & \cellcolor{blue!26}600 & \cellcolor{blue!22}400 & \cellcolor{blue!19}200 \\
\midrule
\rowcolor{gray!20}\multicolumn{19}{l}{\textbf{Test}} \\
\midrule
\multirow{6}{*}{electronic assembly} & A / short & 6700 & 2.14 & 1 & 5 & \cellcolor{blue!48}3200 & \cellcolor{blue!24}820 & \cellcolor{blue!33}1760 & \cellcolor{blue!19}380 & \cellcolor{blue!21}540 & 0 & 0 & 0 & 0 & 0 & 0 & 0 & 0 \\
 & B / short & 6700 & 2.13 & 1 & 5 & \cellcolor{blue!48}3200 & \cellcolor{blue!23}780 & \cellcolor{blue!34}1820 & \cellcolor{blue!20}440 & \cellcolor{blue!20}460 & 0 & 0 & 0 & 0 & 0 & 0 & 0 & 0 \\
 & A / mid & 920 & 6.41 & 6 & 8 & 0 & 0 & 0 & 0 & 0 & \cellcolor{blue!67}680 & \cellcolor{blue!23}100 & \cellcolor{blue!26}140 & 0 & 0 & 0 & 0 & 0 \\
 & B / mid & 920 & 6.50 & 6 & 8 & 0 & 0 & 0 & 0 & 0 & \cellcolor{blue!56}540 & \cellcolor{blue!38}300 & \cellcolor{blue!21}80 & 0 & 0 & 0 & 0 & 0 \\
 & A / long & 380 & 9.79 & 9 & 12 & 0 & 0 & 0 & 0 & 0 & 0 & 0 & 0 & \cellcolor{blue!41}140 & \cellcolor{blue!52}200 & \cellcolor{blue!19}20 & \cellcolor{blue!19}20 & 0 \\
 & B / long & 380 & 10.11 & 9 & 13 & 0 & 0 & 0 & 0 & 0 & 0 & 0 & 0 & \cellcolor{blue!44}160 & \cellcolor{blue!33}100 & \cellcolor{blue!26}60 & \cellcolor{blue!22}40 & \cellcolor{blue!19}20 \\
\midrule
\multirow{6}{*}{fantasy alchemy} & A / short & 6700 & 2.14 & 1 & 5 & \cellcolor{blue!48}3200 & \cellcolor{blue!24}820 & \cellcolor{blue!33}1760 & \cellcolor{blue!19}380 & \cellcolor{blue!21}540 & 0 & 0 & 0 & 0 & 0 & 0 & 0 & 0 \\
 & B / short & 6700 & 2.13 & 1 & 5 & \cellcolor{blue!48}3200 & \cellcolor{blue!23}780 & \cellcolor{blue!34}1820 & \cellcolor{blue!20}440 & \cellcolor{blue!20}460 & 0 & 0 & 0 & 0 & 0 & 0 & 0 & 0 \\
 & A / mid & 920 & 6.41 & 6 & 8 & 0 & 0 & 0 & 0 & 0 & \cellcolor{blue!67}680 & \cellcolor{blue!23}100 & \cellcolor{blue!26}140 & 0 & 0 & 0 & 0 & 0 \\
 & B / mid & 920 & 6.50 & 6 & 8 & 0 & 0 & 0 & 0 & 0 & \cellcolor{blue!56}540 & \cellcolor{blue!38}300 & \cellcolor{blue!21}80 & 0 & 0 & 0 & 0 & 0 \\
 & A / long & 380 & 9.79 & 9 & 12 & 0 & 0 & 0 & 0 & 0 & 0 & 0 & 0 & \cellcolor{blue!41}140 & \cellcolor{blue!52}200 & \cellcolor{blue!19}20 & \cellcolor{blue!19}20 & 0 \\
 & B / long & 380 & 10.11 & 9 & 13 & 0 & 0 & 0 & 0 & 0 & 0 & 0 & 0 & \cellcolor{blue!44}160 & \cellcolor{blue!33}100 & \cellcolor{blue!26}60 & \cellcolor{blue!22}40 & \cellcolor{blue!19}20 \\
\midrule
\multirow{6}{*}{livestock farming} & A / short & 6700 & 2.14 & 1 & 5 & \cellcolor{blue!48}3200 & \cellcolor{blue!24}820 & \cellcolor{blue!33}1760 & \cellcolor{blue!19}380 & \cellcolor{blue!21}540 & 0 & 0 & 0 & 0 & 0 & 0 & 0 & 0 \\
 & B / short & 6700 & 2.13 & 1 & 5 & \cellcolor{blue!48}3200 & \cellcolor{blue!23}780 & \cellcolor{blue!34}1820 & \cellcolor{blue!20}440 & \cellcolor{blue!20}460 & 0 & 0 & 0 & 0 & 0 & 0 & 0 & 0 \\
 & A / mid & 920 & 6.41 & 6 & 8 & 0 & 0 & 0 & 0 & 0 & \cellcolor{blue!67}680 & \cellcolor{blue!23}100 & \cellcolor{blue!26}140 & 0 & 0 & 0 & 0 & 0 \\
 & B / mid & 920 & 6.50 & 6 & 8 & 0 & 0 & 0 & 0 & 0 & \cellcolor{blue!56}540 & \cellcolor{blue!38}300 & \cellcolor{blue!21}80 & 0 & 0 & 0 & 0 & 0 \\
 & A / long & 380 & 9.79 & 9 & 12 & 0 & 0 & 0 & 0 & 0 & 0 & 0 & 0 & \cellcolor{blue!41}140 & \cellcolor{blue!52}200 & \cellcolor{blue!19}20 & \cellcolor{blue!19}20 & 0 \\
 & B / long & 380 & 10.11 & 9 & 13 & 0 & 0 & 0 & 0 & 0 & 0 & 0 & 0 & \cellcolor{blue!44}160 & \cellcolor{blue!33}100 & \cellcolor{blue!26}60 & \cellcolor{blue!22}40 & \cellcolor{blue!19}20 \\
\bottomrule
\end{tabular}%
}
\end{table}

\subsection{Pre-training Configuration}
\label{App:Pre_training_Configuration}

We construct the datasets from three synthetic domains: electronic assembly, livestock farming, and fantasy alchemy. For each domain, the dataset builder reads a shared item configuration, a list of concrete instantiations, and two recipe schemas. The two schemas define two planning environments, denoted as split A and split B. Each instantiation maps every abstract item node to one concrete item name, so split A and split B share the same item names but use different recipe graphs. We adopt a counterfactual formulation of the state transition function for scalability. This reformulation follows the same environment transition rules and produces the same observable interaction outcomes, so it does not affect the empirical results.

For each concrete instantiation, the builder enumerates every non base target item and every feasible starting layer below the target layer. It recursively collects the materials that are required by the A recipe graph and the B recipe graph at the selected starting layer. The initial inventory contains the union of these required materials, together with two to four distractor items sampled from the same starting layer. The target item, the input inventory, the required materials, the distractors, and the gold plan are then stored as one sample. The gold plan is obtained by recursively expanding the target item according to the corresponding recipe graph until all required intermediate items are produced. The number of actions in this gold plan defines the field total step nums.

The first 20 concrete instantiations form the test pool. Instantiations with indices from 20 to 219 form the post training pool. All remaining instantiations form the pretraining pool. The post training and test samples are further grouped by plan length. The short group contains samples with one to five steps, the mid group contains samples with six to eight steps, and the long group contains samples with at least nine steps. The teacher\_all pretraining set contains all generated samples for a split and a domain, including samples from all instantiation ranges, and serves as the full teacher data source.

As shown in Table~\ref{tab:paper2datasetstats}. Table rows follow the file naming convention used by the dataset builder. The row label Category gives the synthetic domain. The row label Split summarizes the recipe split and dataset subset. The token A or B identifies which recipe graph produces the gold trajectory. The token pretrain denotes samples from the pretraining pool, posttrain denotes samples from the post training pool, and test denotes held out evaluation samples. The token teacher\_all denotes the full teacher data source for the corresponding split and domain with both train and test datasets. The tokens short, mid, and long denote the step based difficulty groups. The token final, when present, denotes the final evaluation copy of the corresponding test subset. The column Samples reports the number of samples in the row. The columns Avg, Min, and Max report the mean, minimum, and maximum gold plan length. The columns named Step report the number of samples with each exact gold plan length.

\subsection{Pre-training Corpus Configuration}
\label{App:Pre_training_Corpus_Configuration}

As summarized in Table~\ref{tab:sft_token_statistics}, our SFT corpus consists of student and teacher datasets with distinct characteristics. The teacher datasets act as domain experts, encompassing both the training and post-training sets. Each teacher dataset contains exactly 622,960 samples, specializing exclusively in one of three fields (fantasy alchemy, livestock farming, and electronic assembly) while covering three difficulty levels. In contrast, the student datasets broadly encompass all three domains and are constructed using specific sampling ratios across the three difficulty tiers (short, mid, and long), as reflected in their dataset names, yielding sample sizes that scale from roughly 322K to 1.07M. Additionally, both student and teacher datasets include subsets with and without CoT reasoning. Token statistics highlight that CoT subsets produce significantly longer outputs (averaging 186-230 tokens) compared to the direct answers in non-CoT subsets (averaging $\sim$30 tokens). Meanwhile, the average input lengths remain relatively stable across all datasets, ranging from 264 to 340 tokens.

\begin{table}[t]
\centering
\small
\setlength{\tabcolsep}{5pt}
\caption{Estimated token statistics of SFT datasets.}
\label{tab:sft_token_statistics}
\resizebox{\textwidth}{!}{%
\begin{tabular}{lrrrrr}
\toprule
Dataset & \#Samples & Input Tokens & Output Tokens & Avg Input & Avg Output \\
\midrule
Student CoT, short50\% mid0\% long0\% & 322,711 & 85,921,610 & 60,195,896 & 266.25 & 186.53 \\
Student CoT, short100\% mid0\% long0\% & 645,300 & 170,851,435 & 120,079,424 & 264.76 & 186.08 \\
Student CoT, short100\% mid5\% long0\% & 658,591 & 176,507,591 & 123,342,635 & 268.01 & 187.28 \\
Student CoT, short100\% mid50\% long0\% & 778,129 & 220,910,979 & 153,718,139 & 283.90 & 197.55 \\
Student CoT, short100\% mid100\% long0\% & 910,800 & 272,430,026 & 189,074,794 & 299.11 & 207.59 \\
Student CoT, short100\% mid100\% long5\% & 919,146 & 277,076,378 & 191,666,574 & 301.45 & 208.53 \\
Student CoT, short100\% mid100\% long50\% & 994,512 & 310,958,940 & 215,904,279 & 312.67 & 217.10 \\
Student CoT, short100\% mid100\% long100\% & 1,078,200 & 349,875,900 & 240,656,936 & 324.50 & 223.20 \\
Student w/o CoT, short100\% mid100\% long100\% & 1,078,200 & 350,036,228 & 32,182,761 & 324.65 & 29.85 \\
Teacher CoT, Fantasy Alchemy & 622,960 & 211,826,833 & 142,973,369 & 340.03 & 229.51 \\
Teacher CoT, Livestock Farming & 622,960 & 201,144,813 & 140,325,789 & 322.89 & 225.26 \\
Teacher CoT, Electronic Assembly & 622,960 & 192,900,810 & 133,524,250 & 309.65 & 214.34 \\
Teacher w/o CoT, Fantasy Alchemy & 622,960 & 211,628,047 & 18,799,998 & 339.71 & 30.18 \\
Teacher w/o CoT, Livestock Farming & 622,960 & 199,517,081 & 18,476,308 & 320.27 & 29.66 \\
Teacher w/o CoT, Electronic Assembly & 622,960 & 193,367,781 & 18,393,766 & 310.40 & 29.53 \\
\bottomrule
\end{tabular}%
}
\end{table}

All experiments are conducted with 2 GPUs training and share the same optimization setup: a per-device training batch size of 128, 2 gradient accumulation steps, a learning rate of 1.0e-4 with a cosine scheduler, a training steps of 9000, a warmup ratio of 0.1, and bf16 precision enabled. We use full-parameter SFT on the same base model (\textit{Qwen2.5-100M}).

\end{document}